\documentclass{defaltmlr}

\usepackage{amsmath,amsfonts,bm}

\def\eqref#1{equation~\ref{#1}}

\def\1{\bm{1}}

\DeclareMathAlphabet{\mathsfit}{\encodingdefault}{\sfdefault}{m}{sl}
\SetMathAlphabet{\mathsfit}{bold}{\encodingdefault}{\sfdefault}{bx}{n}

\usepackage{url}
\usepackage[table]{xcolor}
\usepackage[dvipsnames]{xcolor}
\usepackage{graphicx}
\usepackage{geometry}
\usepackage{booktabs}
\usepackage[T1]{fontenc}
\usepackage{fontawesome}
\usepackage{fancyhdr}
\usepackage{tocloft}
\usepackage{enumitem}
\usepackage{etoc}
\usepackage{titletoc}
\usepackage{tcolorbox}
\usepackage{listings}
\newcommand\DoToC{%
  \startcontents
  \printcontents{}{1}{\noindent \textbf{\large{Table of Contents}}\vskip3pt\vskip5pt}
  \vskip3pt\vskip5pt
}

\usepackage{color}
\usepackage[dvipsnames]{xcolor}
\definecolor{JalapenoRed}{RGB}{183,21,64}
\definecolor{Belize}{RGB}{41,128,185}
\definecolor{Amour}{RGB}{238,82,83}
\usepackage{colortbl}

\usepackage{hyperref}

\usepackage{flushend}
\usepackage{caption}
\usepackage{tabularx}
\usepackage{graphicx}
\usepackage{amsfonts}
\usepackage{amsmath}
\usepackage{amsthm}

\usepackage{multirow}
\usepackage{xcolor}
\usepackage{booktabs}
\usepackage{float}
\usepackage{balance}
\usepackage{enumitem}
\usepackage[table]{xcolor}
\usepackage{pgfplots}
\setlist[itemize]{noitemsep, topsep=0pt, partopsep=0pt}

\usepackage{xspace}
\usepackage{threeparttable}
\usepackage{url}
\usepackage{makecell}
\usepackage{titletoc}

\usepackage{caption}
\usepackage{xcolor}
\usepackage{color}
\usepackage{colortbl}
\usepackage{pifont}

\usepackage{graphicx}
\usepackage{enumitem}
\usepackage{longtable}

\usepackage{arydshln}
\usepackage{flushend}
\usepackage{caption}
\usepackage{tabularx}
\usepackage{graphicx}
\usepackage{amsfonts}
\usepackage{amsmath}
\usepackage{amsthm}

\usepackage{multirow}
\usepackage{xcolor}
\usepackage{booktabs}
\usepackage{float}
\usepackage{balance}
\usepackage{enumitem}
\usepackage[table]{xcolor}
\setlist[itemize]{noitemsep, topsep=0pt, partopsep=0pt}

\usepackage{xspace}
\usepackage{threeparttable}
\usepackage{url}

\usepackage{titletoc}
\usepackage{tocloft}

\usepackage{subcaption}

\newcommand{\ie}{\textit{i}.\textit{e}.}
\newcommand{\eg}{\textit{e}.\textit{g}.}

\newcommand{\etc}{\textit{etc.}}

\usepackage{stfloats}
\usepackage{cuted}    
\usepackage{capt-of}  

\usepackage[subrefformat=parens,labelformat=empty]{subcaption}

\usepackage{afterpage}

\def\dataset{\texttt{WorldST}}
\def\format{\texttt{MiniST}}
\def\model{\texttt{UrbanFM}}
\def\benchmark{\texttt{EvalST}}

\definecolor{fbApp}{HTML}{ffe4e3}
\newcommand{\emoji}[2][1.2em]{\raisebox{-0.2\height}{\includegraphics[height=#1]{#2}}}
\newcommand{\rowc}{\rowcolor{gray!10}}

\newcommand{\bluesquare}{
    \tikz[baseline=1.1ex]{
        \node[fill, cyan, opacity=0.05, rounded corners, minimum width=3.ex, minimum height=2.2ex, anchor=center, shape=rectangle] at (1ex,2ex) {};
    }
}

\newcommand{\greensquare}{
    \tikz[baseline=1.1ex]{
        \node[fill, green, opacity=0.05, rounded corners, minimum width=3.ex, minimum height=2.2ex, anchor=center, shape=rectangle] at (1ex,2ex) {};
    }
}

\usepackage[ruled]{algorithm2e} 

\usepackage{amsmath}
\usepackage{amssymb}

\definecolor{myblue}{HTML}{e5f5fd}
\definecolor{mygreen}{HTML}{e6ffe6}
\definecolor{highlightcolor}{HTML}{2da02d}

\usepackage{wrapfig}

\hypersetup{linkcolor=cyan}
\usepackage{mathtools}
\usepackage{ulem}
\usepackage{pifont}
\usepackage{bbding}

\usepackage{tcolorbox}
\usepackage{makecell}

\makeatletter
\let\orig@fnsymbol\@fnsymbol
\def\@fnsymbol#1{\ifcase#1\or\relax\else\orig@fnsymbol{#1}\fi}
\makeatother

\title{UrbanFM: Scaling Urban \\Spatio-Temporal Foundation Models}
\vspace{4cm}

\author{
\parbox{\textwidth}{
Wei Chen\textsuperscript{1,2}, Yuqian Wu\textsuperscript{1}, Junle Chen\textsuperscript{2}, Xiaofang Zhou\textsuperscript{2$^*$}, Yuxuan Liang\textsuperscript{1$^*$}
}
}

\affiliation{\textsuperscript{1}HKUST(GZ), \textsuperscript{2}HKUST}

\abstract{
Urban systems, as dynamic complex systems, continuously generate spatio-temporal data streams that encode the fundamental laws of human mobility and city evolution. While AI for Science has witnessed the transformative power of foundation models in disciplines like genomics and meteorology, urban computing remains fragmented due to ``scenario-specific'' models, which are overfitted to specific regions or tasks, hindering their generalizability. To bridge this gap and advance spatio-temporal foundation models for urban systems, we adopt scaling as the central perspective and systematically investigate two key questions: \textit{\textbf{what}} to scale and \textit{\textbf{how}} to scale. Grounded in first-principles analysis, we identify three critical dimensions: \textbf{\textit{heterogeneity}}, \textbf{\textit{correlation}}, and \textbf{\textit{dynamics}}, aligning these principles with the fundamental scientific properties of urban spatio-temporal data. Specifically, to address heterogeneity through \textbf{\textit{data scaling}}, we construct \dataset. This billion-scale corpus standardizes diverse physical signals, such as traffic flow and speed, from over 100 global cities into a unified data format. To enable \textbf{\textit{computation scaling}} for modeling correlations, we introduce the \format~unit, a novel split mechanism that discretizes continuous spatio-temporal fields into learnable computational units to unify representations of grid-based and sensor-based observations. Finally, addressing dynamics via \textbf{\textit{architecture scaling}}, we propose \model, a minimalist self-attention architecture designed with limited inductive biases to autonomously learn dynamic spatio-temporal dependencies from massive data. Furthermore, to ensure fair evaluation, we establish \benchmark, the largest-scale urban spatio-temporal benchmark to date. Extensive experiments demonstrate that \model~achieves remarkable zero-shot generalization across unseen cities and tasks, marking a pivotal first step toward large-scale pretrained urban spatio-temporal foundation models.
}

\correspondence{onedeanxxx@gmail.com,$^*$zxf@cse.ust.hk,$^*$yuxliang@outlook.com}
\date{\sffamily Feb 9, 2026}

\begin{document}

\maketitle

\makeatletter
\let\@fnsymbol\orig@fnsymbol
\makeatother
\section{Introduction}

Modern cities~\citep{dong2024defining}, functioning as dynamically evolving complex systems, continuously generate multi-source spatio-temporal data streams, encompassing a wide range of mobility information, from traffic flow to population movement. These ubiquitous data streams constantly depict the dynamic pulse of urban systems, encoding the fundamental scientific laws governing urban evolution~\citep{pappalardo2023future}.

Over the past two decade, statistical methods~\citep{cressie2011statistics} and spatio-temporal deep neural networks~\citep{jin2023spatio} have achieved success in specific urban tasks like traffic forecasting~\citep{jin2024survey} and imputation~\citep{gao2022generative}. However, these methods face the dilemma of ``scenario customization'': most models~\citep{marisca2024graph,nie2024imputeformer,cini2022filling,li2018diffusion} are optimized for single urban regions, specific time spans, or isolated scenarios, lacking universal modeling capabilities for general characteristics of urban systems.


\begin{figure}[t!]
    \centering
    \includegraphics[width=1.0\textwidth]{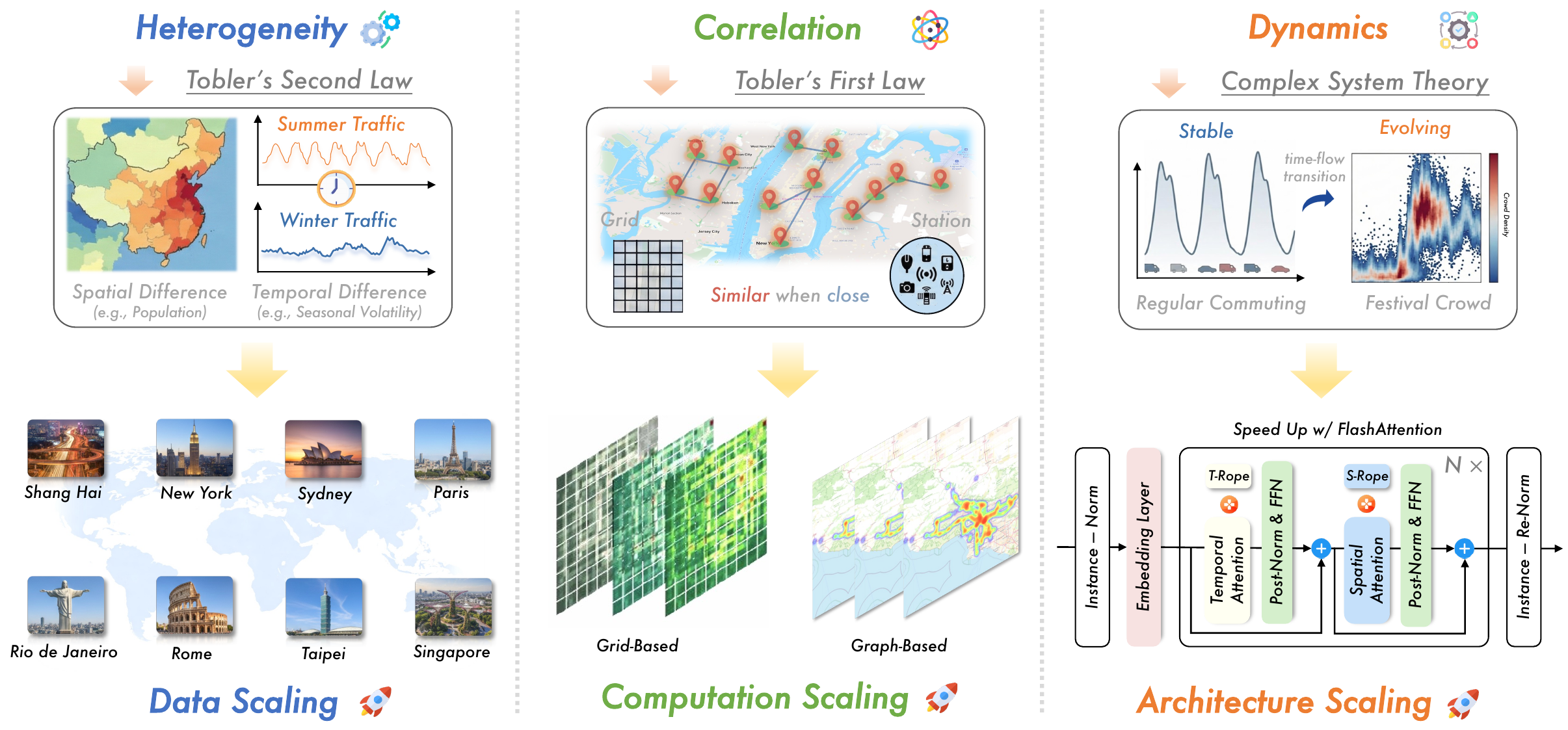}
    \caption{Top: The fundamental nature of spatio-temporal data and challenges of urban ST foundational models. Bottom: Three core perspectives of scaling the urban ST foundation model: data, computation, and architecture.}
    \label{fig:challenge}
\end{figure}

Reflecting on technological history, the ``bitter lesson''~\citep{sutton2019bitter} reveals that when computational scale surpasses critical thresholds, general-purpose paradigms leveraging massive data inevitably outperform complex human-designed heuristics. Although recent foundation models have achieved knowledge compression through scaling~\citep{kaplan2020scaling} and demonstrated remarkable generalization capabilities in language~\citep{brown2020language} and vision~\citep{hurst2024gpt} domains, as illustrated in Figure~\ref{fig:challenge} top, the scaling of spatio-temporal knowledge and intelligence for urban systems is still impeded by three fundamental scientific barriers:

\textit{Challenge \uppercase\expandafter{\romannumeral1}. Heterogeneity:} Tobler's Second Law~\citep{tobler1970computer} of Geography reveals that urban data exhibits significant spatio-temporal heterogeneity across domains. Specifically, the characteristics or relations of variables differ markedly across varying spatial locations or temporal points. This heterogeneity manifests as inherent non-uniformity and complexity, exemplified by spatial variances such as population density differences or temporal fluctuations like seasonal traffic volatility. Consequently, achieving \textit{data scaling}, which involves aggregating massive amounts of multi-source heterogeneous urban data to comprehensively cover the true distribution, serves as the fundamental solution to addressing heterogeneity.

\textit{Challenge \uppercase\expandafter{\romannumeral2}. Correlation:} Tobler's First Law~\citep{tobler1970computer} of Geography dictates that urban data possesses high spatio-temporal autocorrelation, meaning that attributes are interrelated across time and space, with proximity strengthening these associations. However, modeling these dependencies is further complicated by the coexistence of diverse microscopic and macroscopic data types (\eg, grid-based macro mobility patterns versus station-based micro sensor observations~\citep{wang2020deep}). Therefore, achieving \textit{computation scaling}, by leveraging these spatio-temporal autocorrelations to segment data into unified and appropriate computational units, constitutes the primary means to resolve the challenge of spatio-temporal correlation.

\textit{Challenge \uppercase\expandafter{\romannumeral3}. Dynamics:} Complex System Theory~\citep{thurner2018introduction} posits that urban systems are inherently non-stationary. Interactions between entities evolve continuously over time rather than remaining static, as exemplified by the stark contrast between regular commuting flows and irregular festival crowd dynamics. Existing architectures often inject strong inductive biases, such as static graph structures, which fail to capture these shifting patterns. Thus, achieving \textit{architecture scaling}, by designing minimalist architectures with limited inductive biases to autonomously learn time-evolving dependencies, represents the real challenge to mastering system dynamics.

To systematically address these challenges, as illustrated in Figure~\ref{fig:challenge} bottom, we design scalable, principled components tailored to distinct scientific properties. Our contributions are summarized as follows:

\begin{itemize}[leftmargin=4mm,parsep=4pt,topsep=4pt]

\item \textit{Data Scaling:} We construct the \dataset, employing a rigorous standardization pipeline to normalize heterogeneous signals from over 100 global cities into a unified data format, thereby ensuring distributional compatibility.

\item \textit{Computation Scaling:} We develop \format, a novel split mechanism that transforms disparate grid and sensor inputs into unified learnable units, enabling the scalable computation of spatio-temporal correlations.

\item \textit{Architecture Scaling:} We design \model, a minimalist adaptation of the self-attention architecture intended to minimize manual priors, which empowers the model to autonomously capture complex dynamic patterns.

\item Additionally, we establish \benchmark, the largest spatio-temporal evaluation benchmark to date. Experimental results demonstrate that \model~achieves remarkable zero-shot generalization in all tasks. We regard this work as a pivotal advancement toward large-scale spatio-temporal foundation models in urban science.

\end{itemize}

\section{Related Work}

\textbf{Spatio-Temporal Data Science.}
Spatio-temporal data science focuses on modeling the temporal evolution of spatially distributed variables, where forecasting and imputation constitute two fundamental tasks~\citep{liang2025foundation,chen2024deep}. These problems are central to a wide range of urban applications, including traffic management~\citep{ermagun2018spatiotemporal}, environmental monitoring~\citep{dietze2024near}, and public safety~\citep{wu2025beyond,wu2025mas4poi}. Early studies predominantly relied on statistical and classical machine learning approaches~\citep{shi2018machine}, such as ARIMA~\citep{box1970distribution}, VAR~\citep{biller2003modeling}, and Gaussian Processes~\citep{hamelijnck2021spatio}, which explicitly model temporal dependencies but scale poorly to large, heterogeneous urban systems. With the advent of deep learning, spatio-temporal neural models~\citep{shao2024exploring,chen2025select} have become the prevailing paradigm, typically integrating temporal sequence modeling with spatial dependency learning. Representative architectures include CNN-based models~\citep{wang2020deep,zhang2017deep,liu2020dynamic} for capturing local spatial correlations, RNN-~\citep{li2018diffusion,wang2022predrnn} or Transformer-based~\citep{guo2019attention,liu2023spatio} models for temporal dynamics, and graph neural networks~\citep{wu2019graph,song2020spatial,shao2022decoupled,han2024bigst,ma2025less} for explicitly encoding non-Euclidean spatial structures such as road networks or sensor graphs. Recent advances emphasize unified representation learning, with spatio-temporal graph neural networks~\citep{jin2023spatio,jin2024survey} serving as a prominent example. Despite strong empirical performance, most existing approaches remain task-specific and dataset-dependent, often requiring architectural modifications or retraining~\citep{chen2025eac,chen2025stttc} when transferred across domains, cities, or temporal horizons. \textit{These limitations motivate the exploration of more generalizable paradigms that move beyond single-task modeling toward reusable spatio-temporal models.}


\textbf{Urban Foundation Models.}
Inspired by the success of foundation models~\citep{chen2024catastrophic} in language~\citep{brown2020language} and vision~\citep{hurst2024gpt} domains, Urban Foundation Models~\citep{zhang2024urban}, also termed Spatio-Temporal Foundation Models~\citep{fang2026unraveling}, aim to learn general-purpose representations from large-scale, heterogeneous urban data to efficiently adapt to diverse downstream tasks. Early research typically employed self-supervised pre-training with pretext tasks, such as contrastive learning~\citep{qu2022fore,zheng2024treck} or masked auto-encoding~\citep{gao2024spatial,liu2025crossst}, to learn general representations. However, these methods were generally confined to single cities or domains, lacking cross-city zero-shot transferability. With the emergence of the foundation model paradigm, recent studies~\citep{yuan2024unist,yuan2024uniflow} have shifted towards jointly training shared spatio-temporal encoders on multi-city datasets, demonstrating enhanced capabilities for cross-city transfer forecasting. Subsequent research has further explored the efficacy of diverse backbones, including vanilla Transformers~\citep{han2025scalable}, Diffusion Transformers~\citep{yuan2025diffusion}, Graph Conventional Models~\citep{li2024opencity}, Mixture-of-Experts~\citep{tang2025unistd}, and State Space Models~\citep{an2025damba}, while often incorporating external augmentations such as frozen Language Models~\citep{yu2025bigcity,liu2025urbanmind}. Although recent strategies~\citep{zhong2025st,zhong2026st} further advocate for disentangling spatial and temporal components within the ``pre-training \& fine-tuning'' paradigm, the field remains predominantly focused on architectural novelty. \textit{In contrast, we revisit the fundamental scientific properties of urban spatio-temporal data and propose a principled multi-dimensional scaling approach to unlock authentic zero-shot generalization.}

\section{Preliminary}

\noindent \textbf{Definition 1 (\textit{Urban Spatio-temporal Data}).}
Spatio-temporal data describes the continuous observational data that varies over time at different spatial locations in a city. It is uniformly represented as a third-order tensor $ \mathbf{X} \in \mathbb{R}^{N \times T \times C}$, where $N$ denotes the number of spatial locations, $T$ indicates the number of temporal recording steps, and $C$ typically represents a single recorded element feature, including various spatio-temporal data in the city, such as traffic speed, road occupancy rate, crowd flows, taxi demand, bike usage, cellular traffic, and others. Along the temporal dimension, the data can be approximated as a continuously varying system with smooth transitions. However, due to the discrete nature of data collection, the spatial dimension is further divided into two types: \underline{\textit{(\romannumeral1). Sensor-based Type:}} This represents the changes in non-uniformly distributed spatial regions, recorded by $N$ sensors. The spatial relationships can be explicitly constructed by the geographic coordinates (latitude and longitude) of these sensors, forming a topological graph that is represented as an adjacency matrix $A$. \underline{\textit{(\romannumeral2). Grid-based Type:}} This represents the changes in uniformly distributed spatial regions, composed of $N$ grid cells. In this case, the spatial relationships are implicitly expressed by reshaping the $N$ regions into a $W \times H$ format. 

\noindent \textbf{Definition 2 (\textit{Urban Spatio-temporal Task}).}
    Based on these different data, urban spatio-temporal tasks are primarily categorized into the following two types:
    
    \begin{itemize}[leftmargin=2.5mm,parsep=1pt,topsep=0.5em]
        \item \underline{Forecasting:} Given $T_h$ historical time steps of spatio-temporal sample $X \in \mathbb{R}^{N \times T_h \times C}$, predict future signals $\hat{Y}$ over $T_f$ time steps. This task can be further divided into:
        
        \begin{itemize}[leftmargin=*,itemsep=0.3em,topsep=0.5em]
            \item[\scriptsize$\blacktriangleright$] Short-term Forecasting: Where $T_f$ is relatively small (\eg, $T_f \leq 12$, about 1 hour), focusing on near-future predictions.
            \item[\scriptsize$\blacktriangleright$] Long-term Forecasting: Where $T_f$ is larger (\eg, $T_f \textgreater 12$, more than 1 hour), capturing extended trends.
        \end{itemize}
        
        \item \underline{Imputation:} Given an incomplete spatio-temporal sample $X \in \mathbb{R}^{N \times T \times C}$ with missing values indicated by a binary mask $M \in \{0,1\}^{N \times T \times C}$, reconstruct the missing entries to obtain a complete dataset $\hat{Y}$. For each index $(i,t,c)$, we define $M_{i,t,c}=1$ if $X_{i,t,c}$ is observed and $M_{i,t,c}=0$ otherwise. This task can be divided into:
        \begin{itemize}[leftmargin=*,itemsep=0.3em,topsep=0.5em]
            \item[\scriptsize$\blacktriangleright$] Point Imputation: For each discrete coordinate $(i,t,c)$ satisfying $M_{i,t,c}=0$, the task is to estimate the missing value $\hat{Y}_{i,t,c}$.
            \item[\scriptsize$\blacktriangleright$] Block Imputation: For a contiguous set of indices $\mathcal{B} \subseteq \{(i,t,c)\}$ with $M_{i,t,c}=0$ for all $(i,t,c) \in \mathcal{B}$, the block imputation aims to jointly recover all missing contiguous sub-tensor $\hat{\mathbf{Y}}_{\mathcal{B}}$.
        \end{itemize}
    \end{itemize}

\noindent \textbf{Problem (\textit{Urban Spatio-temporal Foundation Models}).}
    Let the pre-training dataset be $\mathcal{D} = \{(X_i, Y_i)\}_{i=1}^{|\mathcal{D}|}$, where $X_i$ represents urban spatio-temporal data (either grid-based or sensor-based) and $Y_i$ denotes the corresponding target (future signals or imputed values). The goal is to train a foundational model $f_\theta$ that generalizes across different tasks $\mathcal{T}$ (\ie, forecasting and imputation) and exhibits:
    \begin{itemize}[leftmargin=2.5mm,parsep=2pt,topsep=0.5em]
        \item \underline{Zero-shot:} For an unseen dataset $\mathcal{D}'$, generate accurate predictions \(\hat{Y_i} = f_\theta(X'_i)\) in any tasks without any datasets-specific examples.
        \item \underline{Few-shot:} For an unseen dataset $\mathcal{D}'$, produce accurate predictions \(\hat{Y_i} = f_\theta(X'_i)\) in any tasks with only a few labeled examples.
    \end{itemize}
    
    This foundation model needs to capture the knowledge of urban spatio-temporal data during pre-training, thereby achieving region and task-agnostic generalization for zero- and few-shot inference.

\section{Methodology}

We investigate scaling mechanisms from three perspectives: data, computation, and architecture. Data serves as the “fuel” for spatio-temporal knowledge and  intelligence; computation defines the “tokenization” mechanism that transforms raw signals into learnable representations; architecture provides an “engine” with minimal inductive bias, designed to maximize scalability and transferability.

\subsection{Data Scaling}

\subsubsection{Intuition.} To effectively scale up urban spatio-temporal data, we need to consider two aspects: \textit{(\romannumeral1). Domain Diversity}: Urban data should cover multiple domains to represent data sources from different physical systems and application categories (\eg, traffic, human, bicycle mobility, \etc). \textit{(\romannumeral2). Spatio-temporal Diversity}: The temporal dimension should cover multi-year changes, while the spatial dimension should encompass geographical patterns of global cities. 

\begin{figure*}[t]
    \centering
    \includegraphics[width=1.0\textwidth]{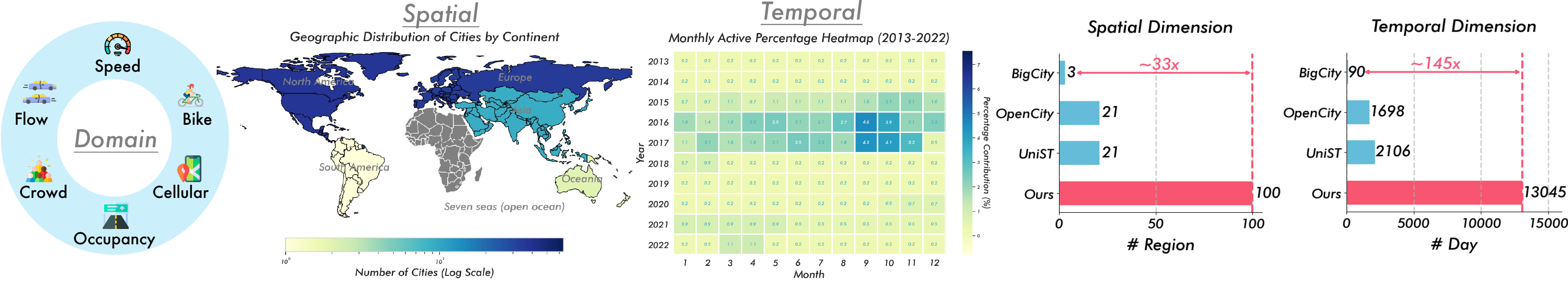}
    \caption{Data Overview: Breakthroughs in multi-domain coverage and spatio-temporal scale. Notably, we far surpass UniST~\citep{yuan2024unist}, OpenCity~\citep{li2024opencity}, and BigCity~\citep{yu2025bigcity} in terms of spatial regions and temporal spans, by as much as 33 to 145 times.}
    \label{fig:worldst_compare}
\end{figure*}

\subsubsection{Solution.} 

Constructing a unified corpus from fragmented urban data silos requires a rigorous data curation pipeline. We propose a three-step protocol to aggregate, clean, and standardize heterogeneous data sources, resulting in a high-quality \dataset~corpus, whose key characteristics are illustrated in Figure~\ref{fig:worldst_compare}.

\paragraph{Step 1: Multi-Source Acquisition \& Aggregation.}
To achieve the critical mass required for foundation models, we implemented a comprehensive data collection campaign targeting diverse urban areas. We aggregated raw data from two primary channels:

\begin{itemize}[leftmargin=2.5mm,itemsep=0.3em,topsep=0.5em]
    \item \textit{Open Government Portals:} We crawled archival data from municipal open data platforms (\eg, NYC Open Data\footnote{\url{https://opendata.cityofnewyork.us/}}, TfL Open Data\footnote{\url{https://tfl.gov.uk/info-for/open-data-users/our-open-data}}, \etc). This provided massive-scale raw logs of public transit swipes, taxi trajectories, and bike-sharing records.
    \item \textit{Established Academic Repositories:} We have also integrated some spatio-temporal datasets that have been open-sourced by academia (such as UCTB~\footnote{\url{https://uctb.github.io/UCTB/index.html}}, UTD-19~\footnote{\url{https://utd19.ethz.ch/}}, \etc) to further enrich data diversity.
    \item \textit{Domain-Specific APIs:} We accessed domain-specific APIs~\footnote{\url{https://pems.dot.ca.gov/}} to retrieve fine-grained sensor readings, such as loop detector data for highway traffic speeds and occupancy rates.
\end{itemize}

Through this campaign, we curated a massive repository spanning 8 distinct domains (including Traffic Speed, Flow, Crow, \etc) across 100 cities globally. This raw collection encompasses over 1 billion data points, characterized by extreme heterogeneity in file formats (\eg, CSV, HDF5, NPZ, \etc) and metadata standards.

\paragraph{\textit{Step 2: Unified Ingestion \& Alignment.}}
The collected raw data exhibits significant inconsistency in storage formats and sampling frequencies. We developed a unified ingestion engine to standardize:

\begin{itemize}[leftmargin=2.5mm,itemsep=0.3em,topsep=0.5em]
    \item \textit{Format Unification:} We built adaptors to parse the diverse raw formats, mapping them into a standardized tensor $\mathbf{X}_{\text{raw}} \in \mathbb{R}^{N \times T \times C}$.
    \item \textit{Frequency Synchronization:} To address temporal heterogeneity, we standardized the time resolution to a base frequency $\Delta t$ (\ie, 5 minutes). We apply \textit{downsampling} via aggregation (sum/mean) for high-frequency streams and \textit{linear interpolation} for low-frequency streams, ensuring temporal alignment across all cities.
\end{itemize}

\paragraph{\textit{Step 3: Quality Control \& Pre-completion.}}
Real-world sensors are prone to hardware failures (leading to missing values) and transmission errors (leading to noise). We implement a strict filtering and pre-completion mechanism to ensure data density:

\begin{itemize}[leftmargin=2.5mm,parsep=2.5pt,topsep=5pt]
    \item \textit{Static Node Removal:} We calculate the variance $\sigma^2_n$ for each spatial node $n$. Nodes with $\sigma^2_n \approx 0$ (indicating dead sensors or constant values) are discarded to prevent learning trivial mappings.
    \item \textit{Outlier Clipping:} To mitigate the impact of extreme sensor noise, we apply the $3\sigma$-rule. Values falling outside $[\mu - 3\sigma, \mu + 3\sigma]$ are treated as anomalies and replaced by boundary values.
    \item \textit{Missing Value Pre-completion:} We ensure dense input tensor by pre-filling missing values. For small gaps, we use linear interpolation to reconstruct signal continuity. This ensures input tensors $\mathbf{X}$ are complete and continuous, stabilizing learning.
\end{itemize}

\subsubsection{Discussion.} 
\textit{Why this pipeline?} 
Our pipeline prioritizes signal integrity and completeness. Simple concatenation of raw city datasets often introduces "poisonous" dead nodes or extensive gaps that destabilize foundation model training. By aggregating from diverse channels and enforcing strict quality control with temporal pre-completion, we transform fragmented urban logs into a consistent, dense "fuel" optimized for large-scale pre-training. Detailed information and data visualizations are provided in Appendix \ref{appendix_data}.

\subsection{Computation Scaling}

\subsubsection{Intuition.}
Standard deep learning models necessitate fixed input dimensions for efficient parallelization; however, urban data exhibits inherent structural elasticity, with varying sensor counts  and temporal spans  across cities. To construct a foundation model, we require a generalized computational unit—analogous to ``tokens'' in NLP or ``patches'' in Vision Transformers—that decouples input size from model architecture while preserving the learnability of local spatio-temporal correlations. Specifically, urban systems manifest distinct correlation patterns: \textit{(\romannumeral1). Temporal Order}: The natural sequential dependency (past influences future) must be strictly preserved. \textit{(\romannumeral2). Spatial Locality}: Spatial correlations are predominantly local but lack a natural 1D sequence. Consequently, the core challenge lies in: \textit{How to efficiently partition continuous spatial fields into discrete, capacity-constrained units that maintain local proximity while enabling the model to learn spatial order as a sequence?}

\begin{figure*}[t!]
    \centering
    \includegraphics[width=\textwidth]{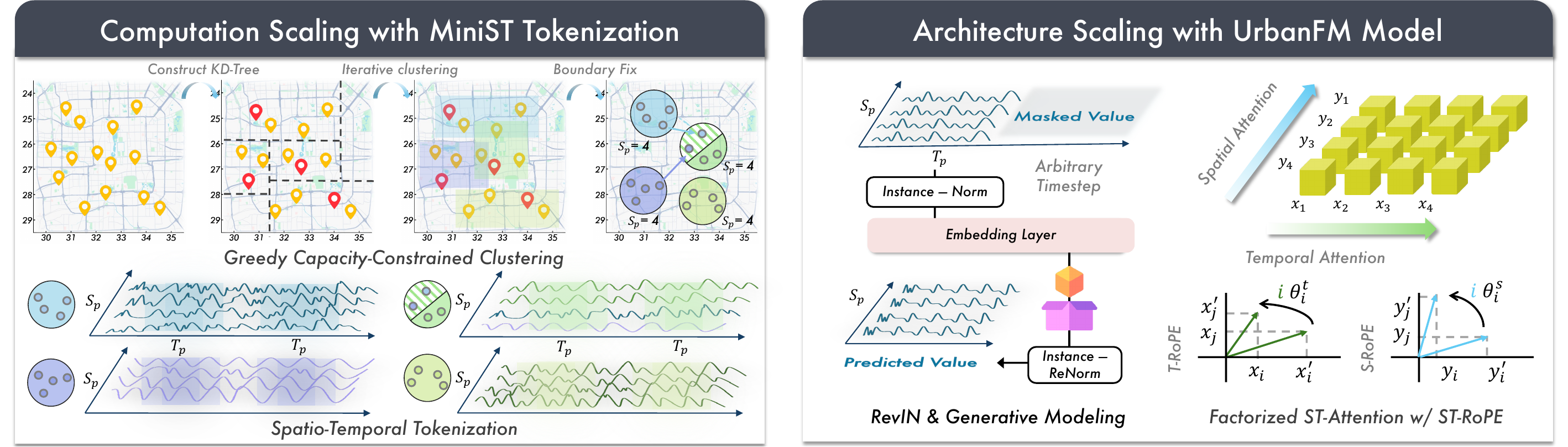}
    \caption{A schematic diagram of the scaling mechanism: \format~tokenization and \model~model.}
    \label{fig:minist_urbanfm}
\end{figure*}

\subsubsection{Solution.}
As shown in Figure~\ref{fig:minist_urbanfm} left, we introduce the \format

\noindent tokenization strategy, utilizing a dynamic split mechanism to transform heterogeneous urban data into unified input samples.

\begin{itemize}[leftmargin=2.5mm,itemsep=0.3em,topsep=0.5em]
\item \textit{Greedy Capacity-Constrained Clustering:} Addressing the challenge of scalable topology modeling for varying $N$, we propose a spatially-aware grouping strategy inspired by KD-Tree logic~\citep{samet2006foundations}, which partitions $N$ spatial nodes into fixed-size clusters of size $S_p$ to linearize spatial structures while preserving spatial relations.
    \begin{itemize}[leftmargin=*,itemsep=0.3em,topsep=0.5em]
    \item[\scriptsize$\blacktriangleright$] We first construct a KD-Tree using geographic coordinates (latitude, longitude) of all sensors or grid cells (idx, idy).
    \item[\scriptsize$\blacktriangleright$] We then iteratively query the tree to group the $S_p$ nearest unassigned neighbors into a cluster. This ensures nodes within a token are geographically proximal, preserving local spatial correlations within the embedding itself.
    \item[\scriptsize$\blacktriangleright$] For clusters with fewer than $S_p$ nodes (boundary cases), we apply binary masking to maintain tensor uniformity.
    \end{itemize}
\item \textit{Spatio-Temporal Patching:} We define a spatio-temporal patch $X_{k,t}$ of dimension $S_p \times T_p$, where $S_p$ is the derived spatial cluster and $T_p$ is the temporal window, effectively transforming irregular spatial structures into regular patch samples.
\end{itemize}

\subsubsection{Discussion.}
\textit{Why Greedy Clustering over Convolution Operator?}
Traditional convolution or graph operator require adjacency matrices $N_A$ , complicating batch processing across heterogeneous cities (where $N_A\neq N_B$). In contrast, our approach renders tokens structurally independent, enabling massive parallel training for efficient scaling. Furthermore, by enforcing local aggregation via KD-Tree, we explicitly encode the geoscience principle that ``near things are more related'' into tokenization. The attention mechanism subsequently learns global correlations among these local samples, effectively modeling the hierarchy of urban dynamics. Detailed pseudocode and visualizations of the greedy capacity-constrained clustering algorithm are provided in Appendix~\ref{appendix_method}.

\subsection{Architecture Scaling}

\subsubsection{Intuition}
Existing spatio-temporal models~\citep{wu2019graph,ma2025less}, even including current foundation models~\citep{li2024opencity,yuan2025diffusion,zhong2025st}, rely heavily on the injection of handcrafted priors, such as auxiliary graph structures or node prompts. While these priors yield benefits on limited datasets, the strong inductive biases they introduce inevitably become bottlenecks for scaling. We posit that the crux of urban modeling lies in capturing dynamic correlations. Consequently, our objective is to construct an architecture with minimal inductive bias, relying exclusively on the attention mechanism to learn evolving dependencies from massive data, rather than hard-coding human heuristics.

\subsubsection{Solution}
As shown in Figure~\ref{fig:minist_urbanfm} right, we design \model~by streamlining standard transformer~\citep{vaswani2017attention} components, only optimizing them specifically for spatio-temporal samples.

\begin{itemize}[leftmargin=2.5mm,itemsep=0.3em,topsep=0.5em]
    \item \textit{Factorized Spatio-Temporal Attention:} To preserve fine-grained dynamic relationships at the point level without incurring the quadratic complexity of standard self-attention on flattened sequences, we factorize the computation into two orthogonal phases:
        \begin{itemize}[leftmargin=*,itemsep=0.3em,topsep=0.5em]
            \item[\scriptsize$\blacktriangleright$] \textit{Temporal Attention:} we first use the self-attention layer to model temporal dependencies within each spatial node. 
            \item[\scriptsize$\blacktriangleright$] \textit{Spatial Attention:} we then use the self-attention layer to model spatial dependencies between nodes at each time step.    
        \end{itemize}        

    \item \textit{Spatio-Temporal RoPE:} Addressing the inherent permutation invariance of attention, we use RoPE~\citep{su2024roformer} to encode relative positions by rotating query and key vectors in the embedding space:
        \begin{itemize}[leftmargin=*,itemsep=0.3em,topsep=0.5em]
            \item[\scriptsize$\blacktriangleright$] \textit{T-RoPE:} Encodes relative temporal distances, enabling the model to generalize to varying historical context lengths.
            \item[\scriptsize$\blacktriangleright$] \textit{S-RoPE:} Encodes relative spatial order within linearized samples, preserving local proximity without adjacency matrix.
        \end{itemize}

    \item \textit{RevIN \& Generative Modeling:} To mitigate the non-stationary statistics of urban flows, we apply instance normalization before and after the backbone. We adopt modern generative modeling objective where future time steps are masked with zero noise values, compelling the model to reconstruct future signals from noisy contexts, thereby unifying forecasting and imputation.
\end{itemize}

\subsubsection{Discussion}
\textit{Rationale for Minimalist Design.} Transformers have demonstrated clear scaling laws~\citep{kaplan2020scaling}, where increasing depth and data volume consistently reduces loss. Therefore, we eschew superfluous structural priors, relying solely on self-attention to capture dynamics. Furthermore, this simplified design allows our model to leverage modern hardware optimizations and architectural iterations~\citep{sun2025speed}, such as Flash Attention~\citep{dao2022flashattention} and Linear Attention~\citep{katharopoulos2020transformers}, facilitating massive scaling. Additionally, our pre-training objective aligns with modern generative prediction paradigms, enabling the model to flexibly support arbitrary-length forecasting, a significant advantage over the fixed-length constraints of traditional models.

\noindent \underline{Ethics, Fairness, and Limitations} are discussed in Appendix~\ref{disscuison}.

\section{Experiments}

In this section, we conduct extensive experiments to investigate the following research questions:

\begin{itemize}[leftmargin=2.5mm,parsep=6pt,topsep=4pt] 
    \item \textbf{RQ1:} Can \model~match or surpass existing models in zero-shot and few-shot evaluations across various urban spatio-temporal analysis tasks? (\textbf{Effectiveness \& Generality})  
    \item \textbf{RQ2:} Can the pre-training phase of \model~benefit from the scaling of datasets and model capability? (\textbf{Scaling Property})  
    \item \textbf{RQ3:} Can the inference phase of \model~robustly tolerate additional noise and meet the requirements of efficient inference in practical scenarios? (\textbf{Robustness \& Efficiency})  
\end{itemize}

\subsection{Experimental Setup}

\noindent\textbf{Benchmark Datasets and Baseline Settings.}
We constructed the largest and most comprehensive urban spatio-temporal benchmark to date, \benchmark, comprising up to 12 datasets, for downstream task evaluation. \benchmark encompasses real-world data in diverse formats (sensor-based and grid-based), spanning various domains such as traffic flow, speed, occupancy, road networks, bicycles, taxis, and trajectories, across 4 countries and 7 cities, and a temporal range exceeding 10 years. To facilitate comprehensive evaluation, we selected 22 diverse baseline models, including specialized expert models and foundation models. For forecasting tasks, the expert models consist of spatio-temporal graph models specifically designed for sensor-based data (STAEformer~\citep{liu2023spatio}, STID~\citep{shao2022spatial}, D2STGNN~\citep{shao2022decoupled}, GWNET~\citep{wu2019graph}) and spatio-temporal grid models tailored for grid-based data (TAU~\citep{tan2023temporal}, PredRNN++~\citep{wang2022predrnn}, DSAN~\citep{lin2020preserving}, ST-ResNet~\citep{zhang2017deep}). Notably, spatio-temporal grid models cannot be applied to sensor-based datasets, and adjacency matrices for spatio-temporal graph models were pre-constructed based on the first-order adjacency of grids for application to grid-based datasets. The foundation models include time series foundation models (TimesFM++~\citep{das2024timesfm}, Moirai~\citep{woo2024moirai}, Time-MoE~\citep{shi2025time}, Chronos~\citep{ansari2024chronos}) and spatio-temporal foundation models (FactoST~\citep{zhong2026st,zhong2025st}, OpenCity~\citep{li2024opencity}). Notably, for open-source foundation models, we evaluate their zero-shot performance using publicly available weights, reporting the optimal results. For closed-source counterparts, we retrain them following the original literature to ensure a fair and unified comparison. 
For imputation tasks, we choose classical method (Mean, \textit{KNN}~\citep{crookston2008yaimpute}, \textit{MICE}~\citep{van2011mice}, and \textit{SVDImpute}~\citep{xu2017interpolating}) as default baselines. More details about the baselines and operations are provided in Appendix~\ref{appendix_baseline} and~\ref{appendix_predefined}.

\noindent\textbf{Evaluation Protocols and Settings.}
Spatio-temporal data in each dataset are chronologically partitioned into training, validation, and testing sets (6:2:2) with early stopping applied. We evaluate three configurations: \textit{full-shot} (standard training), \textit{few-shot} (training / fine-tuning on 10\% of train set), and \textit{zero-shot} (direct inference for foundation models). Following standard protocols~\citep{shao2024exploring,cini2022filling}, the look-back and horizon windows are set to 12 steps for \textit{short-term} and 24 steps for \textit{long-term forecasting}. For imputation tasks, we simulate \textit{block missing} (5\% sensor drop rate per step) and \textit{point missing} (25\% random masking). Performance is measured via MAE, RMSE, and MAPE. Regarding hyper-parameters, patch sizes $S_p$ and $T_p$ for \format~tokenization are set to 16 and 48, while \model~defaults to 8 layers (4 spatial, 4 temporal) with 10 pre-training epochs. Further evaluation and parameter details are provided in Appendix~\ref{appendix_protocol}.

\subsection{Performance Evaluation (RQ1)}

To evaluate the effectiveness of \model, we conducted a comprehensive comparison with 22 baselines, and we present the average MAPE across each dataset as a comprehensive indicator analysis.

\subsubsection{Zero-Shot Generalization}
Figure~\ref{fig:rq1_figure2} illustrates the zero-shot performance of \model~ across both sensor-based and grid-based benchmarks for short-term and long-term forecasting. We observe several key findings: \ding{182} \textit{Significant Superiority over Existing Foundation Models}: \model~ consistently achieves the lowest MAPE across all scenarios. In both sensor- and grid-based zero-shot benchmarks, \model~ substantially outperforms existing spatio-temporal foundation models, with performance gains ranging from 39.0\% to 70.2\%. This suggests that the large-scale pre-training paradigm of \model~ effectively captures universal laws governing urban spatio-temporal patterns. \ding{183} \textit{Outperforming Domain-Specific Experts}: Notably, even in a zero-shot setting, \model~'s performance is comparable to, or even surpasses, expert models trained on the full target dataset (full-shot). For instance, in long-term sensor-based forecasting, \model~ (17.0) outperforms highly specialized models such as D2STGNN (20.5) and STID (19.9). \ding{184} \textit{Robustness Across Data Structures and Horizons}: \model~ maintains a decisive lead in both short- and long-term horizons across grid and sensor data. This demonstrates its architectural flexibility in handling heterogeneous urban data representations and its long-term predictive stability.

\subsubsection{Adaptability and Few-Shot Performance}
\begin{figure*}[t!]
    \centering
    \includegraphics[width=\textwidth]{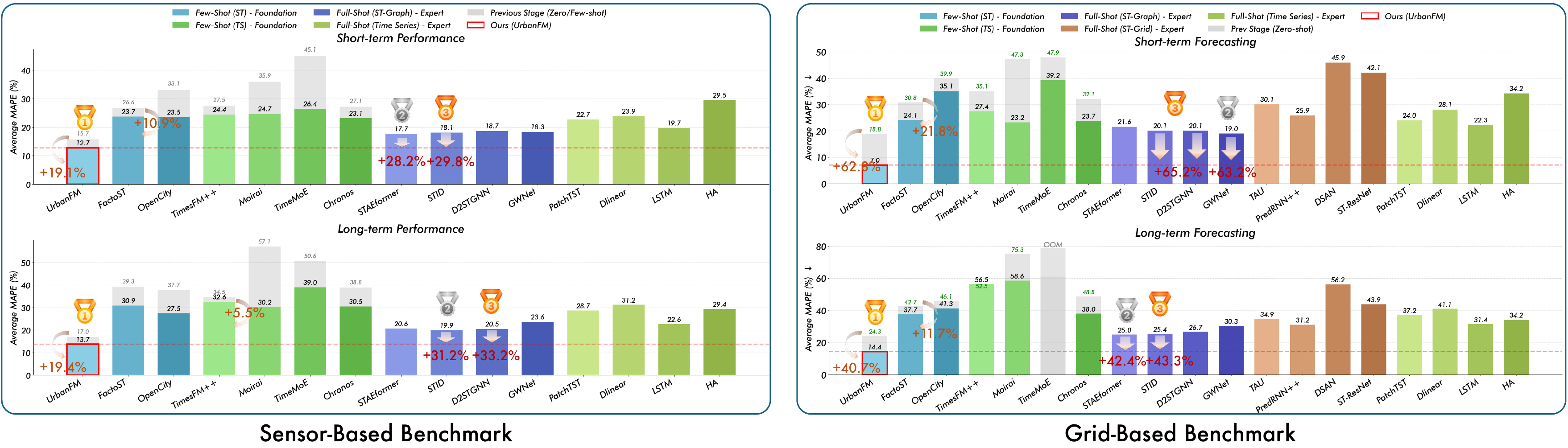}
    \vspace{-2mm}
    \caption{Evaluation of \model's performance gain through few-shot tuning and comparison with full-shot expert models.}
    \vspace{-2mm}
    \label{fig:rq1_figure2}
\end{figure*}

Figure~\ref{fig:rq1_figure2} depicts the performance gains of \model~when adapting to target domains via few-shot fine-tuning. We observe that: \ding{182} \textit{Substantial Performance Gains}: \model~ exhibits remarkable "rapid adaptation" capabilities. With fine-tuning on only a small fraction of target samples, \model~’s performance further exceeds that of full-shot expert models, with relative improvements of 28.2\%--65.2\%. \ding{183} \textit{Efficient Adaptation}: Compared to other foundation models, \model~ achieves superior performance increments. Even while maintaining a lower baseline error, our model yields relative improvements of 19.1\%--62.8\% over competing foundation models. While other models show some gains, their final error rates remain significantly higher than \model~’s few-shot results. \ding{184} \textit{Adaptability to Sparse Data Domains}: The few-shot performance boost is particularly pronounced in grid-based tasks. We attribute this to the fact that while the pre-training set contained fewer grid-style datasets, our data scaling strategy allows the model to quickly align with the target distribution of sparse domains through minimal supervision.

\begin{figure*}[t!]
    \centering
    \includegraphics[width=\textwidth]{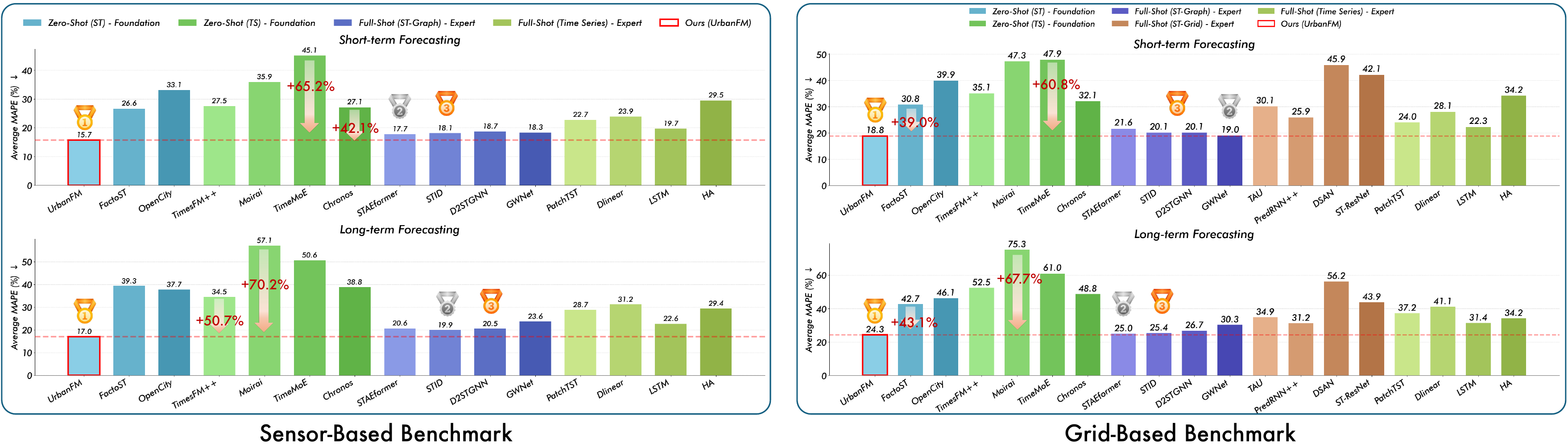}
    \vspace{-2mm}
    \caption{Zero-shot forecasting effectiveness on various spatial-temporal benchmarks (Full results in Table~\ref{tab:zero_few_grid},~\ref{tab:zero_few_graph},~\ref{tab:few_full_graph}, and~\ref{tab:few_full_grid}).}
    \vspace{-2mm}
    \label{fig:rq1_figure1}
\end{figure*}

\begin{figure*}[t!]
    \centering
    \includegraphics[width=\textwidth]{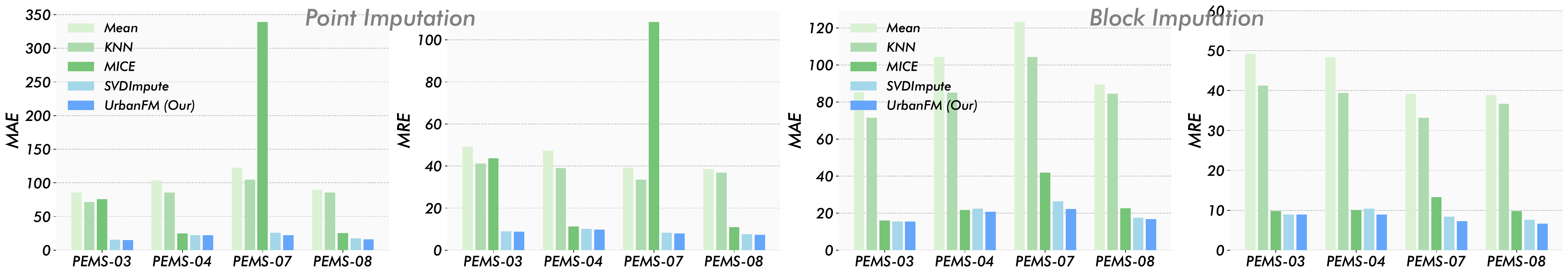}
    \vspace{-2mm}
    \caption{Performance evaluation of \model~and classic methods for spatio-temporal point and block data imputation tasks.}
    \vspace{-2mm}
    \label{fig:rq1_figure3}
\end{figure*}

\subsubsection{Cross-Task Generalization}
Figure~\ref{fig:rq1_figure3} presents the performance of \model~on spatio-temporal data imputation tasks across four PEMS benchmarks. We evaluate the model under two challenging scenarios: \textit{point imputation} (random missing values) and \textit{block imputation} (continuous missing time intervals). \ding{182} \textit{Exceptional Zero-Shot Transferability:} Despite the absence of imputation learning objectives during the pre-training phase, \model~consistently achieves the lowest error rates across all PEMS datasets. We attribute this success to our generative modeling strategy, which enables the model to leverage universal spatio-temporal representations learned from forecasting tasks effectively. \ding{183} \textit{Robustness to Complex Missing Patterns:} In the highly challenging \textit{block imputation} scenario, \model~leverages learned spatio-temporal structural priors to accurately reconstruct long-range missing segments. Consequently, it significantly outperforms traditional baselines such as MICE and SVD. \ding{184} \textit{High Stability Across Datasets:} In contrast to baselines like MICE, which exhibit high variance (\eg, instability on PEMS-07), our model maintains consistently low error bounds across diverse traffic distributions. This validates the robustness of \model~as a reliable backbone for urban data analysis.

\subsection{Scaling Evaluation (RQ2)}

\subsubsection{Scaling Property of Model Side}
Model depth analysis (Figure~\ref{fig:rq2_figure1}, row 1) shows that on all PEMS datasets, increasing the number of Attention layers from 2 to 8 consistently reduces both short-term and long-term prediction errors. This performance improvement stems from the enhanced ability of deeper architectures to capture spatiotemporal hierarchical dependencies. For larger parameters, we consider this a future engineering optimization challenge.

\subsubsection{Scaling Property of Data Side}
\begin{figure*}[t!]
    \centering
    \begin{minipage}{1.0\textwidth}
        \centering
        \includegraphics[width=\linewidth]{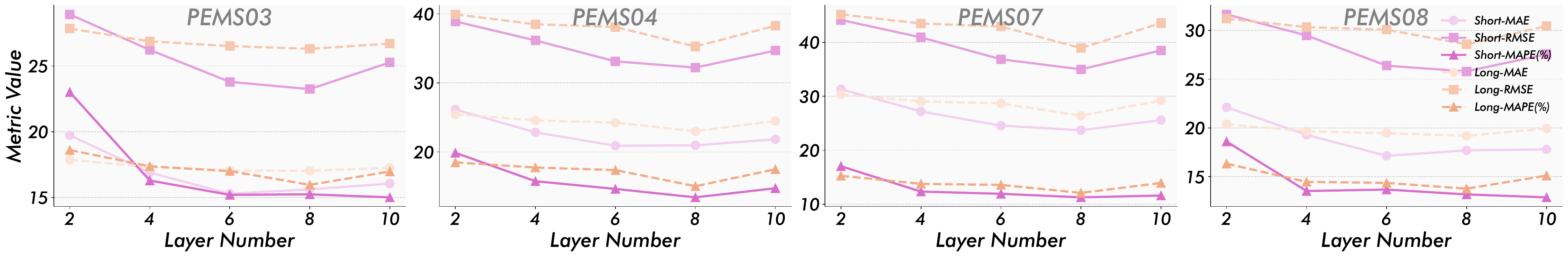}
    \end{minipage}
    \begin{minipage}{1.0\textwidth}
        \centering
        \includegraphics[width=\linewidth]{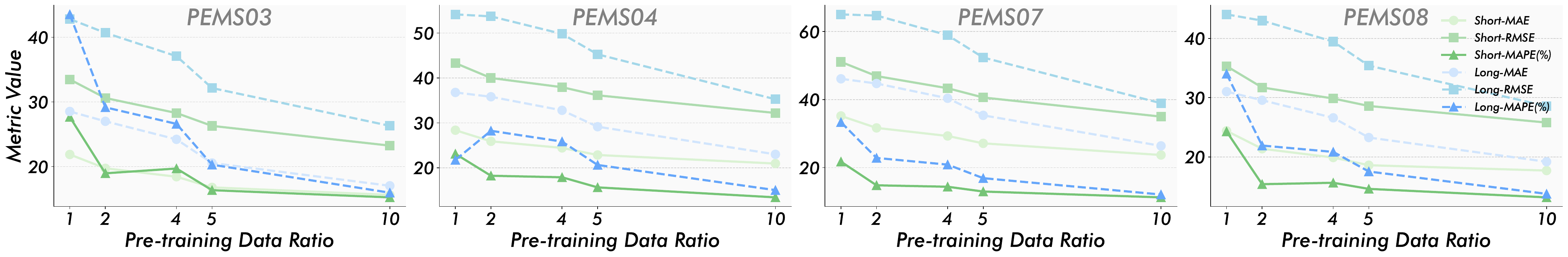}
    \end{minipage}
    \caption{Scaling law analysis across model depth, data volume. (Full results in Table~\ref{tab:ratio} and~\ref{tab:layer}).}
    \label{fig:rq2_figure1}
\end{figure*}

The effect of expanding the pre-training data volume (Figure~\ref{fig:rq2_figure1}, row 2) shows that MAE, RMSE, and MAPE all exhibit power-law decay as the data proportion gradually increases from 1\%. No performance saturation was observed at the maximum test proportion, confirming that \model~ can effectively extract generalized urban dynamic features from large-scale corpora. This indicates that further data expansion will continue to enhance its zero-shot generalization ability.

\begin{figure*}[t!]
    \centering
    \begin{minipage}{1.0\textwidth}
        \centering
        \includegraphics[width=\linewidth]{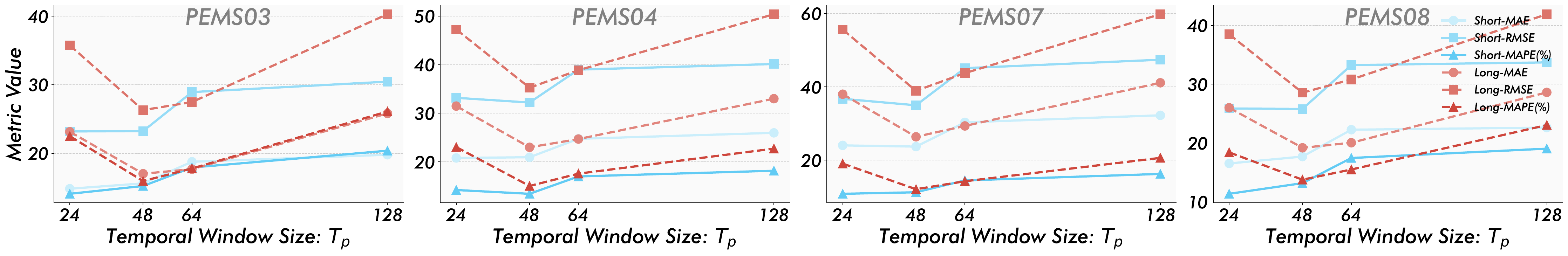}
    \end{minipage}
    \begin{minipage}{1.0\textwidth}
        \centering
        \includegraphics[width=\linewidth]{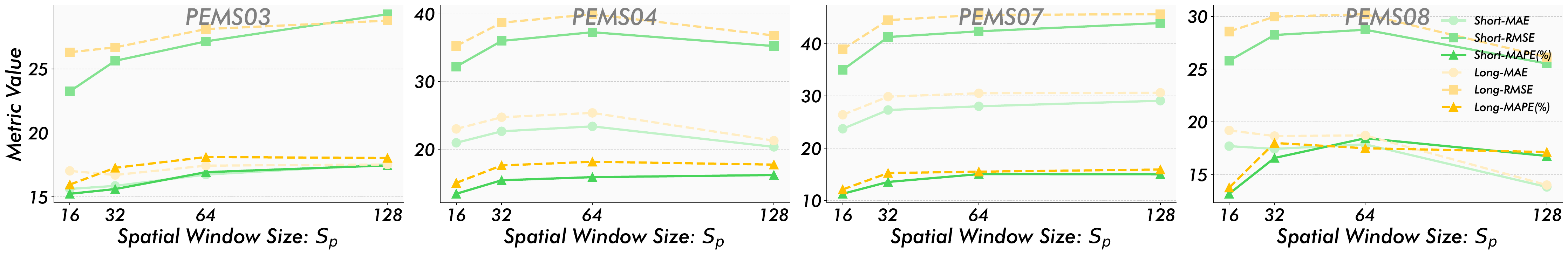}
    \end{minipage}
    \caption{Scaling law analysis of spatio-temporal resolutions (Full results in Table~\ref{tab:temporal} and~\ref{tab:spatial}).}
    \label{fig:rq2_figure2}
\end{figure*}

\subsubsection{Scaling Property of Sample Side} The evaluation of the spatiotemporal window size (Figure~\ref{fig:rq2_figure2}) shows that model performance is highly sensitive to the range of the input context. Although increasing the temporal window ($T_p$) and spatial window ($S_p$) can initially improve accuracy by expanding the receptive field, performance usually tends to stabilize or slightly decrease after exceeding a certain threshold (\eg, 64), reflecting the inherent locality of urban patterns, where excessive context introduces weak noise.

\subsection{Deployment Evaluation (RQ3)}

\begin{figure*}[t!]
    \centering
    \includegraphics[width=1.0\textwidth]{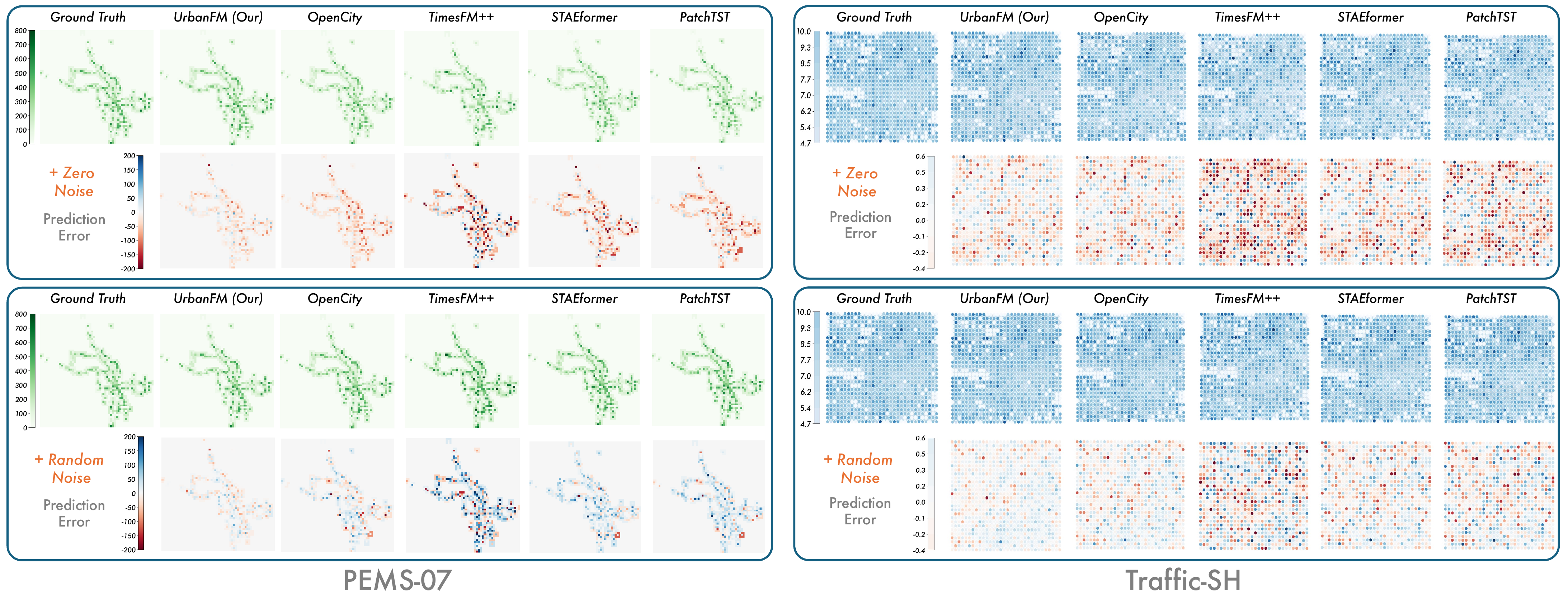}
    \vspace{-4mm}
    \caption{A visualization study of the robustness of predictions in noisy environments using two datasets.}
    \label{fig:robustness}
    \vspace{-4mm}
\end{figure*}

\subsubsection{Robustness Study}
To evaluate \model's resilience against real-world sensor outages and transmission errors, we conducted a case study of stress tests using 30\% zero-masking and 30\% Gaussian noise injection. Observations from Figure~\ref{fig:robustness} reveal three key phenomena: \ding{182} \textit{ST-FM Superiority:} Unlike specialized experts (e.g., STAEformer) that suffer significant degradation due to distribution shifts, our \model~and OpenCity demonstrates superior robustness, avoiding the high-error regions typical of overfitted models. \ding{183} \textit{Role of Spatial Coupling:} While time-series foundation models struggle with spatial consistency, \model~leverages spatio-temporal correlations to rectify local anomalies, resulting in significantly smoother error surfaces. \ding{184} \textit{Intrinsic Denoising of \model:} \model~maintains the highest fidelity across all scenarios. This resilience is attributed to our generative modeling pre-training, which treats signal recovery as a core objective, effectively enabling the model to treat data corruption as a routine inference task.

\subsubsection{Efficiency Study}

\begin{wrapfigure}{r}{8cm}
\begin{center}
\vspace{-11mm}
\includegraphics[width=1.0\linewidth]{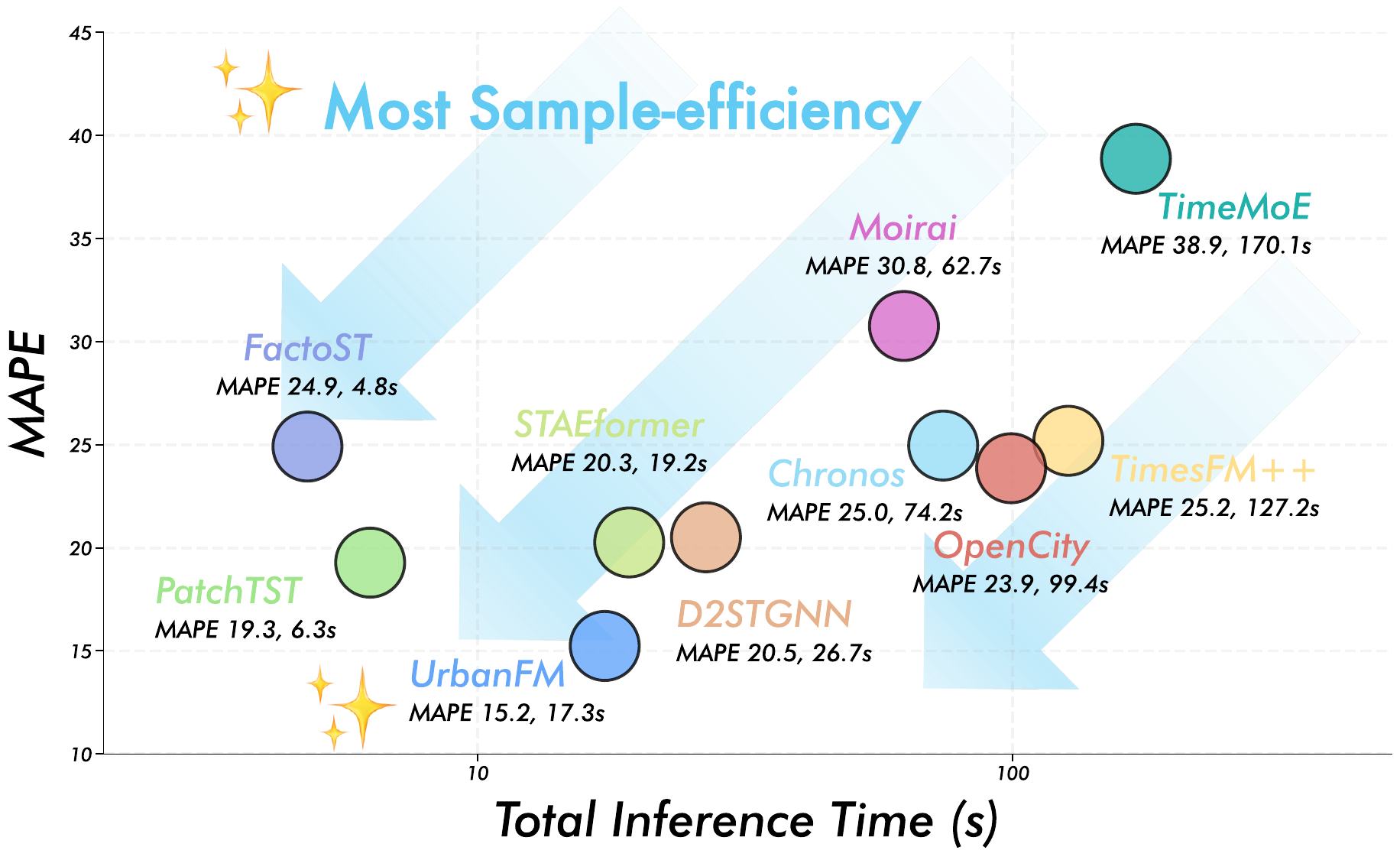}
\setlength{\abovecaptionskip}{-0.25cm}
\caption{Efficiency study of our method.} 
\vspace{-2mm}
\label{fig.efficiency}
\end{center}
\end{wrapfigure} 
We evaluated the total inference time and predictive performance (MAPE) of \model~ against other state-of-the-art time series and foundation models. 
To ensure a fair comparison, we report the total inference time required to process the test set on the same hardware environment (One A100 GPU, 64 batch size, PEMS-03 short). 
As illustrated in Figure~\ref{fig.efficiency}, \model~ strikes a favorable balance between forecasting accuracy and computational efficiency. 
Specifically: \ding{182} Compared to lightweight time series models (\eg, PatchTST), although they achieve marginally lower latency, their simplified designs lead to inferior forecasting performance (higher MAPE). 
\ding{183} Compared to large-scale foundation models (\eg, Chronos, TimeMoE, TimesFM++), our approach achieves superior performance (the lowest MAPE of 15.2\%) while requiring orders of magnitude less inference time (\eg, roughly $4\times$ faster than Chronos and $10\times$ faster than TimeMoE), demonstrating the practical efficiency of our architecture in real-world deployments.

\section{Conclusion}

In this work, we present a systematic investigation into urban spatio-temporal foundation models, aiming to transcend the fragmentation of existing scenario-specific approaches. Guided by the first principles of urban science (\ie, heterogeneity, correlation, and dynamics), we propose a unified scaling framework comprising \dataset, \format, \model, and \benchmark~to address \textit{what} and \textit{how} to scale. Extensive empirical results demonstrate that \model~achieves remarkable zero-shot generalization capabilities previously unattainable by specialized baselines. We believe this work marks a pivotal paradigm shift, providing a blueprint for large-scale urban spatio-temporal intelligence research and advancing our understanding of complex city dynamics.


\clearpage

\definecolor{textgray}{HTML}{6E6E73}
\makeatletter
\newcommand\applefootnote[1]{%
  \begingroup
  \renewcommand\thefootnote{}%
  \renewcommand\@makefntext[1]{\noindent##1}%
  \footnote{#1}%
  \addtocounter{footnote}{-1}%
  \endgroup
}
\makeatother

\bibliography{ref}
\bibliographystyle{plainnat}
\newpage
\appendix

\section*{Appendix}
\appendix
\counterwithin{figure}{section}
\counterwithin{table}{section}
\fancypagestyle{appendixfooter}{
  \fancyhf{} 
  \renewcommand{\headrulewidth}{0pt}
  \renewcommand{\footrulewidth}{0pt}
  \fancyfoot[L]{\hyperlink{appendix-start}{{\textit{Go to Appendix Index}}}}
  \fancyfoot[C]{\thepage}
}

\hypertarget{appendix-start}{}
\pagestyle{appendixfooter}

\vspace{1.5em}

\hrule height .8pt
\DoToC
\hrule height .8pt

\vspace{1.5em}

\clearpage

\section{Datasets Details}\label{appendix_data}

\subsection{Evaluation Benchmark: \benchmark}

\begin{figure*}[htbp!]
    \centering
    \begin{minipage}[b]{0.20\linewidth}
        \includegraphics[width=\linewidth]{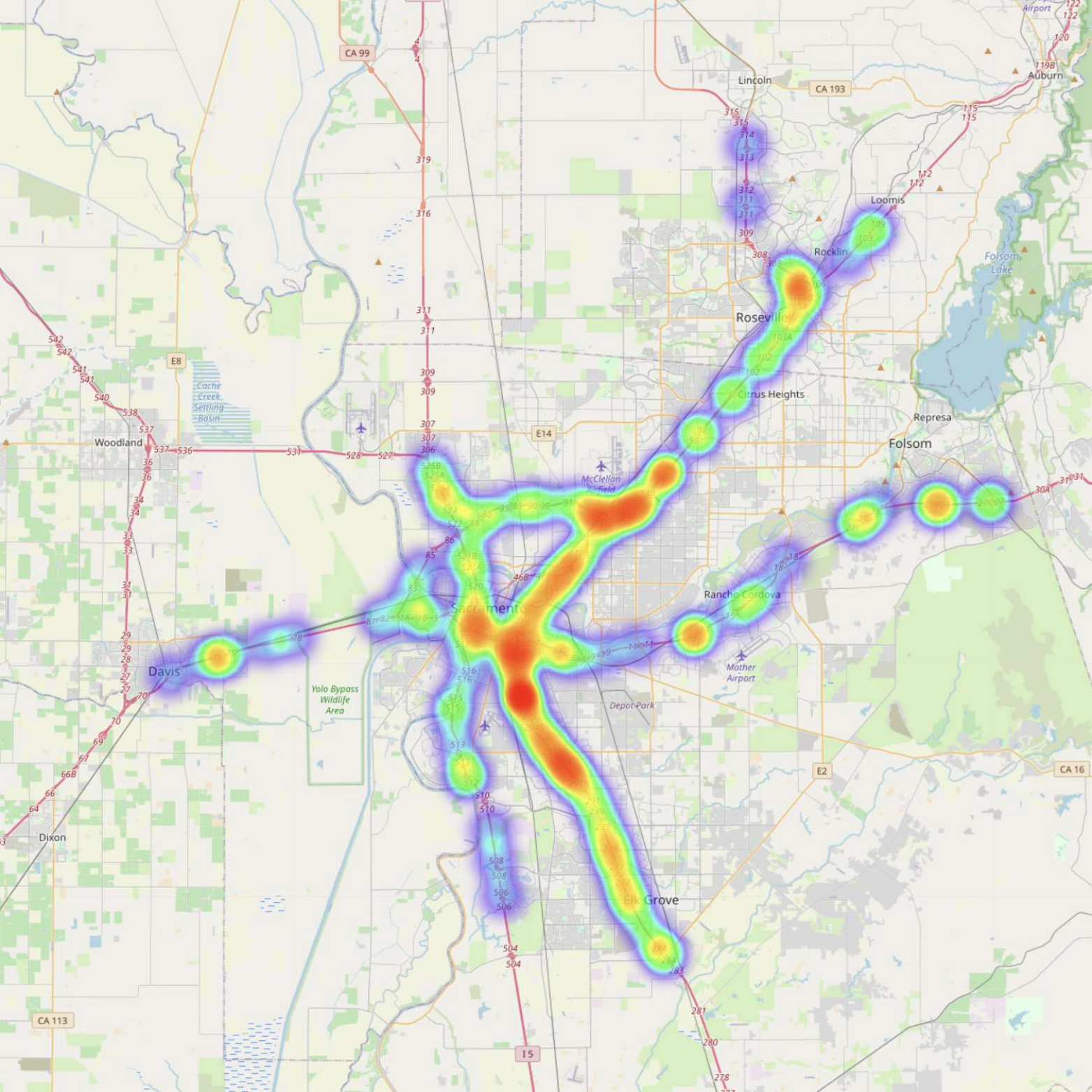}
        \centerline{\footnotesize PEMS-03}
    \end{minipage}
    \hspace{6mm} 
    \begin{minipage}[b]{0.20\linewidth}
        \includegraphics[width=\linewidth]{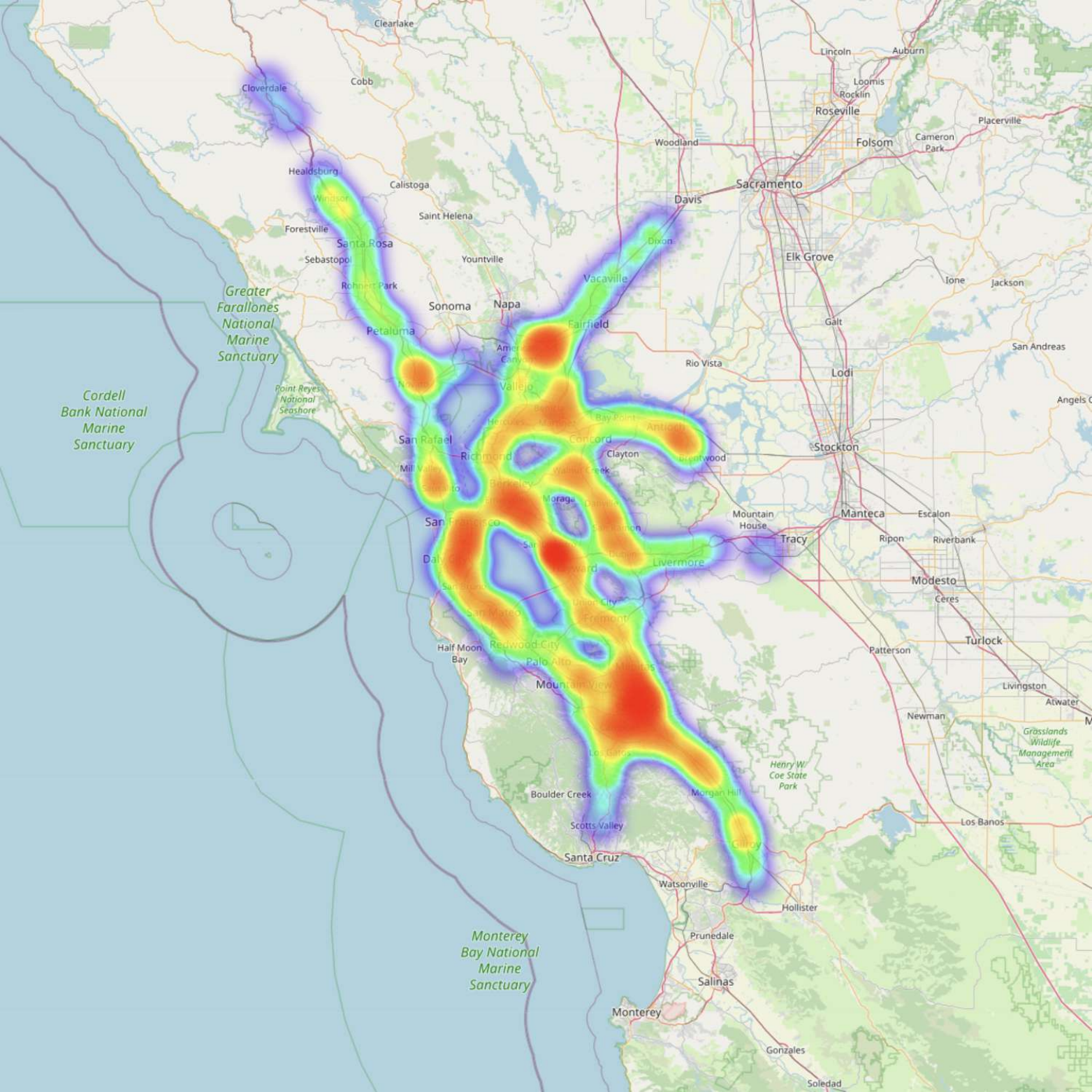}
        \centerline{\footnotesize PEMS-04}
    \end{minipage}
    \hspace{6mm}
    \begin{minipage}[b]{0.20\linewidth}
        \includegraphics[width=\linewidth]{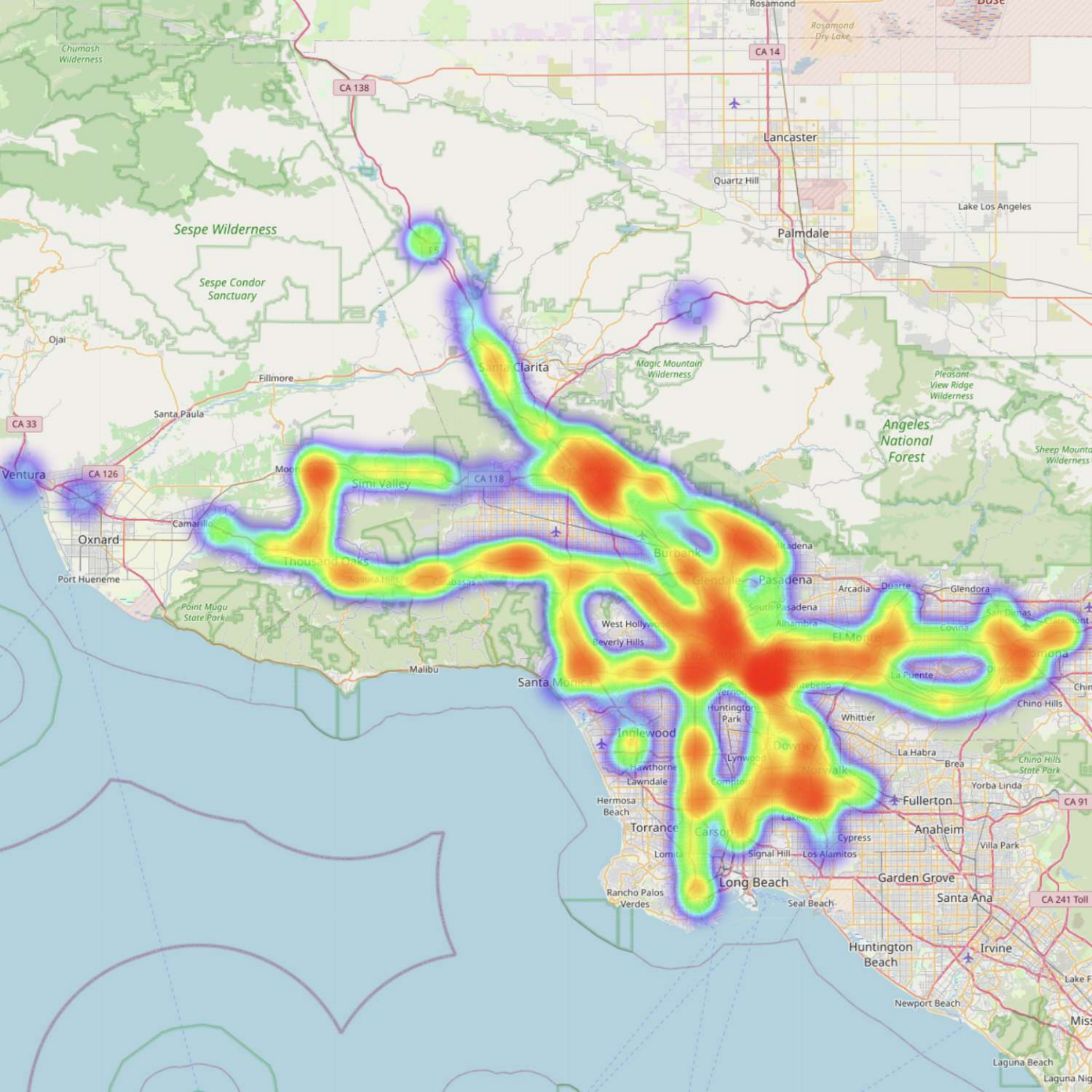}
        \centerline{\footnotesize PEMS-07}
    \end{minipage}
    \hspace{6mm}
    \begin{minipage}[b]{0.20\linewidth}
        \includegraphics[width=\linewidth]{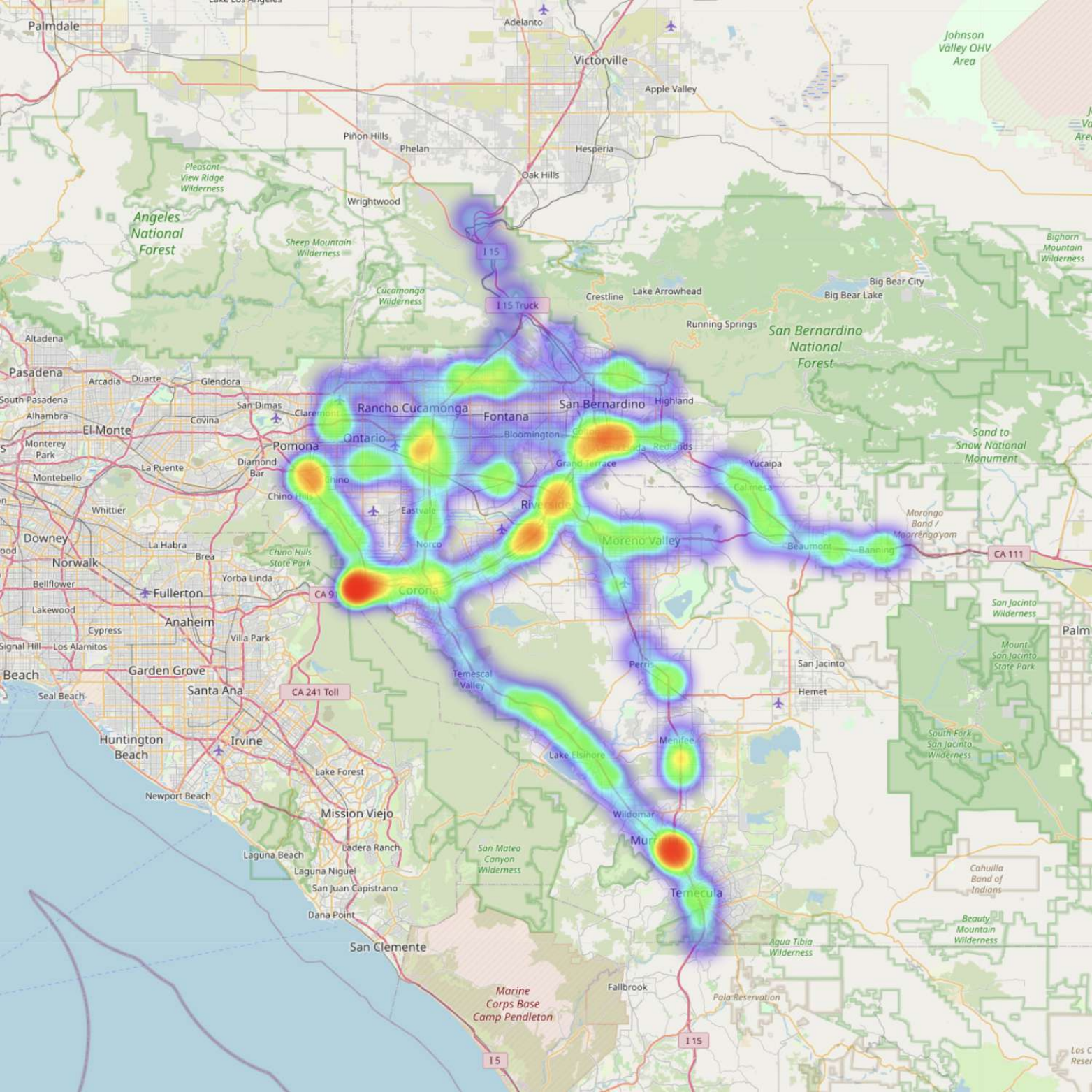}
        \centerline{\footnotesize PEMS-08}
    \end{minipage}\\
    
    \vspace{4mm} 
    
    \begin{minipage}[b]{0.20\linewidth}
        \includegraphics[width=\linewidth]{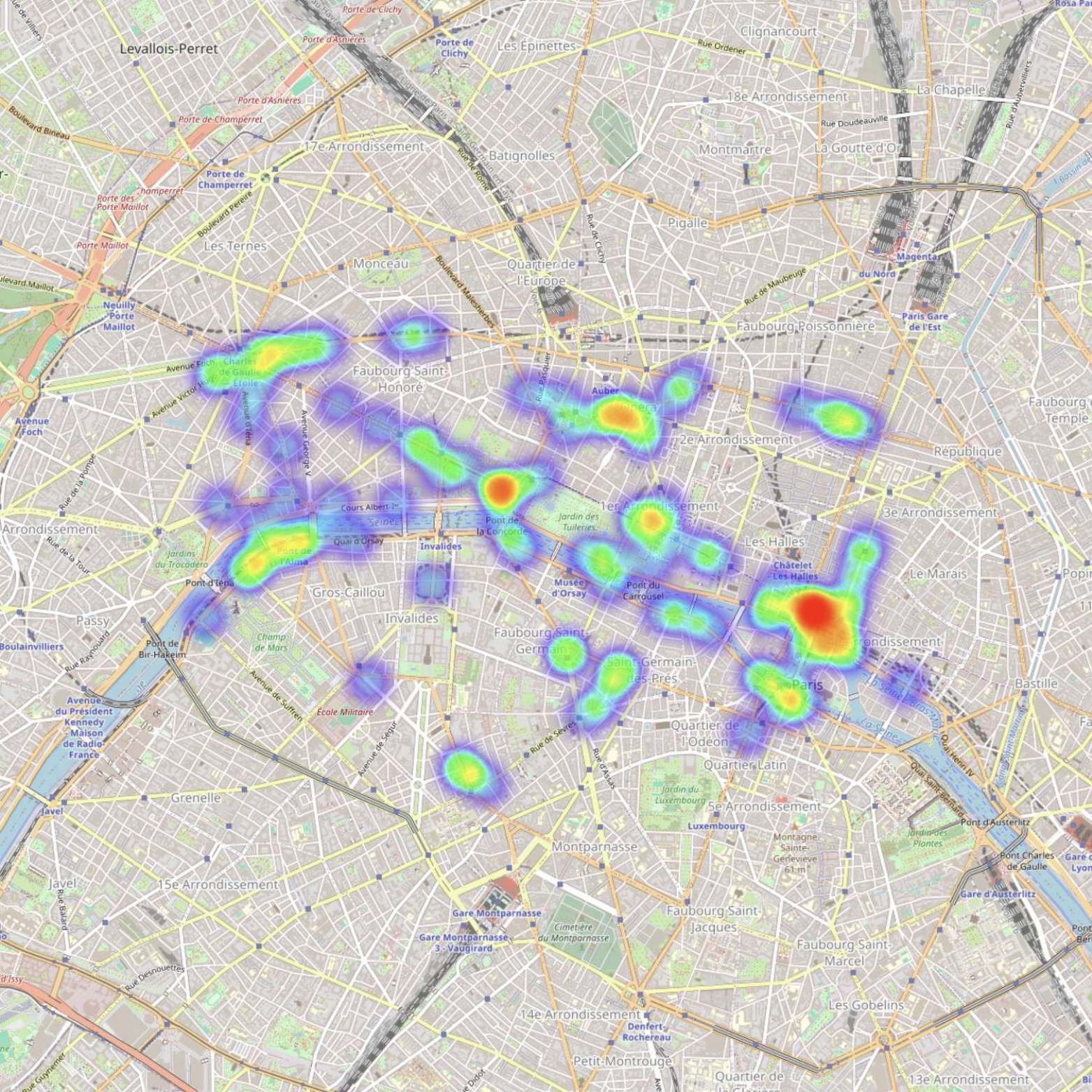}
        \centerline{\footnotesize OCC-Pairs}
    \end{minipage}
    \hspace{6mm}
    \begin{minipage}[b]{0.20\linewidth}
        \includegraphics[width=\linewidth]{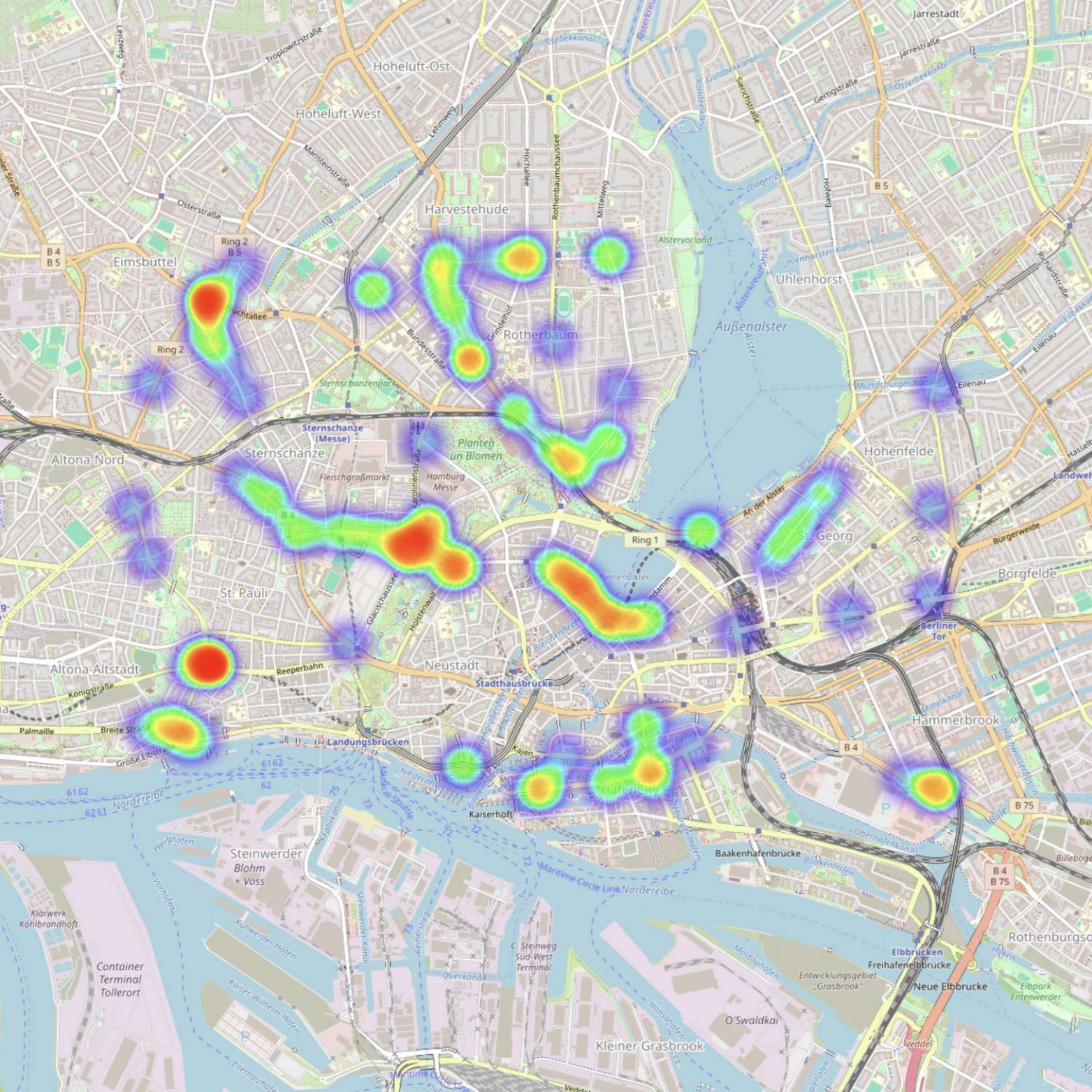}
        \centerline{\footnotesize OCC-Hamburg}
    \end{minipage}
    \hspace{6mm}
    \begin{minipage}[b]{0.20\linewidth}
        \includegraphics[width=\linewidth]{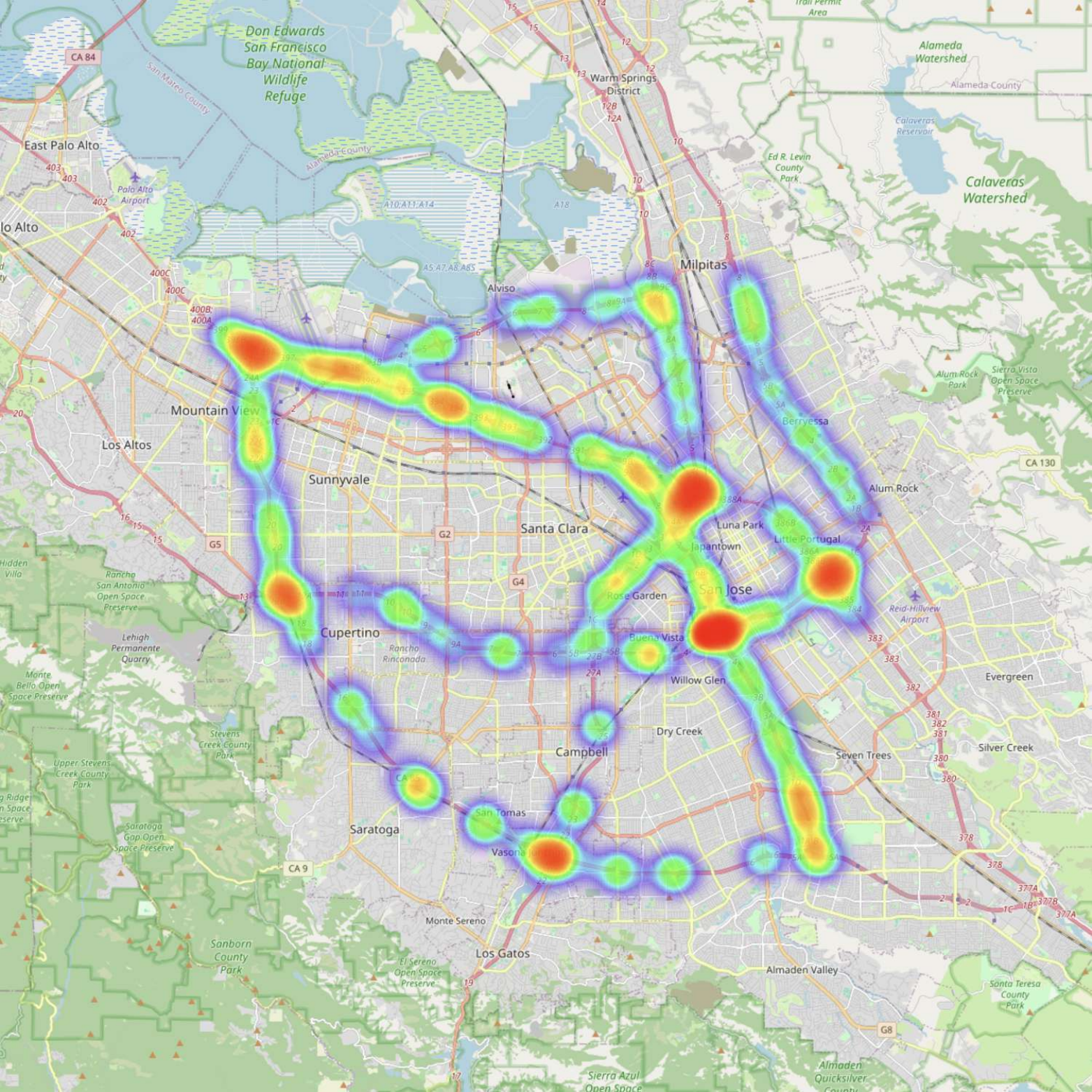}
        \centerline{\footnotesize PEMS-BAY}
    \end{minipage}
    \hspace{6mm}
    \begin{minipage}[b]{0.20\linewidth}
        \includegraphics[width=\linewidth]{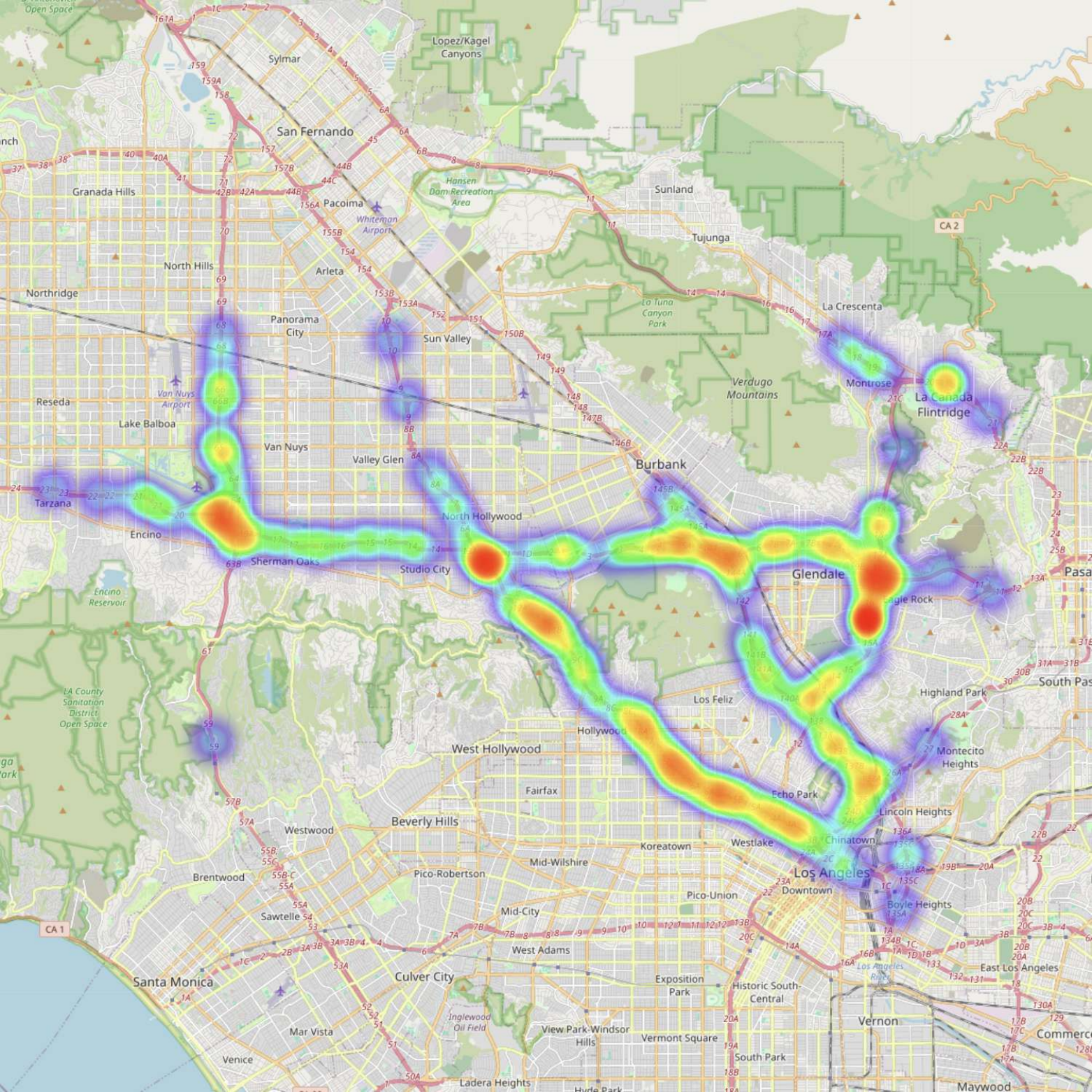}
        \centerline{\footnotesize METR-LA}
    \end{minipage}\\
    
    \vspace{4mm}
    
    \begin{minipage}[b]{0.20\linewidth}
        \includegraphics[width=\linewidth]{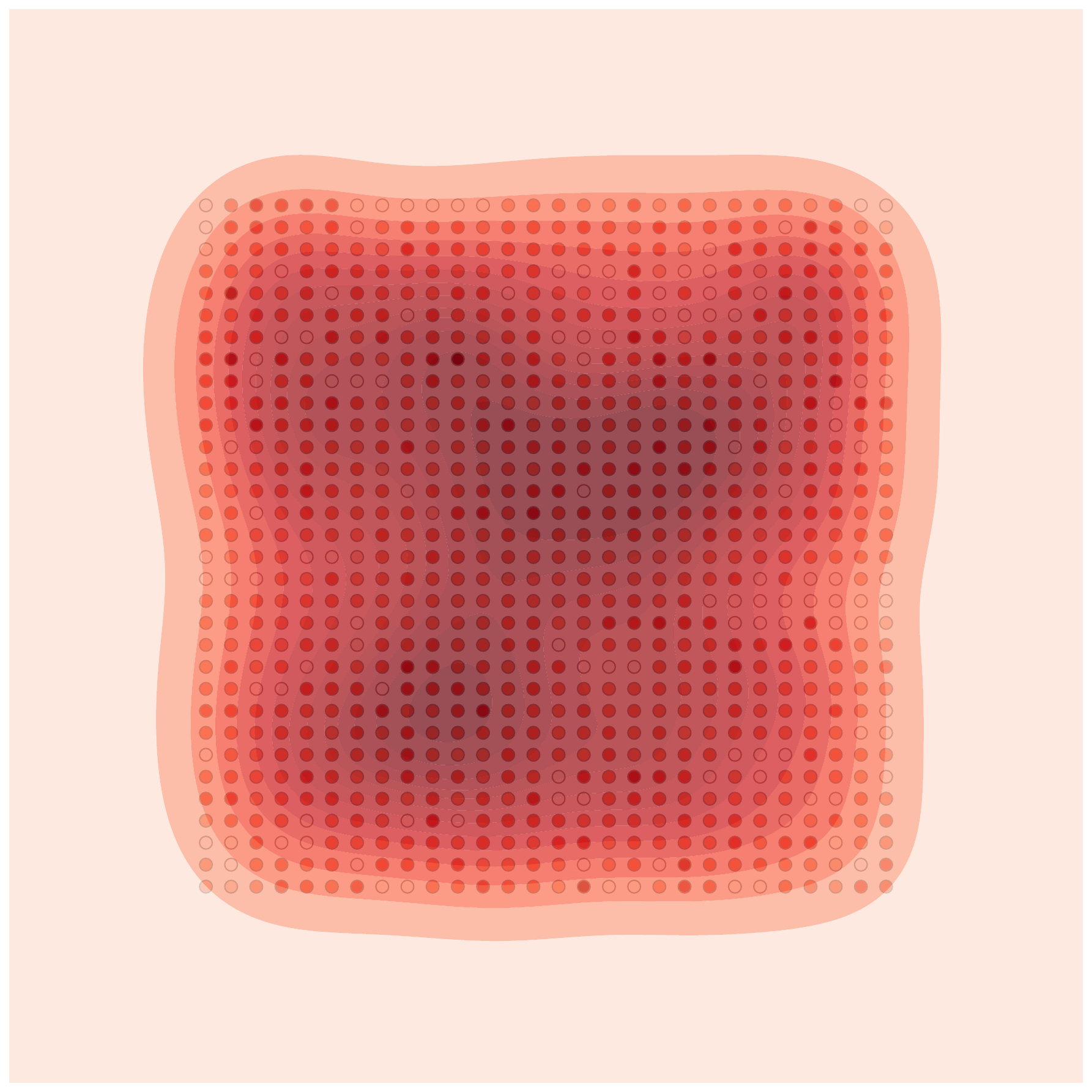}
        \centerline{\footnotesize Traffic-SH}
    \end{minipage}
    \hspace{6mm}
    \begin{minipage}[b]{0.20\linewidth}
        \includegraphics[width=\linewidth]{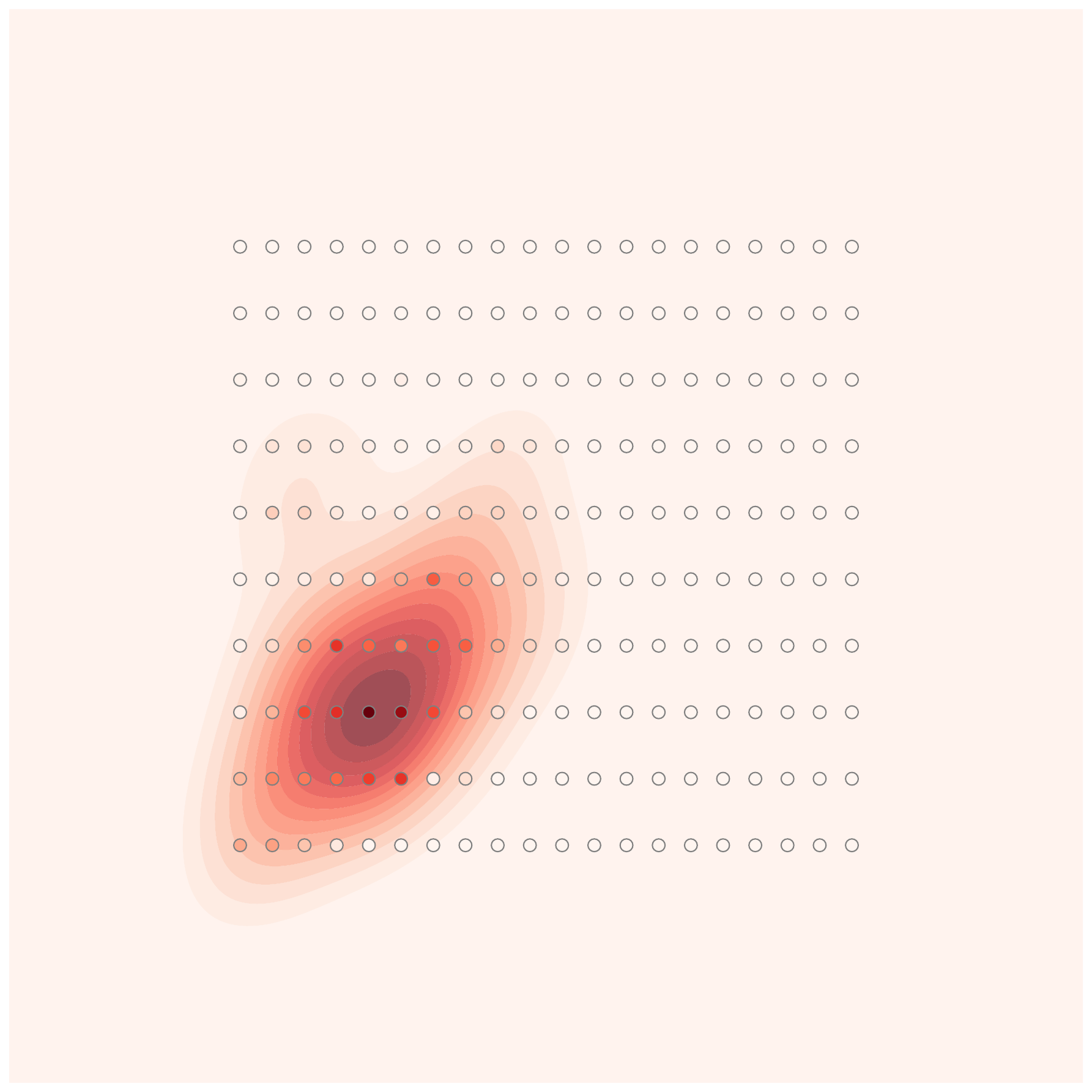}
        \centerline{\footnotesize Bike-NYC}
    \end{minipage}
    \hspace{6mm}
    \begin{minipage}[b]{0.20\linewidth}
        \includegraphics[width=\linewidth]{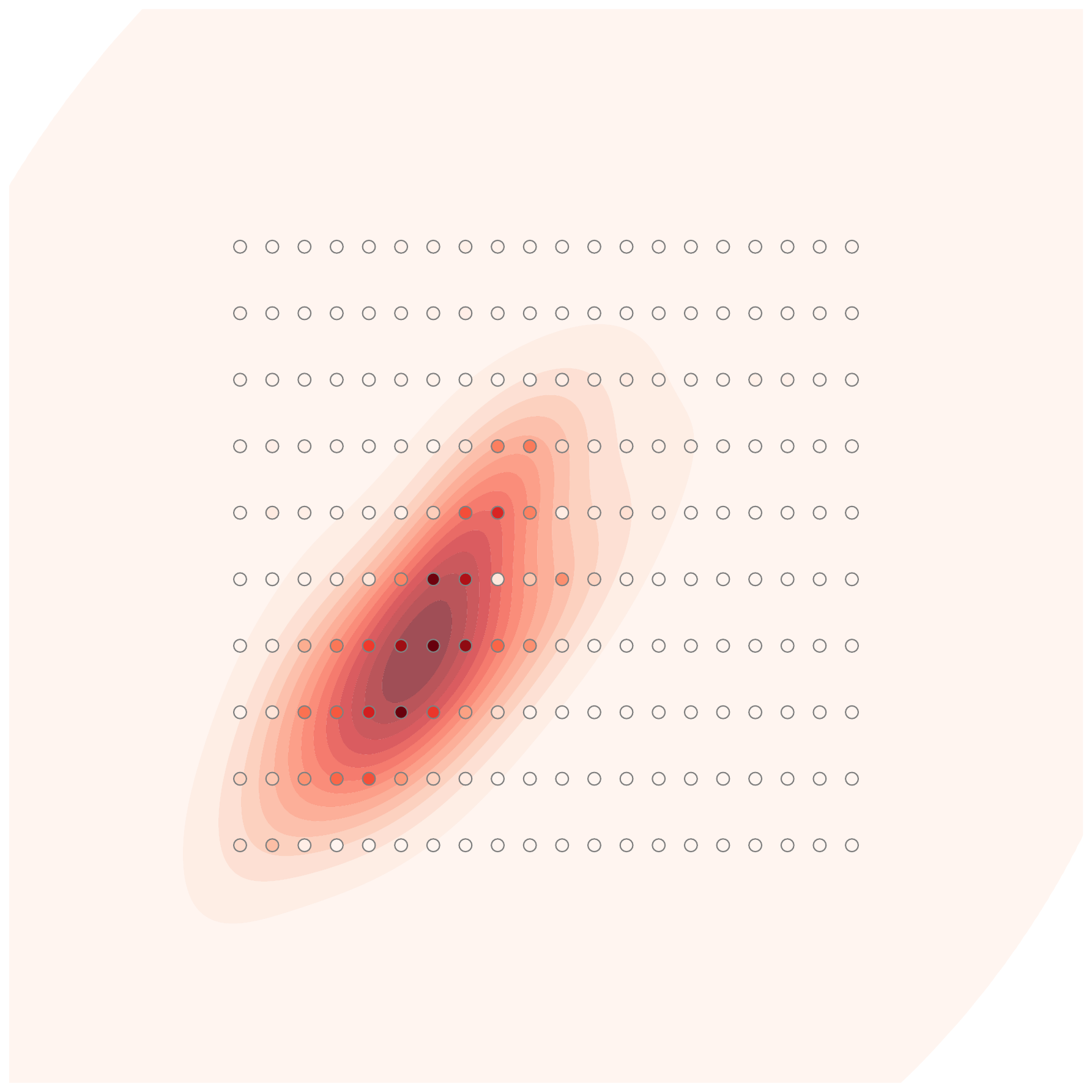}
        \centerline{\footnotesize Taxi-NYC}
    \end{minipage}
    \hspace{6mm}
    \begin{minipage}[b]{0.20\linewidth}
        \includegraphics[width=\linewidth]{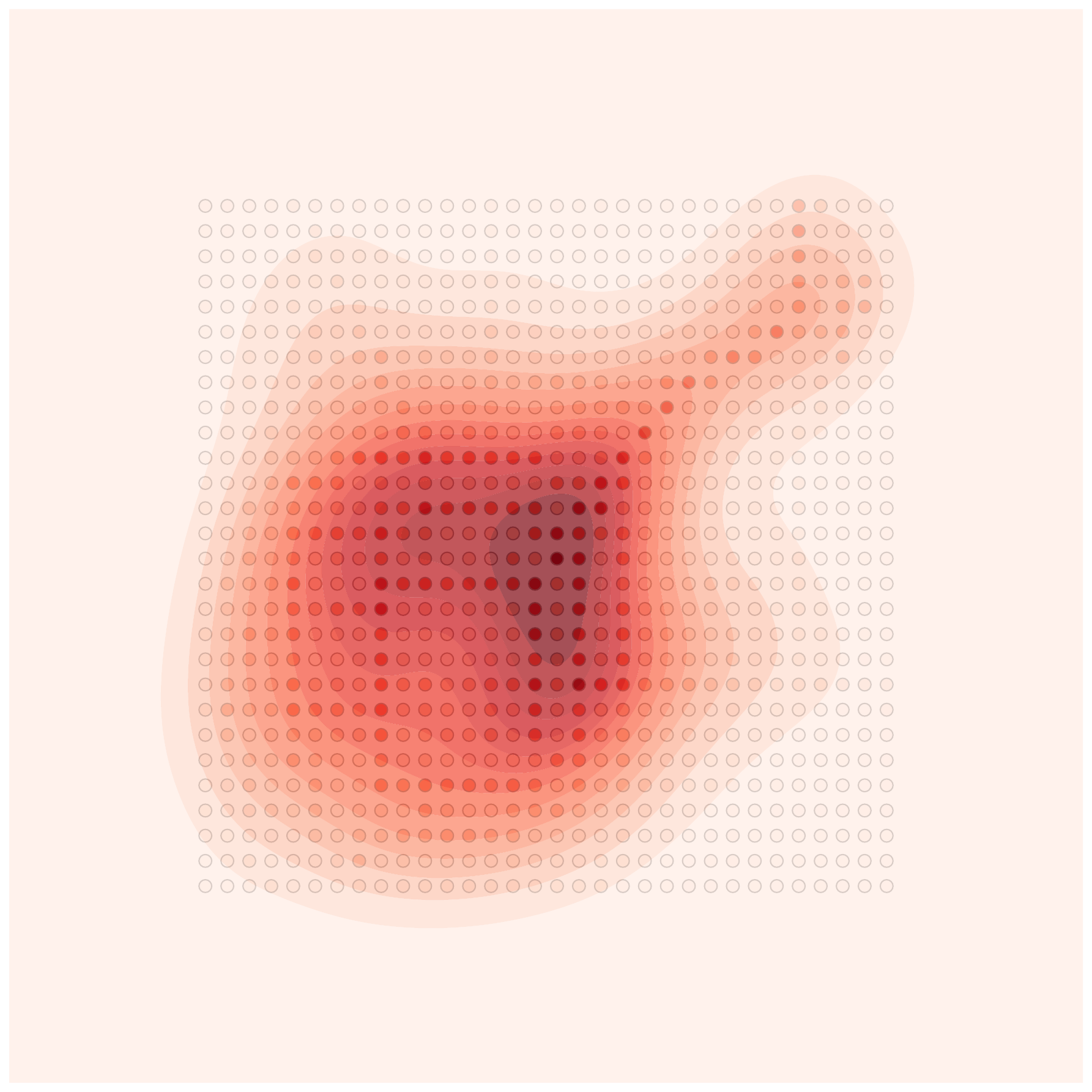}
        \centerline{\footnotesize Tdrive-BJ}
    \end{minipage}
    
    \caption{Spatial Visualization of~\benchmark Benchmark.}
    \vspace{-0.1in}
    \label{fig:benchmark_spatial_vis}
\end{figure*}

\begin{figure*}[htbp!]
    \centering
    \includegraphics[width=1.0\linewidth]{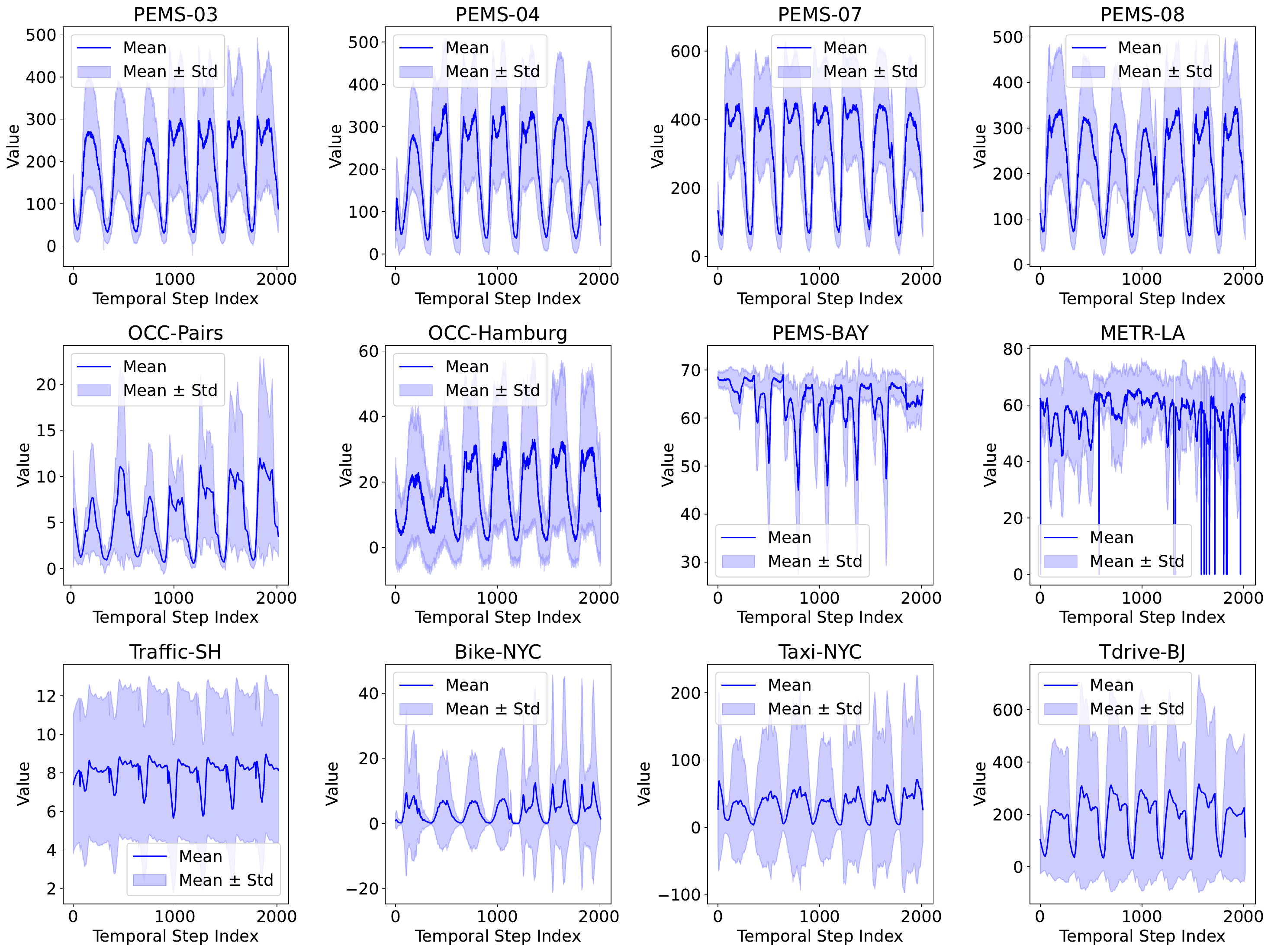}
    \caption{Temporal Visualization of \benchmark~Benchmark.}
    \captionsetup[subfigure]{labelformat=empty} 
    \label{fig:benchmark_t_vis}
\end{figure*}

\begin{figure*}[t!]
    \centering
    \includegraphics[width=1.0\linewidth]{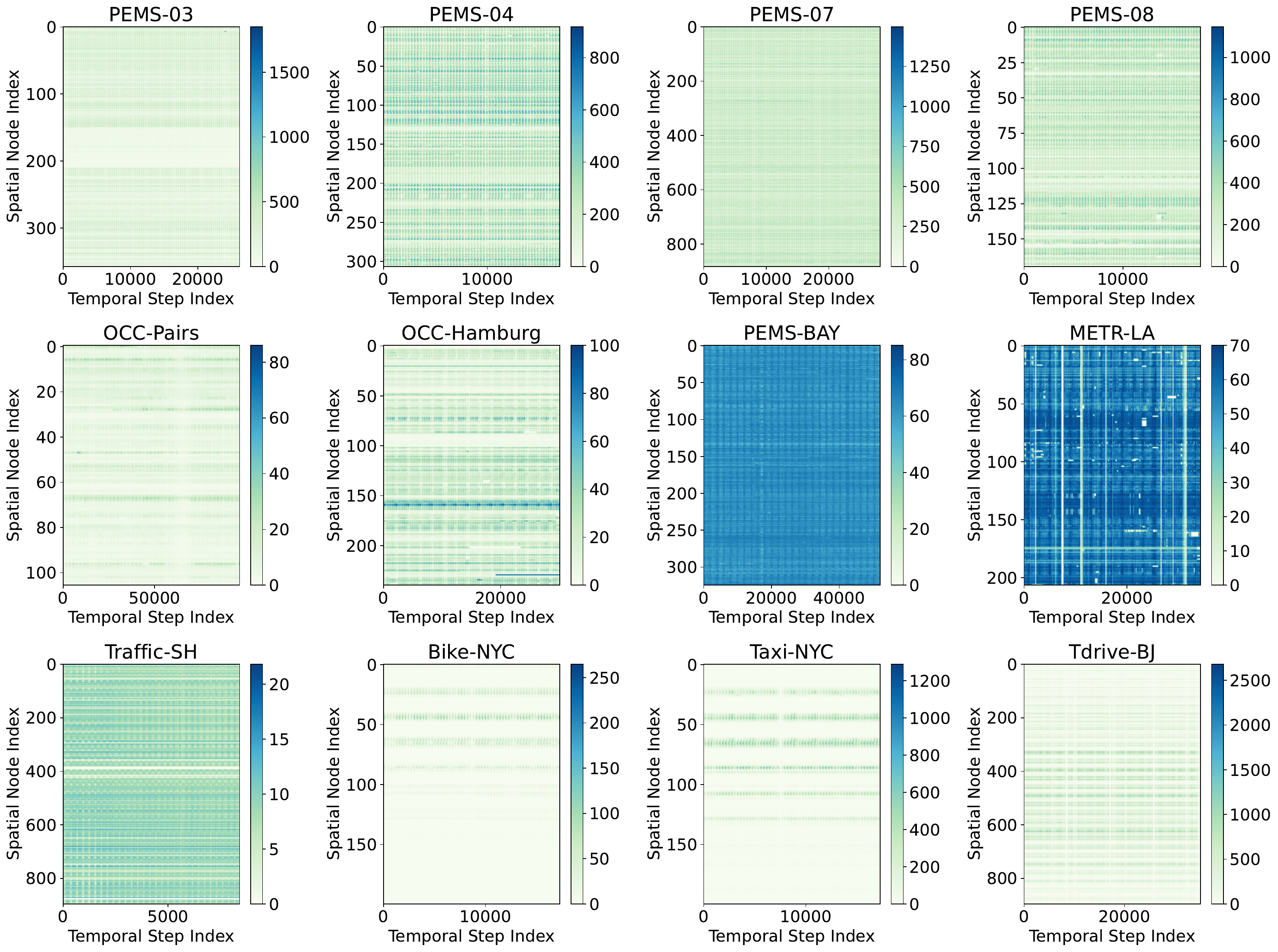}
    \caption{Spatio-Temporal Visualization of \benchmark Benchmark.}
    \captionsetup[subfigure]{labelformat=empty} 
    \label{fig:benchmark_st_vis}
\end{figure*}

We first provide the statistical information of the evaluation benchmark as shown in Table~\ref{tab:benchmark_summary}. Then, we visualize its spatial, temporal, and overall spatio-temporal distribution as shown in Figures~\ref{fig:benchmark_spatial_vis},~\ref{fig:benchmark_t_vis}, and~\ref{fig:benchmark_st_vis}.

\renewcommand{\arraystretch}{1.25}
\setlength{\tabcolsep}{6pt}

\begin{table}[htbp!]
\caption{Evaluation Benchmark Summary}
\label{tab:benchmark_summary}
\centering
\vspace{-2mm}
\resizebox{\linewidth}{!}{
\begin{tabular}{cccccccc}

\toprule

\multirow{2}{*}{\makecell[c]{\textbf{Data}\\\textbf{Format}}} 
& \multirow{2}{*}{\makecell[c]{\textbf{Domain}\\\textbf{Type}}} & \multirow{2}{*}{\makecell[c]{\textbf{Dataset}\\\textbf{Name}}} & \multirow{2}{*}{\makecell[c]{\textbf{Spatial}\\\textbf{Region}}} & \multirow{2}{*}{\makecell[c]{\textbf{\# Spatial}\\\textbf{Location}}} & \multirow{2}{*}{\makecell[c]{\textbf{Temporal}\\\textbf{Range}}} & \multirow{2}{*}{\makecell[c]{\textbf{\# Temporal}\\\textbf{Step}}} & \multirow{2}{*}{\textbf{\textbf{Overlap}}} \\ 

& & & & & & \\ 

\midrule

\multirow{7}{*}{\makecell[c]{Sensor\\-based}} & 

\multirow{4}{*}{Flow} 
& PEMS-03 & Los Angles, US & 358 & 2018/09/01 -- 2018/11/30 & 26208 & $\times$ \\

&  & PEMS-04 & Los Angles, US & 307 & 2018/01/01 -- 2018/02/28 & 16992 & $\times$ \\

&  & PEMS-07 & Los Angles, US & 883 & 2017/05/01 -- 2017/08/06 & 28224 & $\times$ \\

&  & PEMS-08 & Los Angles, US & 170 & 2016/07/01 -- 2016/08/31 & 17856 & $\times$ \\

\cline{2-8}

& \multirow{2}{*}{Occupancy} & OCC-Pairs & Pairs, FR & 106 & 2016/01/01 -- 2016/12/01 & 96410 & $\times$ \\

&  & OCC-Hamburg & Hamburg, DE & 240 & 2016/08/27 -- 2016/12/09 & 30085 & $\times$ \\ 

\cline{2-8}

& \multirow{2}{*}{Speed} 
& PEMS-BAY & Los Angles, US & 325 & 2017/01/01 -- 2017/06/30 & 52116 & $\times$ \\

&  & METR-LA & Los Angles, US & 207 & 2012/03/01 -- 2012/06/27 & 34272 & $\times$ \\

\midrule

\multirow{4}{*}{\makecell[c]{Grid\\-based}} & Road & Traffic-SH & Shanghai, CN & $28 \times 32 = 896$ & 2022/01/27 -- 2022/02/25 & 8413 & $\times$ \\

\cline{2-8}

& Bike & Bike-NYC & New York, US & $20 \times 10 = 200$ & 2016/07/01 -- 2016/08/29 & 17275 & $\times$ \\

\cline{2-8}

& Taxi & Taxi-NYC & New York, US & $20 \times 10 = 200$ & 2015/01/01 -- 2015/03/31 & 17275 & $\times$ \\

\cline{2-8}

& Trajectory & Tdrive-BJ & Beijing, CN & $28 \times 32 = 896$ & 2015/03/01 -- 2015/06/30 & 35125 & $\times$ 

\\ 

\bottomrule

\end{tabular}
}
\end{table}

\subsection{Pre-training Dataset: \dataset}

We then provide statistics on the collected dataset \dataset~that can be used for pre-training, as shown in Table~\ref{tab:dataset_summary}.

\renewcommand{\arraystretch}{1.2}
\setlength{\tabcolsep}{6pt}

\begin{longtable}{cccccc}
\caption{Pre-training Datasets Summary} \label{tab:dataset_summary}\\

\toprule

\multirow{2}{*}{\cellcolor{white}\makecell[c]{\textbf{Data}\\\textbf{Format}}}
& \multirow{2}{*}{\makecell[c]{\textbf{Domain}\\\textbf{Type}}} & \multirow{2}{*}{\makecell[c]{\textbf{Spatial}\\\textbf{Region}}} & \multirow{2}{*}{\makecell[c]{\textbf{\# Spatial}\\\textbf{Location}}} & \multirow{2}{*}{\makecell[c]{\textbf{Temporal}\\\textbf{Range}}} & \multirow{2}{*}{\makecell[c]{\textbf{\# Temporal}\\\textbf{Step}}} \\ 
& & & & & \\ 
\midrule
\endhead

\bottomrule
\endfoot

Sensor & Flow & Augsburg, DE & 40 & 2017/05/06 -- 2017/05/25 & 5757 \\
Sensor & Flow & Basel, CH & 66 & 2016/10/24 -- 2016/10/31 & 2016 \\
Sensor & Flow & Bern, CH & 708 & 2016/10/24 -- 2016/10/31 & 2016 \\
Sensor & Flow & Birmingham, GB & 34 & 2017/10/23 -- 2017/11/18 & 7540 \\
Sensor & Flow & Bolton, GB & 56 & 2017/11/13 -- 2017/11/18 & 1491 \\
Sensor & Flow & Bordeaux, FR & 409 & 2016/11/21 -- 2016/11/27 & 2016 \\
Sensor & Flow & Bremen, DE & 518 & 2016/09/19 -- 2016/10/02 & 4032 \\
Sensor & Flow & Cagliari, IT & 85 & 2016/05/16 -- 2016/07/29 & 21600 \\
Sensor & Flow & Constance, DE & 113 & 2017/02/13 -- 2017/02/19 & 2015 \\
Sensor & Flow & Darmstadt, DE & 163 & 2015/09/21 -- 2016/05/06 & 65952 \\
Sensor & Flow & Essen, DE & 36 & 2017/03/27 -- 2017/09/30 & 54023 \\
Sensor & Flow & Frankfurt, DE & 73 & 2016/12/21 -- 2016/12/21 & 288 \\
Sensor & Flow & Graz, AT & 272 & 2016/04/04 -- 2016/09/23 & 49823 \\
Sensor & Flow & Groningen, NL & 52 & 2017/09/21 -- 2017/10/06 & 4584 \\
Sensor & Flow & Hamburg, DE & 376 & 2016/08/27 -- 2016/12/09 & 30085 \\
Sensor & Flow & Innsbruck, AT & 16 & 2017/04/01 -- 2017/04/30 & 8639 \\
Sensor & Flow & Kassel, DE & 234 & 2016/08/28 -- 2016/09/02 & 1177 \\
Sensor & Flow & London, GB & 4130 & 2015/05/15 -- 2016/05/22 & 107712 \\
Sensor & Flow & Los Angeles, US & 957 & 2017/10/02 -- 2017/10/10 & 2592 \\
Sensor & Flow & Luzern, CH & 134 & 2015/01/01 -- 2016/01/01 & 105121 \\
Sensor & Flow & Madrid, ES & 1065 & 2016/08/29 -- 2017/11/11 & 126549 \\
Sensor & Flow & Manchester, GB & 114 & 2017/09/08 -- 2017/11/18 & 20497 \\
Sensor & Flow & Marseille, FR & 164 & 2017/06/01 -- 2017/07/01 & 8641 \\
Sensor & Flow & Melbourne, AU & 861 & 2018/02/12 -- 2018/02/26 & 4318 \\
Sensor & Flow & Munich, DE & 509 & 2017/02/14 -- 2017/02/15 & 288 \\
Sensor & Flow & Paris, FR & 169 & 2016/01/01 -- 2016/12/01 & 96469 \\
Sensor & Flow & Rotterdam, NL & 230 & 2017/09/21 -- 2017/11/02 & 12276 \\
Sensor & Flow & Santander, ES & 211 & 2016/06/17 -- 2016/12/02 & 48359 \\
Sensor & Flow & Stuttgart, DE & 177 & 2016/03/21 -- 2016/07/22 & 35711 \\
Sensor & Flow & Taipei, CN-TW & 275 & 2017/09/18 -- 2017/10/01 & 4032 \\
Sensor & Flow & Torino, IT & 267 & 2016/09/26 -- 2016/10/16 & 6048 \\
Sensor & Flow & Toronto, CA & 151 & 2016/09/01 -- 2017/01/31 & 44062 \\
Sensor & Flow & Toulouse, FR & 469 & 2008/05/16 -- 2008/06/27 & 12384 \\
Sensor & Flow & Utrecht, NL & 865 & 2017/06/12 -- 2017/06/15 & 1152 \\
Sensor & Flow & Vilnius, LT & 9 & 2015/03/17 -- 2015/03/18 & 289 \\
Sensor & Flow & Wolfsburg, DE & 103 & 2016/09/19 -- 2016/10/02 & 4032 \\
Sensor & Flow & Zurich, CH & 996 & 2015/10/26 -- 2015/11/01 & 2016 \\

Sensor & Flow & Sydney, AU & 27 & 2013/01/02 -- 2024/05/31 & 4169 \\


Sensor & Occupancy & Augsburg, DE & 35 & 2017/05/06 -- 2017/05/25 & 5757 \\
Sensor & Occupancy & Basel, CH & 47 & 2016/10/24 -- 2016/10/31 & 2016 \\
Sensor & Occupancy & Bern, CH & 496 & 2016/10/24 -- 2016/10/31 & 2016 \\
Sensor & Occupancy & Bolton, GB & 49 & 2017/11/13 -- 2017/11/18 & 1491 \\
Sensor & Occupancy & Bordeaux, FR & 198 & 2016/11/21 -- 2016/11/27 & 2016 \\
Sensor & Occupancy & Bremen, DE & 365 & 2016/09/19 -- 2016/10/02 & 4032 \\
Sensor & Occupancy & Cagliari, IT & 61 & 2016/05/16 -- 2016/07/29 & 21600 \\
Sensor & Occupancy & Constance, DE & 103 & 2017/02/13 -- 2017/02/19 & 2015 \\
Sensor & Occupancy & Darmstadt, DE & 101 & 2015/09/21 -- 2016/05/06 & 65952 \\
Sensor & Occupancy & Essen, DE & 24 & 2017/03/27 -- 2017/09/30 & 54023 \\
Sensor & Occupancy & Frankfurt, DE & 59 & 2016/12/21 -- 2016/12/21 & 288 \\
Sensor & Occupancy & Graz, AT & 232 & 2016/04/04 -- 2016/09/23 & 49823 \\
Sensor & Occupancy & Groningen, NL & 42 & 2017/09/21 -- 2017/10/06 & 4584 \\
Sensor & Occupancy & Kassel, DE & 179 & 2016/08/28 -- 2016/09/02 & 1177 \\
Sensor & Occupancy & London, GB & 2793 & 2015/05/15 -- 2016/05/22 & 107712 \\
Sensor & Occupancy & Luzern, CH & 108 & 2015/01/01 -- 2016/01/01 & 105121 \\
Sensor & Occupancy & Madrid, ES & 659 & 2016/08/29 -- 2017/11/11 & 126549 \\
Sensor & Occupancy & Manchester, GB & 15 & 2017/09/08 -- 2017/11/18 & 20497 \\
Sensor & Occupancy & Marseille, FR & 115 & 2017/06/01 -- 2017/07/01 & 8641 \\
Sensor & Occupancy & Munich, DE & 322 & 2017/02/14 -- 2017/02/15 & 288 \\
Sensor & Occupancy & Rotterdam, NL & 108 & 2017/09/21 -- 2017/11/02 & 12276 \\
Sensor & Occupancy & Santander, ES & 157 & 2016/06/17 -- 2016/12/02 & 48359 \\
Sensor & Occupancy & Speyer, DE & 118 & 2016/09/19 -- 2016/10/02 & 4032 \\
Sensor & Occupancy & Strasbourg, FR & 82 & 2017/05/10 -- 2017/11/11 & 53221 \\
Sensor & Occupancy & Stuttgart, DE & 137 & 2016/03/21 -- 2016/07/22 & 35711 \\
Sensor & Occupancy & Taipei, CN-TW & 227 & 2017/09/18 -- 2017/10/01 & 4032 \\
Sensor & Occupancy & Torino, IT & 228 & 2016/09/26 -- 2016/10/16 & 6048 \\
Sensor & Occupancy & Toulouse, FR & 364 & 2008/05/16 -- 2008/06/27 & 12384 \\
Sensor & Occupancy & Vilnius, LT & 8 & 2015/03/17 -- 2015/03/18 & 289 \\
Sensor & Occupancy & Wolfsburg, DE & 60 & 2016/09/19 -- 2016/10/02 & 4032 \\
Sensor & Occupancy & Zurich, CH & 575 & 2015/10/26 -- 2015/11/01 & 2016 \\

Sensor & Speed & Birmingham, GB & 4 & 2017/10/23 -- 2017/11/18 & 7540 \\
Sensor & Speed & Bolton, GB & 16 & 2017/11/13 -- 2017/11/18 & 1491 \\
Sensor & Speed & Constance, DE & 42 & 2017/02/13 -- 2017/02/19 & 2015 \\
Sensor & Speed & Essen, DE & 15 & 2017/03/27 -- 2017/09/30 & 54023 \\
Sensor & Speed & Groningen, NL & 8 & 2017/09/21 -- 2017/10/06 & 4584 \\
Sensor & Speed & Manchester, GB & 29 & 2017/09/08 -- 2017/11/18 & 20497 \\
Sensor & Speed & Rotterdam, NL & 70 & 2017/09/21 -- 2017/11/02 & 12276 \\
Sensor & Speed & Torino, IT & 71 & 2016/09/26 -- 2016/10/16 & 6048 \\

Sensor & Speed & Guangzhou, CN & 214 & 2016/08/01 -- 2016/09/30 & 9216 \\

Sensor & Bus Demand & Montevideo, UY & 675 & 2020/10/01 -- 2020/10/29 & 744 \\

Grid & Taxi Demand & Chicago, US & 77 & 2021/01/01 -- 2021/12/31 & 17520 \\
Grid & Traffic Speed & Zhengzhou, CN & 676 & 2022/03/05 -- 2022/04/05 & 1403 \\
Grid & Traffic Speed & Hangzhou, CN & 672 & 2022/03/05 -- 2022/04/05 & 1403 \\
Grid & Traffic Speed & Chengdu, CN & 728 & 2022/03/05 -- 2022/04/05 & 1403 \\
Grid & Traffic Speed & Jinan, CN & 579 & 2022/03/05 -- 2022/04/05 & 1403 \\
Grid & Traffic Index & Shenzhen, CN & 627 & 2017/01/01 -- 2018/02/28 & 17280 \\
Grid & Traffic Index & Chengdu, CN & 524 & 2018/01/01 -- 2018/02/28 & 17280 \\

Grid & Crowd & Nanjing, CN & 320 & 2020/11/11 -- 2021/05/31 & 6980 \\
Grid & Cellular & Nanjing, CN & 320 & 2020/11/11 -- 2021/05/31 & 6980 \\

\end{longtable}

\vspace{-4mm}

\section{Method Detail}\label{appendix_method}

\begin{figure*}[htbp!]
    \centering
    \includegraphics[width=1.0\textwidth]{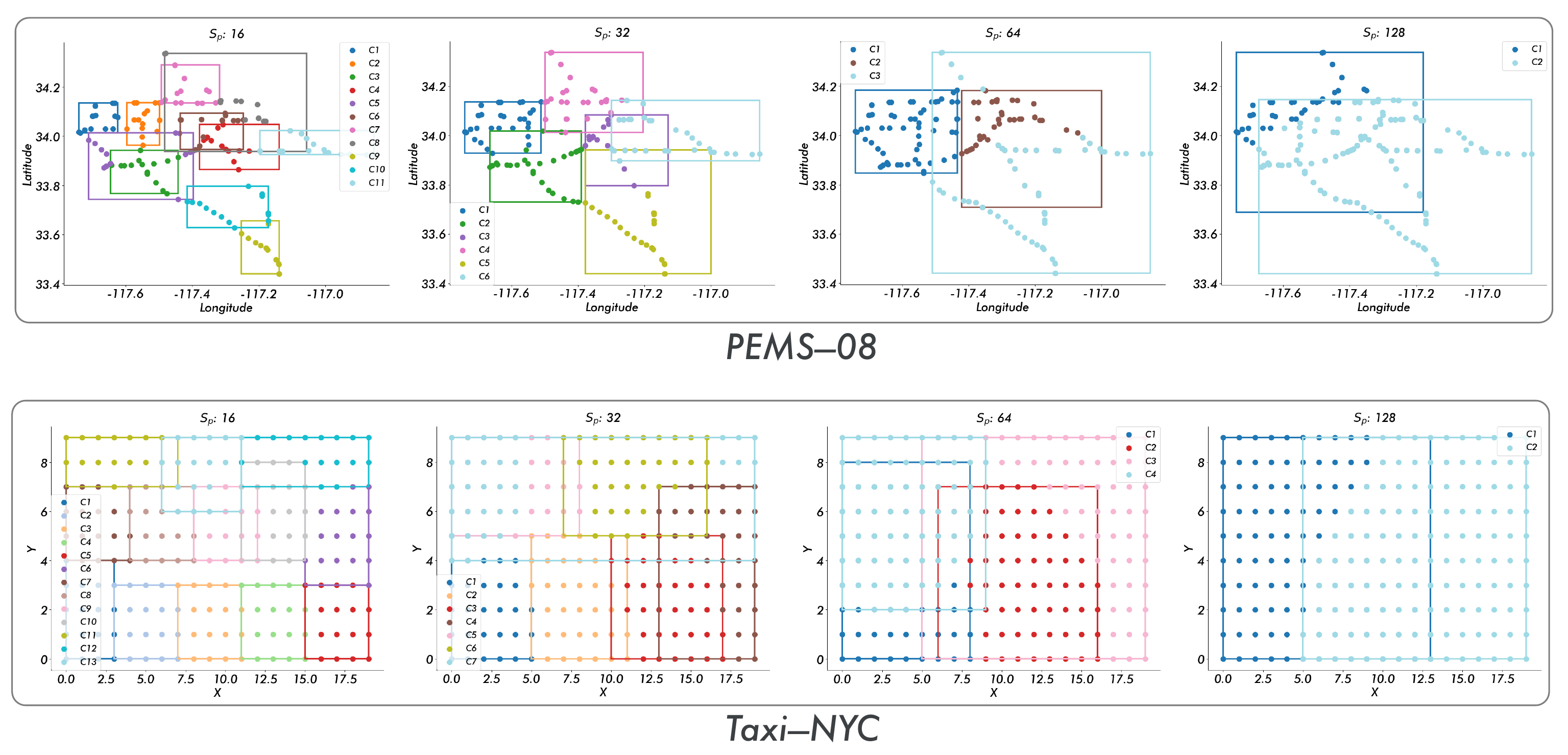}
    \caption{Visualization of greedy capacity-constrained clustering with different $S_p$ on PEMS-08 and Taxi-NYC.}
    \label{fig:cluster_vis}
\end{figure*}

We first summarize the specific implementation details of our proposed Greedy Capacity-Constrained Clustering in Algorithm workflow~\ref{alg:kd_cluster}. Furthermore, we provide schematic diagrams~\ref{fig:cluster_vis} of clustering visualization effects under different capacities for two types of datasets: the sensor-based PEMS-08 and the grid-based Taxi-NYC.

\begin{algorithm}[H]
    \fontsize{8.3pt}{8.3pt}\selectfont
    \caption{Greedy Capacity-Constrained Clustering Algorithm Workflow.}
    \label{alg:kd_cluster}
    
    \SetKwInOut{Input}{Input}
    \SetKwInOut{Output}{Output}
    \SetKwComment{tcp}{// }{}
    
    \Input{A set of points $P = \{p_1, \dots, p_n\}$; capacity parameter $d$}
    \Output{Clusters $C$ and corresponding index sets $C_{idx}$}
    
    $n \gets |P|$\;
    Build a KDTree $T$ using the points in $P$; Initialize $Assigned[1 \dots n] \gets \text{False}$; $C, C_{idx} \gets \emptyset$\;
    
    \If{$d \leq n$}{
        \For{$i \gets 1$ \KwTo $n$}{
            \If{$Assigned[i] = \text{False}$}{
                $(\text{distances}, I) \gets \text{KDTreeQuery}(T, p_i, k=d)$\;
                $cluster \gets \emptyset, cluster\_idx \gets \emptyset$\;
                \ForEach{$idx \in I$}{
                    \If{$Assigned[idx] = \text{False}$}{
                        Add $p_{idx}$ to $cluster$; Add $idx$ to $cluster\_idx$\;
                        \If{$|cluster| = d$}{\textbf{break}\;}
                    }
                }
                \If{$|cluster| < d$}{
                    \tcp{If less than $d$ points found, query all points}
                    $(\text{distances}_{ext}, I_{ext}) \gets \text{KDTreeQuery}(T, p_i, k=n)$\;
                    \ForEach{$idx \in I_{ext}$}{
                        \If{$Assigned[idx] = \text{False}$ \textbf{and} $idx \notin cluster\_idx$}{
                            Add $p_{idx}$ to $cluster$; Add $idx$ to $cluster\_idx$\;
                            \If{$|cluster| = d$}{\textbf{break}\;}
                        }
                    }
                }
                \ForEach{$idx \in cluster\_idx$}{$Assigned[idx] \gets \text{True}$\;}
                Append $cluster$ to $C$; Append $cluster\_idx$ to $C_{idx}$\;
            }
        }
        \tcp{Handle the last cluster if its size is less than $d$}
        \If{$C \neq \emptyset$ \textbf{and} $|C[\text{last}]| < d$}{
            $last\_c \gets \text{last}(C); last\_idx \gets \text{last}(C_{idx}); needed \gets d - |last\_c|$\;
            \ForEach{$p \in last\_c$}{
                $(\text{distances}, I') \gets \text{KDTreeQuery}(T, p, k=n)$\;
                \ForEach{$idx \in I'$}{
                    \If{$p_{idx} \notin last\_c$}{
                        Add $p_{idx}$ to $last\_c$; Add $idx$ to $last\_idx$; $needed \gets needed - 1$\;
                        \If{$needed = 0$}{\textbf{break}\;}
                    }
                }
                \If{$needed = 0$}{\textbf{break}\;}
            }
            $C[\text{last}] \gets last\_c; C_{idx}[\text{last}] \gets last\_idx$\;
        }
    }
    \Else{
        \tcp{Case when $d > n$: fill by repeating points}
        $full \gets \lfloor d / n \rfloor, rem \gets d \bmod n$\;
        \For{$j \gets 1$ \KwTo $full$}{
            Append $P$ to $C$; Append $\{1, \dots, n\}$ to $C_{idx}$\;
        }
        \If{$rem > 0$}{
            Append first $rem$ points of $P$ to $C$\;
            Append $\{1, \dots, rem\}$ to $C_{idx}$\;
        }
    }
    \Return{$C, C_{idx}$}\;
\end{algorithm}

\section{Experimental Details}\label{appendix_setup}

\subsection{Baseline Details}\label{appendix_baseline}

In this appendix, we provide detailed descriptions of the expert and foundation baseline models used in our default evaluation. 

\subsubsection{Time Series Models} 

\vspace{0.5em}
\begin{itemize}[noitemsep, topsep=8pt, partopsep=0pt, leftmargin=6mm,parsep=8pt]
\setlength\itemsep{0mm}
\item \textit{PatchTST}~\citep{nie2022time}: 
\textit{PtachTST} is a Transformer-based model that segments time series into sub-series patches. By leveraging patching and channel independence, it effectively reduces computational complexity while capturing both local semantic patterns and global long-term dependencies. \url{https://github.com/yuqinie98/PatchTST}

\item \textit{Dlinear}~\citep{zeng2023transformers}: \textit{Dlinear} is a simple yet robust baseline that decomposes time series into trend and remainder components. It models each component with a single-layer linear network, demonstrating that simple linear mappings can achieve high accuracy and robustness against distribution shifts. \url{https://github.com/cure-lab/LTSF-Linear}

\item \textit{LSTM}~\citep{hochreiter1997long}: 
\textit{LSTM} is a classic temporal modeling architecture designed to handle sequential dependencies. In the context of spatio-temporal learning, it serves as a fundamental building block for capturing non-linear temporal correlations and long-term trends across time steps. \url{https://github.com/liuxu77/LargeST/blob/main/src/models/lstm.py}

\item \textit{HA}~\citep{cui2021historical}:
\textit{HA} is a naive baseline that exploits the inherent historical inertia in time series data, directly utilizing the most recent input sequence as the prediction. \url{https://github.com/GestaltCogTeam/BasicTS/blob/master/src/basicts/models/HI/arch/hi_arch.py}

\end{itemize}

\subsubsection{ST-Grid Models} 

\vspace{0.5em}
\begin{itemize}[noitemsep, topsep=8pt, partopsep=0pt, leftmargin=6mm,parsep=8pt]
\setlength\itemsep{0mm}
\item \textit{TAU}~\citep{tan2023temporal}: 
\textit{TAU} is a non-recurrent spatio-temporal model designed to parallelize the learning of temporal evolution. It introduces a novel attention mechanism that splits the temporal context into static (invariant) and dynamic (changing) components, allowing for effective global temporal modeling without the sequential constraints of RNNs.
\url{https://github.com/chengtan9907/OpenSTL}

\item \textit{PredRNN++}~\citep{wang2022predrnn}: \textit{PredRNN++} is a spatio-temporal recurrent network that introduces the Causal LSTM structure to cascade dual memory states and employs gradient highway units to enable efficient training across long sequences.
\url{https://github.com/Yunbo426/predrnn-pp}

\item \textit{DSAN}~\citep{lin2020preserving}:  \textit{DSAN} is a spatial-temporal forecasting framework designed to achieve effective long-term spatial-temporal prediction by filtering out spatial noise and alleviating long-term error propagation. \url{https://github.com/haoxingl/DSAN}

\item \textit{ST-ResNet}~\citep{zhang2017deep}: \textit{ST-ResNet} is a grid-based spatial-temporal framework designed to model temporal closeness, period, and trend properties using three separate residual neural networks and aggregates them to capture citywide spatial-temporal dependencies.
\url{https://github.com/topazape/ST-ResNet}
\end{itemize}

\subsubsection{ST-Graph Models} 

\vspace{0.5em}
\begin{itemize}[noitemsep, topsep=8pt, partopsep=0pt, leftmargin=6mm,parsep=8pt]
\setlength\itemsep{0mm}
\item \textit{STAEformer}~\citep{liu2023spatio}: \textit{STAEformer} is a spatial-temporal transformer architecture utilizing spatial-temporal adaptive embeddings to enhance representation learning for traffic forecasting tasks. \url{https://github.com/XDZhelheim/STAEformer}
\item \textit{STID}~\citep{shao2022spatial}: \textit{STID} is a scalable model that addresses sample indistinguishability by introducing learnable spatial-temporal identity embeddings. These embeddings attach unique context (e.g., time-of-day, sensor ID) to the input, enabling the efficient capture of node-specific dynamics. \url{https://github.com/GestaltCogTeam/STID}
\item \textit{D2STGNN}~\citep{shao2022decoupled}: \textit{D2STGNN} is a graph neural network that decouples spatial and temporal dependencies to prevent signal mixing. It utilizes separate graph convolution and recurrent modules to independently model diffusion signals and inherent traffic patterns for precise spatio-temporal forecasting.\url{https://github.com/GestaltCogTeam/D2STGNN}

\item \textit{GWNet}~\citep{wu2019graph}: \textit{GWNet} is a spatial-temporal graph neural network that integrates diffusion graph convolutions with dilated causal temporal convolutions, utilizing a learnable self-adaptive adjacency matrix to capture latent spatial dependencies.
\url{https://github.com/nnzhan/Graph-WaveNet}
\end{itemize}

\subsubsection{Time Series Foundation Models}

\vspace{0.5em}
\begin{itemize}[noitemsep, topsep=8pt, partopsep=0pt, leftmargin=6mm,parsep=8pt]
\setlength\itemsep{0mm}
\item \textit{TimesFM++}~\citep{das2024timesfm}: 
\textit{TimesFM++} is a large-scale pre-trained decoder-only foundation model developed by Google Research, capable of delivering high-accuracy univariate forecasting across diverse domains and frequencies. For this study, we utilize the \textit{google/timesfm-2.0-500m-pytorch} checkpoint.
\url{https://github.com/google-research/timesfm}

\item \textit{Moirai}~\citep{woo2024moirai}: \textit{Moirai} is a large-scale pre-trained encoder-only time series foundation model developed by Salesforce AI Research. It is specifically designed to deliver universal forecasting capabilities across diverse domains, frequencies, and variable types. For this study, we utilize the \textit{moirai-1.0-R-base} checkpoint.
\url{https://github.com/SalesforceAIResearch/uni2ts}

\item \textit{TimeMoE}~\citep{shi2025time}: 
\textit{TimeMoE} is a scalable forecasting foundation model leveraging a sparse Mixture-of-Experts (MoE) architecture. 
By dynamically activating only a subset of experts for each prediction, it significantly reduces inference costs while maintaining high model capacity.
For this study, we utilize the \textit{Maple728/TimeMoE-50M} checkpoints.
\url{https://github.com/Time-MoE/Time-MoE}
\item 

\textit{Chronos}~\citep{ansari2024chronos}: 
\textit{Chronos} is a probabilistic time series foundation model that adapts the language modeling paradigm to the temporal domain. 
By tokenizing continuous values into a fixed vocabulary through scaling and quantization, it trains off-the-shelf Transformer architectures to learn universal temporal patterns from diverse datasets.
For this study, we utilize the \textit{amazon/chronos-bolt-base} checkpoints.
\url{https://github.com/amazon-science/chronos-forecasting}
\end{itemize}

\subsubsection{Spatio-Tempoal Foundation Models}

\vspace{0.5em}
\begin{itemize}[noitemsep, topsep=8pt, partopsep=0pt, leftmargin=6mm,parsep=8pt]
\setlength\itemsep{0mm}

\item \textit{OpenCity}~\citep{li2024opencity}: 
\textit{OpenCity} is a spatial-temporal foundation model that supports zero-shot and few-shot forecasting across diverse city-level applications.
For this study, we utilize the \textit{Opencity-plus.pth} checkpoints.
\url{https://github.com/HKUDS/OpenCity}

\item \textit{FactoST}~\citep{zhong2026st}: 
\textit{FactoST} is a factorized spatio-temporal foundation model that decouples universal temporal learning from domain-specific spatial adaptation, employing a minimalist temporal encoder and a lightweight spatial adapter to achieve efficient zero-shot and few-shot forecasting across diverse domains.
For this study, we utilize the \textit{factost\_utp2\_tiny\_4sta.pt} checkpoints. \url{https://github.com/CityMind-Lab/FactoST}
\end{itemize}

\subsubsection{Imputation Models}

\vspace{0.5em}
\begin{itemize}[noitemsep, topsep=8pt, partopsep=0pt, leftmargin=6mm,parsep=8pt]
\setlength\itemsep{0mm}

\item \textit{Mean}: 
\textit{Mean} is a baseline imputation method that fills missing entries using the node-level average of the observed data.

\item \textit{KNN}~\citep{crookston2008yaimpute}: 
\textit{KNN} performs imputation by averaging the values of the $k = 10$ neighboring nodes with the highest weight in the adjacency matrix. \url{https://github.com/iskandr/fancyimpute}

\item \textit{MICE}~\citep{van2011mice}: 
\textit{MICE} (Multivariate Imputation by Chained Equations) is a multiple imputation technique. For this study, the maximum number of iterations is limited to 100, and the number of nearest features is set to 10. \url{https://github.com/Graph-Machine-Learning-Group/grin/blob/main/scripts/run_baselines.py}

\item \textit{SVDImpute}~\citep{xu2017interpolating}: 
\textit{SVDImpute} employs Singular Value Decomposition (SVD) to estimate missing values by iteratively refining the low-rank approximation of the data matrix to capture global spatial-temporal correlations. \url{https://github.com/iskandr/fancyimpute}

\end{itemize}

\subsection{Details of Predefined Graph Construction}\label{appendix_predefined}

We follow conventional practices~\citep{li2018diffusion} to define the graph topology based on the spatial distribution of sensors. Depending on the data structure, we employ two distinct strategies to construct the adjacency matrix: a distance-based threshold Gaussian kernel for station-based data, and a grid-based topology for raster data.

\subsubsection{Distance-based Graph.} 

For irregularly distributed sensors (e.g., air quality and wind power datasets), we construct the adjacency matrix $A$ using a threshold Gaussian kernel. Let $d_{ij}$ denote the Haversine distance between sensor $i$ and sensor $j$. The weighted adjacency matrix is defined as:
$$
A_{ij} = 
\begin{cases} 
\exp\left(-\frac{d_{ij}^2}{\sigma^2}\right) & \text{if } \exp\left(-\frac{d_{ij}^2}{\sigma^2}\right) \geq r \quad \text{and} \quad i \neq j, \\ 
0 & \text{otherwise},
\end{cases}
$$
where $\sigma$ is the standard deviation of distances between all valid sensor pairs, and $r$ is the sparsity threshold. Empirically, we set $r$ to 0.5 for the air quality dataset and 0.99 for the wind power dataset.

\subsubsection{Grid-based Graph.} 

For datasets structured as a regular Euclidean grid (\eg, rasterized meteorological data), we define the graph topology using the Moore neighborhood (8-connectivity). Let $(u_i, v_i)$ represent the grid coordinates (longitude and latitude indices) of node $i$. The binary adjacency matrix is constructed based on the first-order spatial proximity:
$$
A_{ij} = 
\begin{cases} 
1 & \text{if } |u_i - u_j| \leq 1 \quad \text{and} \quad |v_i - v_j| \leq 1 \quad \text{and} \quad i \neq j, \\ 
0 & \text{otherwise}.
\end{cases}
$$
This ensures that each node is connected to its immediate spatial neighbors, including diagonals, while self-loops are excluded.

\subsection{Protocol Details}\label{appendix_protocol}

\subsubsection{Metrics Detail.}

We use different metrics such as MAE, RMSE, MRE, and MAPE. Formally, these metrics are formulated as follows:
\begin{align*}
\text{MAE} &= \frac{1}{n} \sum_{i=1}^n |y_i - \hat{y}_i|, &
\text{RMSE} &= \sqrt{\frac{1}{n} \sum_{i=1}^n (y_i - \hat{y}_i)^2}, &
\text{MRE} &= \frac{\sum_{i=1}^n |y_i - \hat{y}_i|}{\sum_{i=1}^n |y_i|}, &
\text{MAPE} &= \frac{100\%}{n} \sum_{i=1}^n \left| \frac{\hat{y}_i - y_i}{y_i} \right|,
\end{align*}
where $n$ represents the indices of all observed samples, $y_i$ denotes the $i$-th actual sample and $\hat{y}_i$ is the corresponding prediction.

\subsubsection{Parameter Detail.}

For all expert model baselines, hyper-parameters were set according to the recommendations of the corresponding papers and adjusted to achieve optimal results. The batch size was 128 by default; if it exceeded the GPU memory, it was halved until it was completely unusable. For fair comparison, all models were trained for 100 epochs by default, and an early stopping strategy was used to prevent overfitting; that is, if the loss on the validation set did not improve for 10 consecutive epochs, training was stopped. For the foundation model baselines, we used the corresponding public checkpoints for testing. In addition, the fine-tuning phase uniformly used 10 epochs of full parameter fine-tuning. All experiments were conducted on a Linux server equipped with (5 × 1) × Intel(R) Xeon(R) Gold 6248R CPU @ 3.00GHz (512GB memory) and (5 × 8) × NVIDIA A100 (80GB memory) GPUs. 

\section{More Experimental Results}
\label{appendix_more_exp}

\subsection{More Result on Performance Evaluation}
\label{appendix_more_rq1}

We provide complete experimental results on sensor-based and grid-based dataset benchmakr, including the zero-shot of the foundation model compared with the few-shot of the expert model, and the few-shot of the foundation model compared with the full-shot of the expert model, which are presented in Table~\ref{tab:zero_few_grid},~\ref{tab:zero_few_graph},~\ref{tab:few_full_graph}, and~\ref{tab:few_full_grid} respectively.

\begin{table*}[htbp!]
  \centering
  \caption{Comparison of the \textcolor{cyan}{\textit{\textbf{zero-shot}}} results of the foundation models with the \textcolor{cyan}{\textit{\textbf{few-shot}}} results of the expert models on the \textcolor{brown}{\textit{\textbf{short-term}}} ($12\rightarrow12_{average}$) and the \textcolor{brown}{\textit{\textbf{long-term}}} ($24\rightarrow24_{average}$) \textit{\textbf{forecasting}} based on the grid dataset benchmark. \protect\bluesquare{} indicates best,  \protect\greensquare{} indicates second best.}

\setlength{\tabcolsep}{1pt}
\renewcommand{\arraystretch}{1.6}
\resizebox{\linewidth}{!}{

\begin{tabular}{ccccccccccccccccccccccccccc}

\toprule

\multicolumn{2}{c}{} &   \quad\quad   & \multicolumn{9}{c}{\textbf{\emoji{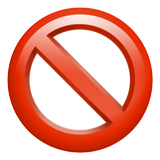} Zero-Shot of Foundation Model~~\emoji{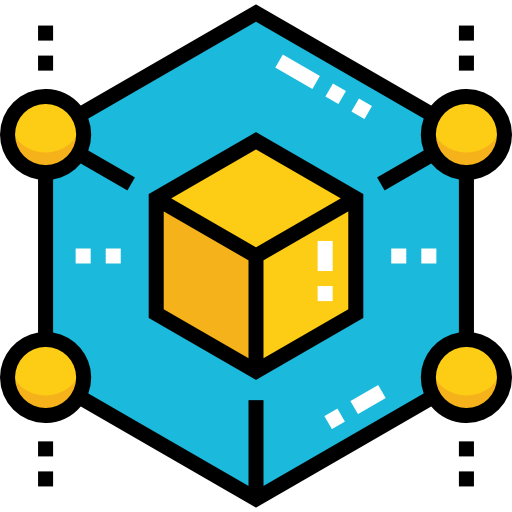}}} &  \quad\quad    & \multicolumn{14}{c}{\textbf{\emoji{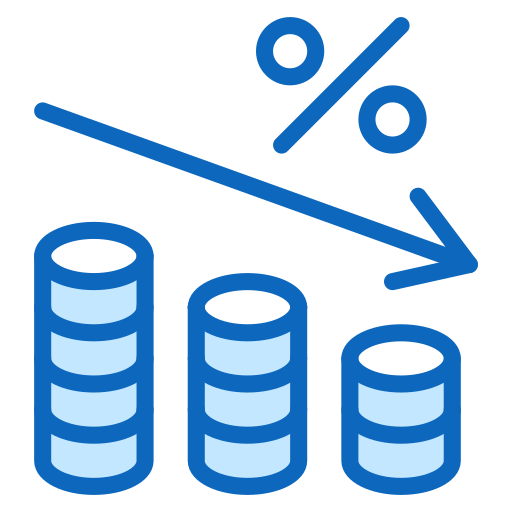} Few-Shot \textcolor{gray}{(10\% Training Set)} of Expert Model~~\emoji{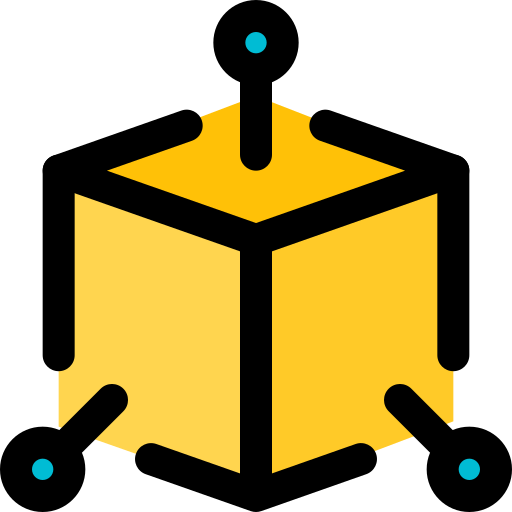}}} \\
\cmidrule{4-11}\cmidrule{13-27}

\multicolumn{3}{c}{\textbf{Method Type}} & \multicolumn{3}{c}{\textit{\textbf{\emoji{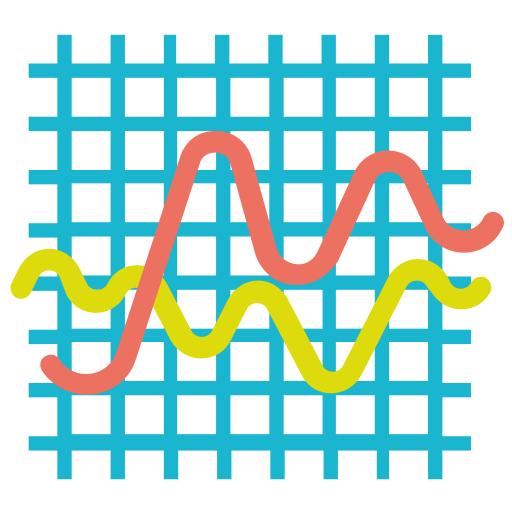} ST}}}  &  \quad\quad  & \multicolumn{4}{c}{\textit{\textbf{\emoji{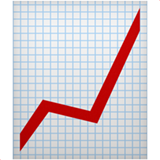} TS}}} &  \quad\quad   & \multicolumn{4}{c}{\textit{\textbf{\emoji{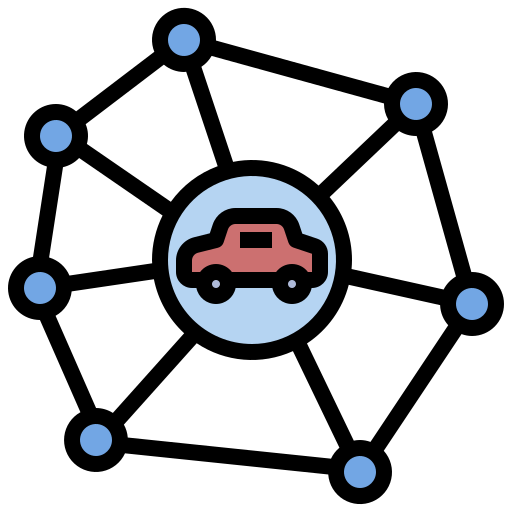} ST-Graph Method}}} &  \quad\quad  & \multicolumn{4}{c}{\textit{\textbf{\emoji{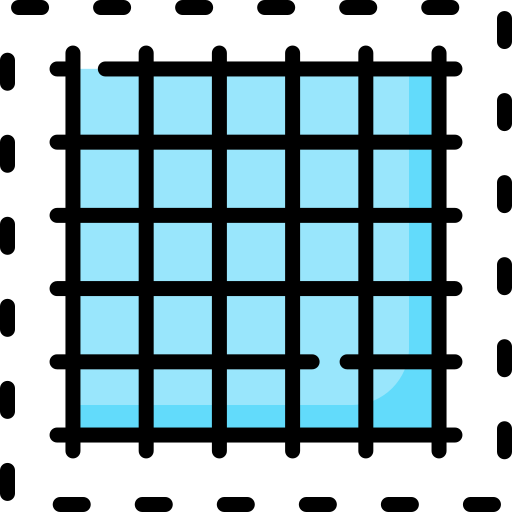} ST-Grid Method}}} &  \quad\quad & \multicolumn{4}{c}{\textit{\textbf{\emoji{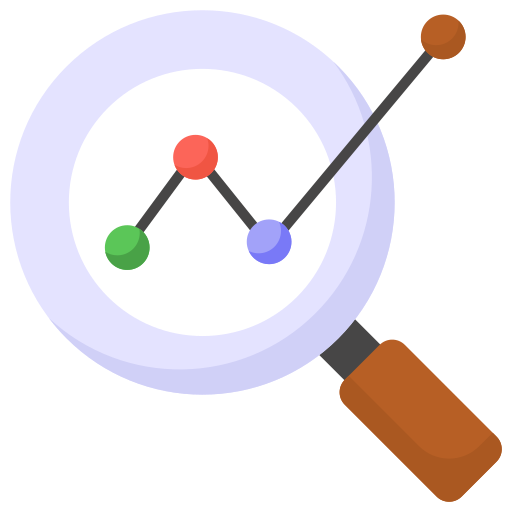} Time Series Method}}} \\

\cmidrule{4-6}\cmidrule{8-11}\cmidrule{13-16}\cmidrule{18-21}\cmidrule{23-27}

\multicolumn{3}{c}{\textbf{Method}} & \textbf{\model} &\textbf{FactoST} & \textbf{OpenCity} & & \textbf{TimesFM++} & \textbf{Moirai} &  \textbf{TimeMoE} & \textbf{Chronos} & & \textbf{STAEformer} & \textbf{STID} & \textbf{D2STGNN} & \textbf{GWNet} &  & \textbf{TAU} & \textbf{PredRNN++} & \textbf{DSAN} & \textbf{ST-ResNet} &  & \textbf{PatchTST} & \textbf{Dlinear} & \textbf{LSTM} & \textbf{HA} \\

\cmidrule{1-3}\cmidrule{4-6}\cmidrule{8-11}\cmidrule{13-16}\cmidrule{18-21}\cmidrule{23-27}

\multirow{6}{*}{Traffic-SH}

  & \multirow{3}{*}{Short}
& MAE & \textbf{0.44} & 0.40 & 0.45 & & 0.49 & 0.46 & 0.43 & 0.45 & 
& $1.26\textcolor{gray}{\text{\scriptsize±0.06}}$ & $0.47\textcolor{gray}{\text{\scriptsize±0.01}}$
& $0.39\textcolor{gray}{\text{\scriptsize±0.02}}$ & $0.37\textcolor{gray}{\text{\scriptsize±0.00}}$ &  
& $0.56\textcolor{gray}{\text{\scriptsize±0.01}}$ & $0.66\textcolor{gray}{\text{\scriptsize±0.01}}$
& $1.11\textcolor{gray}{\text{\scriptsize±0.00}}$ & $1.50\textcolor{gray}{\text{\scriptsize±0.00}}$ &  
& \cellcolor{cyan!8}$0.31\textcolor{gray}{\text{\scriptsize±0.00}}$ & $0.46\textcolor{gray}{\text{\scriptsize±0.00}}$
& \cellcolor{green!10}$0.36\textcolor{gray}{\text{\scriptsize±0.00}}$ & $0.53\textcolor{gray}{\text{\scriptsize±0.00}}$ \\

&
& RMSE & \cellcolor{cyan!8}\textbf{0.61} & 0.70 & 0.82  & & 1.14 & 0.91 & 0.74 & 0.80 & 
& $1.87\textcolor{gray}{\text{\scriptsize±0.08}}$ & $0.77\textcolor{gray}{\text{\scriptsize±0.02}}$
& \cellcolor{green!10}$0.69\textcolor{gray}{\text{\scriptsize±0.02}}$ & \cellcolor{green!10}$0.69\textcolor{gray}{\text{\scriptsize±0.01}}$ &  
& $0.98\textcolor{gray}{\text{\scriptsize±0.05}}$ & $1.02\textcolor{gray}{\text{\scriptsize±0.01}}$
& $1.84\textcolor{gray}{\text{\scriptsize±0.00}}$ & $2.11\textcolor{gray}{\text{\scriptsize±0.01}}$ &  
& \cellcolor{cyan!8}$0.61\textcolor{gray}{\text{\scriptsize±0.00}}$ & $0.79\textcolor{gray}{\text{\scriptsize±0.00}}$
& $0.71\textcolor{gray}{\text{\scriptsize±0.00}}$ & $0.92\textcolor{gray}{\text{\scriptsize±0.00}}$ \\

&
& MAPE(\%) & \cellcolor{cyan!8}\textbf{5.39} & 5.84 & 7.21 & & 7.31 & 8.76 & 6.65 & 6.77 & 
& $18.45\textcolor{gray}{\text{\scriptsize±0.12}}$ & $7.96\textcolor{gray}{\text{\scriptsize±0.14}}$
& $6.33\textcolor{gray}{\text{\scriptsize±0.10}}$ & $6.21\textcolor{gray}{\text{\scriptsize±0.06}}$ &  
& $9.41\textcolor{gray}{\text{\scriptsize±0.30}}$ & $11.03\textcolor{gray}{\text{\scriptsize±0.12}}$
& $10.74\textcolor{gray}{\text{\scriptsize±0.01}}$ & $17.16\textcolor{gray}{\text{\scriptsize±0.17}}$ &  
& \cellcolor{green!10}$5.47\textcolor{gray}{\text{\scriptsize±0.03}}$ & $8.18\textcolor{gray}{\text{\scriptsize±0.00}}$
& $6.44\textcolor{gray}{\text{\scriptsize±0.03}}$ & $7.76\textcolor{gray}{\text{\scriptsize±0.00}}$ \\

\cmidrule{3-27}

& \multirow{3}{*}{Long}
& MAE & \cellcolor{green!10}\textbf{0.49} & 0.58 & \cellcolor{cyan!8}0.48 &  & 0.65 & 0.66 & 0.60 & 0.67 & 
& $2.94\textcolor{gray}{\text{\scriptsize±0.52}}$ & $0.60\textcolor{gray}{\text{\scriptsize±0.01}}$
& $0.56\textcolor{gray}{\text{\scriptsize±0.02}}$ & $0.68\textcolor{gray}{\text{\scriptsize±0.01}}$ &  
& $0.70\textcolor{gray}{\text{\scriptsize±0.03}}$ & $0.80\textcolor{gray}{\text{\scriptsize±0.03}}$
& $1.21\textcolor{gray}{\text{\scriptsize±0.00}}$ & $1.54\textcolor{gray}{\text{\scriptsize±0.01}}$ &  
& \cellcolor{cyan!8}$0.48\textcolor{gray}{\text{\scriptsize±0.00}}$ & $0.59\textcolor{gray}{\text{\scriptsize±0.00}}$
& $0.51\textcolor{gray}{\text{\scriptsize±0.00}}$ & $0.53\textcolor{gray}{\text{\scriptsize±0.00}}$ \\

&
& RMSE & \cellcolor{cyan!8}\textbf{0.71} & 0.97 & 0.81 &  & 1.15 & 1.22 & 0.99 & 1.17 & 
& $3.82\textcolor{gray}{\text{\scriptsize±0.61}}$ & $0.95\textcolor{gray}{\text{\scriptsize±0.01}}$
& $0.90\textcolor{gray}{\text{\scriptsize±0.01}}$ & $1.14\textcolor{gray}{\text{\scriptsize±0.00}}$ &  
& $1.13\textcolor{gray}{\text{\scriptsize±0.02}}$ & $1.20\textcolor{gray}{\text{\scriptsize±0.04}}$
& $1.91\textcolor{gray}{\text{\scriptsize±0.00}}$ & $2.13\textcolor{gray}{\text{\scriptsize±0.01}}$ &  
& \cellcolor{green!10}$0.88\textcolor{gray}{\text{\scriptsize±0.00}}$ & $0.97\textcolor{gray}{\text{\scriptsize±0.00}}$
& $0.95\textcolor{gray}{\text{\scriptsize±0.01}}$ & $0.92\textcolor{gray}{\text{\scriptsize±0.00}}$ \\

&
& MAPE(\%) & \cellcolor{cyan!8}\textbf{6.23} & 9.30 & \cellcolor{green!10}7.41 &  & 10.16 & 12.72 & 9.57 & 9.99 & 
& $40.09\textcolor{gray}{\text{\scriptsize±7.07}}$ & $10.24\textcolor{gray}{\text{\scriptsize±0.05}}$
& $9.45\textcolor{gray}{\text{\scriptsize±0.29}}$ & $11.81\textcolor{gray}{\text{\scriptsize±0.17}}$ &  
& $11.25\textcolor{gray}{\text{\scriptsize±0.45}}$ & $13.49\textcolor{gray}{\text{\scriptsize±0.64}}$
& $12.29\textcolor{gray}{\text{\scriptsize±0.01}}$ & $17.98\textcolor{gray}{\text{\scriptsize±0.09}}$ &  
& $8.62\textcolor{gray}{\text{\scriptsize±0.00}}$ & $10.74\textcolor{gray}{\text{\scriptsize±0.01}}$
& $9.58\textcolor{gray}{\text{\scriptsize±0.08}}$ & $7.76\textcolor{gray}{\text{\scriptsize±0.00}}$ \\

\midrule

\multirow{6}{*}{Bike-NYC}

& \multirow{3}{*}{Short}
& MAE & \textbf{2.12} &  3.77 & 4.67 &   & 3.98 & 1.63 & 3.90 & 4.08 & 
& $1.52\textcolor{gray}{\text{\scriptsize±0.17}}$ & $1.12\textcolor{gray}{\text{\scriptsize±0.03}}$
& $1.06\textcolor{gray}{\text{\scriptsize±0.02}}$ & \cellcolor{cyan!8}$0.94\textcolor{gray}{\text{\scriptsize±0.03}}$ &  
& $1.28\textcolor{gray}{\text{\scriptsize±0.02}}$ & $1.79\textcolor{gray}{\text{\scriptsize±0.02}}$
& $3.12\textcolor{gray}{\text{\scriptsize±0.00}}$ & $3.74\textcolor{gray}{\text{\scriptsize±0.02}}$ &  
& \cellcolor{green!10}$0.99\textcolor{gray}{\text{\scriptsize±0.01}}$ & $1.18\textcolor{gray}{\text{\scriptsize±0.00}}$
& $1.02\textcolor{gray}{\text{\scriptsize±0.01}}$ & $5.22\textcolor{gray}{\text{\scriptsize±0.00}}$ \\

&
& RMSE & \textbf{3.64} &  7.11 & 11.03 &  & 7.37 & 5.87 & 8.08 & 7.64 & 
& $5.62\textcolor{gray}{\text{\scriptsize±0.62}}$ & \cellcolor{green!10}$3.39\textcolor{gray}{\text{\scriptsize±0.08}}$
& $3.90\textcolor{gray}{\text{\scriptsize±0.14}}$ & \cellcolor{cyan!8}$3.32\textcolor{gray}{\text{\scriptsize±0.08}}$ &  
& $3.93\textcolor{gray}{\text{\scriptsize±0.12}}$ & $7.23\textcolor{gray}{\text{\scriptsize±0.21}}$
& $12.26\textcolor{gray}{\text{\scriptsize±0.00}}$ & $12.46\textcolor{gray}{\text{\scriptsize±0.01}}$ &  
& $3.40\textcolor{gray}{\text{\scriptsize±0.02}}$ & $3.83\textcolor{gray}{\text{\scriptsize±0.02}}$
& $3.77\textcolor{gray}{\text{\scriptsize±0.04}}$ & $10.80\textcolor{gray}{\text{\scriptsize±0.00}}$ \\

&
& MAPE(\%) & \cellcolor{cyan!8}\textbf{24.39} &  54.36 & 73.36&  & 65.52 & 85.18 & 64.20 & 57.62 &  
& $57.83\textcolor{gray}{\text{\scriptsize±4.43}}$ & $46.32\textcolor{gray}{\text{\scriptsize±1.29}}$
& $46.61\textcolor{gray}{\text{\scriptsize±1.85}}$ & \cellcolor{green!10}$41.82\textcolor{gray}{\text{\scriptsize±1.37}}$ &  
& $55.65\textcolor{gray}{\text{\scriptsize±1.31}}$ & $61.26\textcolor{gray}{\text{\scriptsize±1.12}}$
& $70.80\textcolor{gray}{\text{\scriptsize±0.16}}$ & $83.24\textcolor{gray}{\text{\scriptsize±2.34}}$ &  
& $43.76\textcolor{gray}{\text{\scriptsize±0.43}}$ & $52.45\textcolor{gray}{\text{\scriptsize±0.20}}$
& $46.36\textcolor{gray}{\text{\scriptsize±0.39}}$ & $67.98\textcolor{gray}{\text{\scriptsize±0.00}}$ \\

\cmidrule{3-27}

& \multirow{3}{*}{Long}
& MAE & \textbf{2.81} & 5.89 & 5.17 & & 5.74 & 2.67 & 5.52 & 6.17 & 
& $2.44\textcolor{gray}{\text{\scriptsize±0.38}}$ & \cellcolor{green!10}$1.59\textcolor{gray}{\text{\scriptsize±0.01}}$
& \cellcolor{cyan!8}$1.44\textcolor{gray}{\text{\scriptsize±0.01}}$ & $1.71\textcolor{gray}{\text{\scriptsize±0.03}}$ &  
& $1.60\textcolor{gray}{\text{\scriptsize±0.00}}$ & $1.93\textcolor{gray}{\text{\scriptsize±0.04}}$
& $3.25\textcolor{gray}{\text{\scriptsize±0.00}}$ & $3.85\textcolor{gray}{\text{\scriptsize±0.08}}$ &  
& $1.69\textcolor{gray}{\text{\scriptsize±0.01}}$ & $1.77\textcolor{gray}{\text{\scriptsize±0.00}}$
& $1.63\textcolor{gray}{\text{\scriptsize±0.07}}$ & $5.18\textcolor{gray}{\text{\scriptsize±0.00}}$ \\

&
& RMSE & \cellcolor{green!10}\textbf{5.13} & 11.21 & 11.79 &  & 11.17 & 9.78 & 11.33 & 12.19 & 
& $7.97\textcolor{gray}{\text{\scriptsize±0.72}}$ & $5.39\textcolor{gray}{\text{\scriptsize±0.06}}$
& $5.41\textcolor{gray}{\text{\scriptsize±0.03}}$ & $6.29\textcolor{gray}{\text{\scriptsize±0.06}}$ &  
& \cellcolor{cyan!8}$5.08\textcolor{gray}{\text{\scriptsize±0.09}}$ & $7.43\textcolor{gray}{\text{\scriptsize±0.17}}$
& $12.39\textcolor{gray}{\text{\scriptsize±0.00}}$ & $12.53\textcolor{gray}{\text{\scriptsize±0.01}}$ &  
& $5.96\textcolor{gray}{\text{\scriptsize±0.07}}$ & $6.06\textcolor{gray}{\text{\scriptsize±0.00}}$
& $5.96\textcolor{gray}{\text{\scriptsize±0.14}}$ & $10.77\textcolor{gray}{\text{\scriptsize±0.00}}$ \\

&
& MAPE(\%) & \cellcolor{cyan!8}\textbf{29.53} & 75.63 & 85.08&  & 92.49 & 134.82 & 83.27 & 87.06 & 
& $99.60\textcolor{gray}{\text{\scriptsize±21.14}}$ & $56.47\textcolor{gray}{\text{\scriptsize±0.59}}$
& \cellcolor{green!10}$55.85\textcolor{gray}{\text{\scriptsize±0.62}}$ & $66.85\textcolor{gray}{\text{\scriptsize±0.48}}$ &  
& $63.65\textcolor{gray}{\text{\scriptsize±0.75}}$ & $68.35\textcolor{gray}{\text{\scriptsize±0.86}}$
& $80.26\textcolor{gray}{\text{\scriptsize±0.08}}$ & $86.39\textcolor{gray}{\text{\scriptsize±3.40}}$ &  
& $65.15\textcolor{gray}{\text{\scriptsize±0.60}}$ & $70.93\textcolor{gray}{\text{\scriptsize±0.40}}$
& $65.76\textcolor{gray}{\text{\scriptsize±4.13}}$ & $67.38\textcolor{gray}{\text{\scriptsize±0.00}}$ \\

\midrule

\multirow{6}{*}{Taxi-NYC}

& \multirow{3}{*}{Short}
& MAE & \textbf{5.60} &  8.84 & 12.23 &  & 8.95 & 7.58 & 10.05 & 9.48 & 
& $8.60\textcolor{gray}{\text{\scriptsize±0.54}}$ & $6.18\textcolor{gray}{\text{\scriptsize±0.41}}$
& $5.34\textcolor{gray}{\text{\scriptsize±0.21}}$ & $5.04\textcolor{gray}{\text{\scriptsize±0.09}}$ &  
& $6.59\textcolor{gray}{\text{\scriptsize±0.12}}$ & $10.80\textcolor{gray}{\text{\scriptsize±0.17}}$
& $22.56\textcolor{gray}{\text{\scriptsize±0.01}}$ & $24.04\textcolor{gray}{\text{\scriptsize±0.05}}$ &  
& \cellcolor{cyan!8}$4.08\textcolor{gray}{\text{\scriptsize±0.01}}$ & \cellcolor{green!10}$4.77\textcolor{gray}{\text{\scriptsize±0.00}}$
& $4.86\textcolor{gray}{\text{\scriptsize±0.06}}$ & $18.83\textcolor{gray}{\text{\scriptsize±0.00}}$ \\

&
& RMSE & \cellcolor{cyan!8}\textbf{11.65} &  21.68 & 41.34  & & 20.04 & 22.72 & 23.59 & 21.69 & 
& $27.34\textcolor{gray}{\text{\scriptsize±1.58}}$ & $15.29\textcolor{gray}{\text{\scriptsize±0.40}}$
& $16.17\textcolor{gray}{\text{\scriptsize±0.23}}$ & $16.67\textcolor{gray}{\text{\scriptsize±0.06}}$ &  
& $19.01\textcolor{gray}{\text{\scriptsize±0.59}}$ & $35.48\textcolor{gray}{\text{\scriptsize±0.76}}$
& $78.95\textcolor{gray}{\text{\scriptsize±0.00}}$ & $79.20\textcolor{gray}{\text{\scriptsize±0.03}}$ &  
& \cellcolor{green!10}$12.34\textcolor{gray}{\text{\scriptsize±0.04}}$ & $13.66\textcolor{gray}{\text{\scriptsize±0.01}}$
& $15.85\textcolor{gray}{\text{\scriptsize±0.21}}$ & $52.17\textcolor{gray}{\text{\scriptsize±0.00}}$ \\

&
& MAPE(\%) & \cellcolor{cyan!8}\textbf{29.55} & 42.33 & 53.82  &   & 49.50 & 63.33 & 89.85 & 43.22 & 
& $61.05\textcolor{gray}{\text{\scriptsize±2.76}}$ & $106.73\textcolor{gray}{\text{\scriptsize±25.20}}$
& $49.62\textcolor{gray}{\text{\scriptsize±8.52}}$ & $44.47\textcolor{gray}{\text{\scriptsize±2.06}}$ &  
& $68.53\textcolor{gray}{\text{\scriptsize±0.78}}$ & $74.84\textcolor{gray}{\text{\scriptsize±5.67}}$
& $55.68\textcolor{gray}{\text{\scriptsize±0.23}}$ & $122.35\textcolor{gray}{\text{\scriptsize±5.68}}$ &  
& \cellcolor{green!10}$36.24\textcolor{gray}{\text{\scriptsize±0.28}}$ & $39.47\textcolor{gray}{\text{\scriptsize±0.28}}$
& $39.96\textcolor{gray}{\text{\scriptsize±0.32}}$ & $72.60\textcolor{gray}{\text{\scriptsize±0.00}}$ \\

\cmidrule{3-27}

& \multirow{3}{*}{Long}
& MAE & \cellcolor{cyan!8}\textbf{7.74} &  13.98 & 13.50 &  & 14.44 & 13.04 & 15.21 & 14.12 & 
& $12.81\textcolor{gray}{\text{\scriptsize±0.28}}$ & $8.30\textcolor{gray}{\text{\scriptsize±0.11}}$
& $7.20\textcolor{gray}{\text{\scriptsize±0.08}}$ & $10.41\textcolor{gray}{\text{\scriptsize±0.18}}$ &  
& $9.01\textcolor{gray}{\text{\scriptsize±0.23}}$ & $12.67\textcolor{gray}{\text{\scriptsize±0.18}}$
& $23.65\textcolor{gray}{\text{\scriptsize±0.01}}$ & $24.99\textcolor{gray}{\text{\scriptsize±0.08}}$ &  
& \cellcolor{green!10}$7.70\textcolor{gray}{\text{\scriptsize±0.02}}$ & $7.83\textcolor{gray}{\text{\scriptsize±0.00}}$
& $8.24\textcolor{gray}{\text{\scriptsize±0.09}}$ & $18.79\textcolor{gray}{\text{\scriptsize±0.00}}$ \\

&
& RMSE & \cellcolor{cyan!8}\textbf{16.33} & 33.21 &40.83 &   & 33.44 & 40.13 & 37.26 & 33.02 & 
& $40.60\textcolor{gray}{\text{\scriptsize±1.45}}$ & \cellcolor{green!10}$23.44\textcolor{gray}{\text{\scriptsize±0.25}}$
& $24.28\textcolor{gray}{\text{\scriptsize±0.43}}$ & $32.32\textcolor{gray}{\text{\scriptsize±0.48}}$ &  
& $26.81\textcolor{gray}{\text{\scriptsize±0.56}}$ & $39.88\textcolor{gray}{\text{\scriptsize±0.99}}$
& $79.90\textcolor{gray}{\text{\scriptsize±0.01}}$ & $80.07\textcolor{gray}{\text{\scriptsize±0.03}}$ &  
& $23.65\textcolor{gray}{\text{\scriptsize±0.12}}$ & $23.67\textcolor{gray}{\text{\scriptsize±0.01}}$
& $26.36\textcolor{gray}{\text{\scriptsize±0.15}}$ & $52.16\textcolor{gray}{\text{\scriptsize±0.00}}$ \\

&
& MAPE(\%) & \cellcolor{cyan!8}\textbf{40.77} & 54.49 & 60.91 &  & 66.83 & 96.72 & 94.20 & 63.58 & 
& $80.98\textcolor{gray}{\text{\scriptsize±8.31}}$ & $103.69\textcolor{gray}{\text{\scriptsize±11.22}}$
& \cellcolor{green!10}$50.48\textcolor{gray}{\text{\scriptsize±3.09}}$ & $62.74\textcolor{gray}{\text{\scriptsize±2.21}}$ &  
& $83.72\textcolor{gray}{\text{\scriptsize±1.15}}$ & $80.96\textcolor{gray}{\text{\scriptsize±8.23}}$
& $66.39\textcolor{gray}{\text{\scriptsize±0.26}}$ & $137.27\textcolor{gray}{\text{\scriptsize±2.10}}$ &  
& $52.54\textcolor{gray}{\text{\scriptsize±0.38}}$ & $52.19\textcolor{gray}{\text{\scriptsize±0.55}}$
& $51.01\textcolor{gray}{\text{\scriptsize±0.43}}$ & $72.64\textcolor{gray}{\text{\scriptsize±0.00}}$ \\

\midrule

\multirow{6}{*}{Tdrive-BJ}

& \multirow{3}{*}{Short}
& MAE & \textbf{9.57} & 17.26 & 24.11 &  & 13.90 & 21.73 & 21.40 & 16.91 & 
& \cellcolor{green!10}$8.21\textcolor{gray}{\text{\scriptsize±0.30}}$ & $9.63\textcolor{gray}{\text{\scriptsize±0.37}}$
& $8.89\textcolor{gray}{\text{\scriptsize±0.35}}$ & $11.16\textcolor{gray}{\text{\scriptsize±2.24}}$ &  
& $16.16\textcolor{gray}{\text{\scriptsize±0.15}}$ & $17.29\textcolor{gray}{\text{\scriptsize±0.18}}$
& $44.18\textcolor{gray}{\text{\scriptsize±0.03}}$ & $58.21\textcolor{gray}{\text{\scriptsize±0.27}}$ &  
& \cellcolor{cyan!8}$7.12\textcolor{gray}{\text{\scriptsize±0.12}}$ & $10.59\textcolor{gray}{\text{\scriptsize±0.01}}$
& $9.35\textcolor{gray}{\text{\scriptsize±0.69}}$ & $31.67\textcolor{gray}{\text{\scriptsize±0.00}}$ \\

&
& RMSE & \cellcolor{cyan!8}\textbf{16.69} & 35.23 & 71.15 &  & 30.48 & 52.16 & 36.67 & 35.01 & 
& \cellcolor{green!10}$18.54\textcolor{gray}{\text{\scriptsize±0.62}}$ & $20.66\textcolor{gray}{\text{\scriptsize±1.26}}$
& $18.62\textcolor{gray}{\text{\scriptsize±0.70}}$ & $22.23\textcolor{gray}{\text{\scriptsize±2.17}}$ &  
& $30.79\textcolor{gray}{\text{\scriptsize±0.68}}$ & $33.61\textcolor{gray}{\text{\scriptsize±0.75}}$
& $131.72\textcolor{gray}{\text{\scriptsize±0.01}}$ & $139.91\textcolor{gray}{\text{\scriptsize±0.12}}$ &  
& $19.80\textcolor{gray}{\text{\scriptsize±0.29}}$ & $23.70\textcolor{gray}{\text{\scriptsize±0.02}}$
& $21.58\textcolor{gray}{\text{\scriptsize±1.06}}$ & $63.53\textcolor{gray}{\text{\scriptsize±0.00}}$ \\

&
& MAPE(\%) & \textbf{15.92} & 20.61 & 25.17 &  & 18.26 & 31.99 &  30.76 & 20.79 & 
& \cellcolor{green!10}$14.67\textcolor{gray}{\text{\scriptsize±0.61}}$ & $21.53\textcolor{gray}{\text{\scriptsize±2.61}}$
& $23.10\textcolor{gray}{\text{\scriptsize±4.61}}$ & $55.30\textcolor{gray}{\text{\scriptsize±34.98}}$ &  
& $59.91\textcolor{gray}{\text{\scriptsize±1.07}}$ & $30.97\textcolor{gray}{\text{\scriptsize±2.13}}$
& $76.56\textcolor{gray}{\text{\scriptsize±2.54}}$ & $123.20\textcolor{gray}{\text{\scriptsize±5.41}}$ &  
& \cellcolor{cyan!8}$11.08\textcolor{gray}{\text{\scriptsize±0.25}}$ & $19.97\textcolor{gray}{\text{\scriptsize±0.26}}$
& $19.80\textcolor{gray}{\text{\scriptsize±2.12}}$ & $35.44\textcolor{gray}{\text{\scriptsize±0.00}}$ \\

\cmidrule{3-27}

& \multirow{3}{*}{Long}
& MAE & \cellcolor{cyan!8}\textbf{12.92} & 26.28 & 50.05 &  & 29.89 & 38.79 & 30.94 & 27.54 & 
& $20.48\textcolor{gray}{\text{\scriptsize±0.41}}$ & $19.56\textcolor{gray}{\text{\scriptsize±0.08}}$
& \cellcolor{green!10}$15.52\textcolor{gray}{\text{\scriptsize±0.40}}$ & $22.89\textcolor{gray}{\text{\scriptsize±0.31}}$ &  
& $26.23\textcolor{gray}{\text{\scriptsize±1.05}}$ & $25.33\textcolor{gray}{\text{\scriptsize±0.51}}$
& $49.94\textcolor{gray}{\text{\scriptsize±0.02}}$ & $60.27\textcolor{gray}{\text{\scriptsize±0.19}}$ &  
& $18.52\textcolor{gray}{\text{\scriptsize±0.08}}$ & $21.72\textcolor{gray}{\text{\scriptsize±0.04}}$
& $16.44\textcolor{gray}{\text{\scriptsize±0.18}}$ & $31.69\textcolor{gray}{\text{\scriptsize±0.00}}$ \\

&
& RMSE & \cellcolor{cyan!8}\textbf{23.54} & 55.77 & 69.42 &  & 61.15 & 89.24 & 55.75 & 57.26 & 
& $43.94\textcolor{gray}{\text{\scriptsize±1.62}}$ & $44.76\textcolor{gray}{\text{\scriptsize±1.22}}$
& \cellcolor{green!10}$32.02\textcolor{gray}{\text{\scriptsize±0.97}}$ & $45.60\textcolor{gray}{\text{\scriptsize±0.28}}$ &  
& $56.30\textcolor{gray}{\text{\scriptsize±4.81}}$ & $50.52\textcolor{gray}{\text{\scriptsize±0.85}}$
& $134.92\textcolor{gray}{\text{\scriptsize±0.01}}$ & $140.68\textcolor{gray}{\text{\scriptsize±0.17}}$ &  
& $43.34\textcolor{gray}{\text{\scriptsize±0.06}}$ & $45.11\textcolor{gray}{\text{\scriptsize±0.08}}$
& $35.94\textcolor{gray}{\text{\scriptsize±0.43}}$ & $63.58\textcolor{gray}{\text{\scriptsize±0.00}}$ \\

&
& MAPE(\%) & \cellcolor{cyan!8}\textbf{20.62} & 31.31 & 30.96 &   & 40.51 & 56.76 & 57.15 & 34.69 & 
& $39.76\textcolor{gray}{\text{\scriptsize±20.85}}$ & $30.34\textcolor{gray}{\text{\scriptsize±2.96}}$
& \cellcolor{green!10}$25.10\textcolor{gray}{\text{\scriptsize±1.67}}$ & $53.90\textcolor{gray}{\text{\scriptsize±6.60}}$ &  
& $86.47\textcolor{gray}{\text{\scriptsize±1.45}}$ & $39.13\textcolor{gray}{\text{\scriptsize±2.59}}$
& $82.85\textcolor{gray}{\text{\scriptsize±1.27}}$ & $128.38\textcolor{gray}{\text{\scriptsize±1.49}}$ &  
& $25.15\textcolor{gray}{\text{\scriptsize±0.01}}$ & $35.17\textcolor{gray}{\text{\scriptsize±0.90}}$
& $27.40\textcolor{gray}{\text{\scriptsize±1.50}}$ & $35.49\textcolor{gray}{\text{\scriptsize±0.00}}$ \\

\midrule

\rowc

\multicolumn{3}{c}{\textbf{$\bf 1^{st}$ count}} 
& \textbf{15} & 0 & 1 &  & 0 & 0 &0 & 0 & & 0 & 0 & 1 & 2 & & 1 & 0 & 0  & 0  & & 6 & 0 & 0  & 0\\

\bottomrule

\end{tabular}%

 }
\label{tab:zero_few_grid}
\end{table*}

\begin{table*}[htbp!]
  \centering
  \caption{Comparison of the \textcolor{cyan}{\textit{\textbf{zero-shot}}} results of the foundation models with the \textcolor{cyan}{\textit{\textbf{few-shot}}} results of the expert models on the \textcolor{brown}{\textit{\textbf{short-term}}} ($12\rightarrow12_{average}$) and the \textcolor{brown}{\textit{\textbf{long-term}}} ($24\rightarrow24_{average}$) \textit{\textbf{forecasting}} based on the \textit{\textbf{sensor}} dataset benchmark. 
  \protect\bluesquare{} indicates best,  \protect\greensquare{} indicates second best.
  OOM indicates Out-Of-Memory (NVIDIA A100-80G).}

\setlength{\tabcolsep}{1pt}
\renewcommand{\arraystretch}{1.6}
\resizebox{\linewidth}{!}{

\begin{tabular}{ccccccccccccccccccccc}

\toprule

\multicolumn{2}{c}{} &   \quad\quad   & \multicolumn{8}{c}{\textbf{\emoji{figure/emoji/em-zero_shot} Zero-Shot of Foundation Model~~\emoji{figure/emoji/em-foundation_model}}} &  \quad\quad    & \multicolumn{9}{c}{\textbf{\emoji{figure/emoji/em-few_shot} Few-Shot \textcolor{gray}{(10\% Training Set)} of Expert Model~~\emoji{figure/emoji/em-expert_model}}} \\
\cmidrule{4-11}\cmidrule{13-21}

\multicolumn{3}{c}{\textbf{Method Type}} & \multicolumn{3}{c}{\textit{\textbf{\emoji{figure/emoji/em_st_foundation} ST}}}  &  \quad\quad  & \multicolumn{4}{c}{\textit{\textbf{\emoji{figure/emoji/em_ts_foundation} TS}}} &  \quad\quad   & \multicolumn{4}{c}{\textit{\textbf{\emoji{figure/emoji/em-st_graph} ST-Graph Method}}} &  \quad\quad & \multicolumn{4}{c}{\textit{\textbf{\emoji{figure/emoji/em-ts_expert} Time Series Method}}} \\

\cmidrule{4-6}\cmidrule{8-11}\cmidrule{13-16}\cmidrule{18-21}

\multicolumn{3}{c}{\textbf{Model}} & \textbf{\model} & \textbf{FactoST} & \textbf{OpenCity} &  \quad\quad & \textbf{TimesFM++} & \textbf{Moirai} & \textbf{TimeMoE} & \textbf{Chronos} & &\textbf{STAEformer} & \textbf{STID} & \textbf{D2STGNN} & \textbf{GWNet} &  & \textbf{PatchTST} & \textbf{Dlinear} & \textbf{LSTM} & \textbf{HA} \\

\cmidrule{1-3}\cmidrule{4-6}\cmidrule{8-11}\cmidrule{13-16}\cmidrule{18-21}

\multirow{6}{*}{PEMS-03}
& \multirow{3}{*}{Short}
& MAE & \textbf{\cellcolor{cyan!8}{15.62}} & 25.84& 25.57 &  & 27.13 & 29.02 & 29.12 & 26.48 &
& $20.07\textcolor{gray}{\text{\scriptsize±0.39}}$ & $17.88\textcolor{gray}{\text{\scriptsize±0.12}}$
& \cellcolor{green!10}$17.78\textcolor{gray}{\text{\scriptsize±0.17}}$ & $17.96\textcolor{gray}{\text{\scriptsize±0.68}}$ &   
& $21.04\textcolor{gray}{\text{\scriptsize±0.07}}$ & $21.17\textcolor{gray}{\text{\scriptsize±0.00}}$
& $19.08\textcolor{gray}{\text{\scriptsize±0.38}}$ & $32.01\textcolor{gray}{\text{\scriptsize±0.00}}$ \\
\bottomrule

\end{tabular}%

 }
\label{tab:zero_few_graph}
\end{table*}

\begin{table*}[htbp!]
  \centering
  \caption{Comparison of the \textcolor{cyan}{\textit{\textbf{few-shot}}} results of the foundation models with the \textcolor{cyan}{\textit{\textbf{full-shot}}} results of the expert models on the \textcolor{brown}{\textit{\textbf{short-term}}} ($12\rightarrow12_{average}$) and the \textcolor{brown}{\textit{\textbf{long-term}}} ($24\rightarrow24_{average}$) \textit{\textbf{forecasting}} based on the sensor dataset benchmark. \protect\bluesquare{} indicates best,  \protect\greensquare{} indicates second best.}

\vspace{-4mm}

\setlength{\tabcolsep}{1pt}
\renewcommand{\arraystretch}{1.6}
\resizebox{\linewidth}{!}{

\begin{tabular}{ccccccccccccccccccccc}

\toprule

\multicolumn{2}{c}{} &   \quad\quad   & \multicolumn{8}{c}{\textbf{\emoji{figure/emoji/em-few_shot} Few-Shot \textcolor{gray}{(10\% Training Set)} of Foundation Model~~\emoji{figure/emoji/em-foundation_model}}} &  \quad\quad    & \multicolumn{9}{c}{\textbf{\emoji{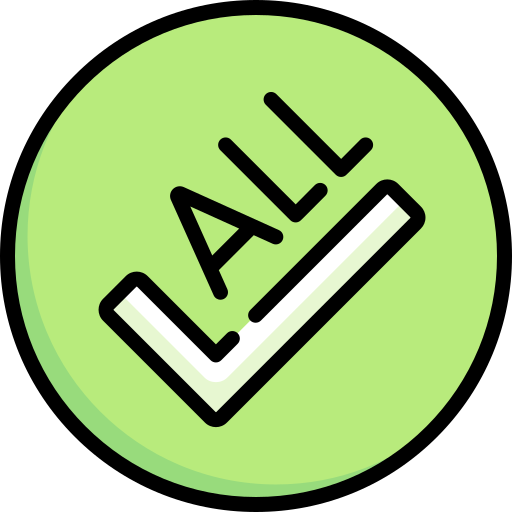} Full-Shot of Expert Model~~\emoji{figure/emoji/em-expert_model}}} \\
\cmidrule{4-11}\cmidrule{13-21}

\multicolumn{3}{c}{\textbf{Method Type}} & \multicolumn{3}{c}{\textit{\textbf{\emoji{figure/emoji/em_st_foundation} ST}}}  &  \quad\quad  & \multicolumn{4}{c}{\textit{\textbf{\emoji{figure/emoji/em_ts_foundation} TS}}} &  \quad\quad   & \multicolumn{4}{c}{\textit{\textbf{\emoji{figure/emoji/em-st_graph} ST-Graph Method}}} &  \quad\quad & \multicolumn{4}{c}{\textit{\textbf{\emoji{figure/emoji/em-ts_expert} Time Series Method}}} \\

\cmidrule{4-6}\cmidrule{8-11}\cmidrule{13-16}\cmidrule{18-21}

\multicolumn{3}{c}{\textbf{Model}} & \textbf{\model} & \textbf{FactoST} & \textbf{OpenCity} &  \quad\quad & \textbf{TimesFM++} & \textbf{Moirai} & \textbf{TimeMoE} & \textbf{Chronos} & &\textbf{STAEformer} & \textbf{STID} & \textbf{D2STGNN} & \textbf{GWNet} &  & \textbf{PatchTST} & \textbf{Dlinear} & \textbf{LSTM} & \textbf{HA} \\

\cmidrule{1-3}\cmidrule{4-6}\cmidrule{8-11}\cmidrule{13-16}\cmidrule{18-21}

\multirow{6}{*}{PEMS-03}

& \multirow{3}{*}{Short}
& MAE 
& \cellcolor{cyan!8}\textbf{13.00} & 21.08  & 16.91 &  & 21.56 & 20.54  & 20.59  & 21.31  &
& \cellcolor{green!10}$15.13\textcolor{gray}{\text{\scriptsize±0.10}}$ & $15.21\textcolor{gray}{\text{\scriptsize±0.03}}$
& $14.88\textcolor{gray}{\text{\scriptsize±0.11}}$ & $14.91\textcolor{gray}{\text{\scriptsize±0.12}}$ &   
& $20.84\textcolor{gray}{\text{\scriptsize±0.01}}$ & $21.04\textcolor{gray}{\text{\scriptsize±0.00}}$
& $16.69\textcolor{gray}{\text{\scriptsize±0.06}}$ & $26.17\textcolor{gray}{\text{\scriptsize±0.00}}$ \\
\bottomrule

\end{tabular}%

 }
\label{tab:few_full_graph}
\end{table*}

\begin{table*}[htbp!]
  \centering
  \caption{Comparison of the \textcolor{cyan}{\textit{\textbf{few-shot}}} results of the foundation models with the \textcolor{cyan}{\textit{\textbf{full-shot}}} results of the expert models on the \textcolor{brown}{\textit{\textbf{short-term}}} ($12\rightarrow12_{average}$) and the \textcolor{brown}{\textit{\textbf{long-term}}} ($24\rightarrow24_{average}$) \textit{\textbf{forecasting}} based on the grid dataset benchmark. \protect\bluesquare{} indicates best,  \protect\greensquare{} indicates second best.}


\setlength{\tabcolsep}{1pt}
\renewcommand{\arraystretch}{1.6}
\resizebox{\linewidth}{!}{

\begin{tabular}{cccccccccccccccccccccccccccc}

\toprule

\multicolumn{2}{c}{} &   \quad\quad   & \multicolumn{9}{c}{\textbf{\emoji{figure/emoji/em-few_shot} Few-Shot \textcolor{gray}{(10\% Training Set)} of Foundation Model~~\emoji{figure/emoji/em-foundation_model}}} &  \quad\quad    & \multicolumn{14}{c}{\textbf{\emoji{figure/emoji/em-full_shot} Full-Shot of Expert Model~~\emoji{figure/emoji/em-expert_model}}} \\
\cmidrule{4-11}\cmidrule{13-26}

\multicolumn{3}{c}{\textbf{Method Type}} & \multicolumn{3}{c}{\textit{\textbf{\emoji{figure/emoji/em_st_foundation} ST}}}  &  \quad\quad  & \multicolumn{4}{c}{\textit{\textbf{\emoji{figure/emoji/em_ts_foundation} TS}}} &  \quad\quad   & \multicolumn{4}{c}{\textit{\textbf{\emoji{figure/emoji/em-st_graph} ST-Graph Method}}} &  \quad\quad  & \multicolumn{4}{c}{\textit{\textbf{\emoji{figure/emoji/em-st_grid} ST-Grid Method}}} &  \quad\quad & \multicolumn{4}{c}{\textit{\textbf{\emoji{figure/emoji/em-ts_expert} Time Series Method}}} \\

\cmidrule{4-6}\cmidrule{8-11}\cmidrule{13-16}\cmidrule{18-21}\cmidrule{23-27}

\multicolumn{3}{c}{\textbf{Method}} & \textbf{\model} &\textbf{FactoST} & \textbf{OpenCity} &  & \textbf{TimesFM++} & \textbf{Moirai} &  \textbf{TimeMoE} & \textbf{Chronos} & & \textbf{STAEformer} & \textbf{STID} & \textbf{D2STGNN} & \textbf{GWNet} &  & \textbf{TAU} & \textbf{PredRNN++} & \textbf{DSAN} & \textbf{ST-ResNet} &  & \textbf{PatchTST} & \textbf{Dlinear} & \textbf{LSTM} & \textbf{HA} \\

\cmidrule{1-3}\cmidrule{4-6}\cmidrule{8-11}\cmidrule{13-16}\cmidrule{18-21}\cmidrule{23-27}

\multirow{6}{*}{Traffic-SH}

& \multirow{3}{*}{Short}
& MAE & \cellcolor{cyan!8}\textbf{0.10} &  0.34 &  0.44 & & 0.39  & 0.34 &  OOM & 0.34 &
& \cellcolor{green!10}$0.24\textcolor{gray}{\text{\scriptsize±0.00}}$ & $0.25\textcolor{gray}{\text{\scriptsize±0.00}}$
& \cellcolor{green!10}$0.24\textcolor{gray}{\text{\scriptsize±0.00}}$ & $0.27\textcolor{gray}{\text{\scriptsize±0.01}}$ &  
& $0.40\textcolor{gray}{\text{\scriptsize±0.00}}$ & $0.32\textcolor{gray}{\text{\scriptsize±0.00}}$
& $1.04\textcolor{gray}{\text{\scriptsize±0.00}}$ & $1.09\textcolor{gray}{\text{\scriptsize±0.00}}$ &  
& $0.30\textcolor{gray}{\text{\scriptsize±0.00}}$ & $0.43\textcolor{gray}{\text{\scriptsize±0.00}}$
& $0.31\textcolor{gray}{\text{\scriptsize±0.00}}$ & $0.92\textcolor{gray}{\text{\scriptsize±0.00}}$ \\

&
& RMSE & \cellcolor{cyan!8}\textbf{0.26} & 0.62 & 0.77 &   & 0.69 & 0.62 &  OOM & 0.64 &
& $0.57\textcolor{gray}{\text{\scriptsize±0.00}}$ & \cellcolor{green!10}$0.50\textcolor{gray}{\text{\scriptsize±0.01}}$
& $0.56\textcolor{gray}{\text{\scriptsize±0.02}}$ & $0.56\textcolor{gray}{\text{\scriptsize±0.02}}$ &  
& $0.74\textcolor{gray}{\text{\scriptsize±0.01}}$ & $0.62\textcolor{gray}{\text{\scriptsize±0.00}}$
& $1.80\textcolor{gray}{\text{\scriptsize±0.00}}$ & $1.80\textcolor{gray}{\text{\scriptsize±0.00}}$ &  
& $0.61\textcolor{gray}{\text{\scriptsize±0.00}}$ & $0.76\textcolor{gray}{\text{\scriptsize±0.00}}$
& $0.62\textcolor{gray}{\text{\scriptsize±0.01}}$ & $1.22\textcolor{gray}{\text{\scriptsize±0.00}}$ \\

&
& MAPE(\%) & \cellcolor{cyan!8}\textbf{1.47} & 5.25 & 7.33 &  & 6.14 & 5.31 &  OOM & 5.29 &
& $4.41\textcolor{gray}{\text{\scriptsize±0.00}}$ & \cellcolor{green!10}$4.23\textcolor{gray}{\text{\scriptsize±0.08}}$
& $4.38\textcolor{gray}{\text{\scriptsize±0.08}}$ & $4.78\textcolor{gray}{\text{\scriptsize±0.18}}$ &  
& $6.86\textcolor{gray}{\text{\scriptsize±0.05}}$ & $5.65\textcolor{gray}{\text{\scriptsize±0.09}}$
& $9.58\textcolor{gray}{\text{\scriptsize±0.01}}$ & $10.70\textcolor{gray}{\text{\scriptsize±0.04}}$ &  
& $5.34\textcolor{gray}{\text{\scriptsize±0.05}}$ & $7.72\textcolor{gray}{\text{\scriptsize±0.02}}$
& $5.57\textcolor{gray}{\text{\scriptsize±0.06}}$ & $12.76\textcolor{gray}{\text{\scriptsize±0.00}}$ \\

\cmidrule{3-27}

& \multirow{3}{*}{Long}
& MAE & \cellcolor{cyan!8}\textbf{0.20} & 0.54 & 0.48 &  & 0.57 & OOM &  OOM & 0.55 &
& $0.33\textcolor{gray}{\text{\scriptsize±0.00}}$ & $0.35\textcolor{gray}{\text{\scriptsize±0.00}}$
& \cellcolor{green!10}$0.32\textcolor{gray}{\text{\scriptsize±0.00}}$ & $0.48\textcolor{gray}{\text{\scriptsize±0.01}}$ &  
& $0.44\textcolor{gray}{\text{\scriptsize±0.01}}$ & $0.41\textcolor{gray}{\text{\scriptsize±0.01}}$
& $1.16\textcolor{gray}{\text{\scriptsize±0.00}}$ & $1.13\textcolor{gray}{\text{\scriptsize±0.01}}$ &  
& $0.47\textcolor{gray}{\text{\scriptsize±0.00}}$ & $0.56\textcolor{gray}{\text{\scriptsize±0.00}}$
& $0.41\textcolor{gray}{\text{\scriptsize±0.00}}$ & $0.93\textcolor{gray}{\text{\scriptsize±0.00}}$ \\

&
& RMSE & \cellcolor{cyan!8}\textbf{0.40} & 0.91 & 0.78 &  & 0.95 & OOM &  OOM & 0.94 &
& $0.66\textcolor{gray}{\text{\scriptsize±0.01}}$ & $0.66\textcolor{gray}{\text{\scriptsize±0.00}}$
& \cellcolor{green!10}$0.64\textcolor{gray}{\text{\scriptsize±0.01}}$ & $0.89\textcolor{gray}{\text{\scriptsize±0.03}}$ &  
& $0.78\textcolor{gray}{\text{\scriptsize±0.01}}$ & $0.76\textcolor{gray}{\text{\scriptsize±0.01}}$
& $1.88\textcolor{gray}{\text{\scriptsize±0.00}}$ & $1.82\textcolor{gray}{\text{\scriptsize±0.00}}$ &  
& $0.88\textcolor{gray}{\text{\scriptsize±0.00}}$ & $0.94\textcolor{gray}{\text{\scriptsize±0.00}}$
& $0.76\textcolor{gray}{\text{\scriptsize±0.00}}$ & $1.23\textcolor{gray}{\text{\scriptsize±0.00}}$ \\

&
& MAPE(\%) & \cellcolor{cyan!8}\textbf{2.78} & 8.54 & 7.58 &  & 9.07 & OOM & OOM & 8.63 &
& $5.83\textcolor{gray}{\text{\scriptsize±0.04}}$ & $6.01\textcolor{gray}{\text{\scriptsize±0.06}}$
& \cellcolor{green!10}$5.73\textcolor{gray}{\text{\scriptsize±0.07}}$ & $8.48\textcolor{gray}{\text{\scriptsize±0.28}}$ &  
& $7.39\textcolor{gray}{\text{\scriptsize±0.10}}$ & $7.28\textcolor{gray}{\text{\scriptsize±0.13}}$
& $11.45\textcolor{gray}{\text{\scriptsize±0.01}}$ & $11.26\textcolor{gray}{\text{\scriptsize±0.11}}$ &  
& $8.46\textcolor{gray}{\text{\scriptsize±0.05}}$ & $10.27\textcolor{gray}{\text{\scriptsize±0.03}}$
& $7.45\textcolor{gray}{\text{\scriptsize±0.08}}$ & $12.82\textcolor{gray}{\text{\scriptsize±0.00}}$ \\

\midrule

\multirow{6}{*}{Bike-NYC}

& \multirow{3}{*}{Short}
& MAE & \textbf{0.73} & 3.05 & 3.58 &   & 3.22 & 3.00 &  3.18 & 3.32 &
& \cellcolor{cyan!8}$0.61\textcolor{gray}{\text{\scriptsize±0.01}}$ & \cellcolor{green!10}$0.64\textcolor{gray}{\text{\scriptsize±0.01}}$
& \cellcolor{green!10}$0.64\textcolor{gray}{\text{\scriptsize±0.01}}$ & \cellcolor{cyan!8}$0.61\textcolor{gray}{\text{\scriptsize±0.01}}$ &  
& $0.81\textcolor{gray}{\text{\scriptsize±0.01}}$ & $1.00\textcolor{gray}{\text{\scriptsize±0.01}}$
& $2.66\textcolor{gray}{\text{\scriptsize±0.00}}$ & $2.70\textcolor{gray}{\text{\scriptsize±0.01}}$ &  
& $0.95\textcolor{gray}{\text{\scriptsize±0.01}}$ & $1.07\textcolor{gray}{\text{\scriptsize±0.00}}$
& $0.74\textcolor{gray}{\text{\scriptsize±0.01}}$ & $3.68\textcolor{gray}{\text{\scriptsize±0.00}}$ \\

&
& RMSE & \cellcolor{green!10}\textbf{1.46} & 5.89 & 7.33 &   & 6.16 & 5.87 &  7.12 & 6.51 &
& \cellcolor{cyan!8}$1.86\textcolor{gray}{\text{\scriptsize±0.01}}$ & $1.93\textcolor{gray}{\text{\scriptsize±0.04}}$
& $1.99\textcolor{gray}{\text{\scriptsize±0.04}}$ & $1.89\textcolor{gray}{\text{\scriptsize±0.02}}$ &  
& $2.31\textcolor{gray}{\text{\scriptsize±0.01}}$ & $3.33\textcolor{gray}{\text{\scriptsize±0.14}}$
& $11.20\textcolor{gray}{\text{\scriptsize±0.00}}$ & $11.17\textcolor{gray}{\text{\scriptsize±0.00}}$ &  
& $3.21\textcolor{gray}{\text{\scriptsize±0.02}}$ & $3.40\textcolor{gray}{\text{\scriptsize±0.00}}$
& $2.33\textcolor{gray}{\text{\scriptsize±0.05}}$ & $6.77\textcolor{gray}{\text{\scriptsize±0.00}}$ \\

&
& MAPE(\%) & \cellcolor{cyan!8}\textbf{9.11} & 43.86 &65.40 &   & 50.27 & 43.20 & 43.18 & 45.85 &
& $33.33\textcolor{gray}{\text{\scriptsize±0.43}}$ & $32.69\textcolor{gray}{\text{\scriptsize±0.17}}$
& $33.26\textcolor{gray}{\text{\scriptsize±1.05}}$ & \cellcolor{green!10}$31.94\textcolor{gray}{\text{\scriptsize±0.23}}$ &  
& $39.05\textcolor{gray}{\text{\scriptsize±0.33}}$ & $45.65\textcolor{gray}{\text{\scriptsize±0.85}}$
& $62.57\textcolor{gray}{\text{\scriptsize±0.16}}$ & $57.51\textcolor{gray}{\text{\scriptsize±0.32}}$ &  
& $43.97\textcolor{gray}{\text{\scriptsize±0.47}}$ & $49.77\textcolor{gray}{\text{\scriptsize±0.16}}$
& $37.91\textcolor{gray}{\text{\scriptsize±0.72}}$ & $55.02\textcolor{gray}{\text{\scriptsize±0.00}}$ \\

\cmidrule{3-27}

& \multirow{3}{*}{Long}
& MAE & \textbf{1.55} &  5.22 & 4.76 &  & 5.70 & 5.27 & OOM & 5.42 &
& $0.90\textcolor{gray}{\text{\scriptsize±0.01}}$ & \cellcolor{green!10}$0.85\textcolor{gray}{\text{\scriptsize±0.01}}$
& \cellcolor{cyan!8}$0.84\textcolor{gray}{\text{\scriptsize±0.02}}$ & $1.08\textcolor{gray}{\text{\scriptsize±0.01}}$ &  
& $0.97\textcolor{gray}{\text{\scriptsize±0.00}}$ & $1.12\textcolor{gray}{\text{\scriptsize±0.01}}$
& $2.87\textcolor{gray}{\text{\scriptsize±0.00}}$ & $2.77\textcolor{gray}{\text{\scriptsize±0.05}}$ &  
& $1.56\textcolor{gray}{\text{\scriptsize±0.00}}$ & $1.65\textcolor{gray}{\text{\scriptsize±0.00}}$
& $1.01\textcolor{gray}{\text{\scriptsize±0.00}}$ & $3.69\textcolor{gray}{\text{\scriptsize±0.00}}$ \\

&
& RMSE & \textbf{2.83} & 10.42 &  10.12 &  & 11.27 & 10.44 & OOM & 11.13 &
& $2.69\textcolor{gray}{\text{\scriptsize±0.03}}$ & \cellcolor{green!10}$2.60\textcolor{gray}{\text{\scriptsize±0.05}}$
& \cellcolor{cyan!8}$2.56\textcolor{gray}{\text{\scriptsize±0.06}}$ & $3.40\textcolor{gray}{\text{\scriptsize±0.05}}$ &  
& $2.81\textcolor{gray}{\text{\scriptsize±0.04}}$ & $3.60\textcolor{gray}{\text{\scriptsize±0.08}}$
& $11.37\textcolor{gray}{\text{\scriptsize±0.00}}$ & $11.23\textcolor{gray}{\text{\scriptsize±0.01}}$ &  
& $5.31\textcolor{gray}{\text{\scriptsize±0.02}}$ & $5.54\textcolor{gray}{\text{\scriptsize±0.00}}$
& $3.19\textcolor{gray}{\text{\scriptsize±0.02}}$ & $6.78\textcolor{gray}{\text{\scriptsize±0.00}}$ \\

&
& MAPE(\%) & \cellcolor{cyan!8}\textbf{18.17} & 65.67 &  76.80 &  & 72.42 & 64.14 & OOM & 66.84 &
& $45.35\textcolor{gray}{\text{\scriptsize±0.70}}$ & \cellcolor{green!10}$40.64\textcolor{gray}{\text{\scriptsize±0.13}}$
& $43.38\textcolor{gray}{\text{\scriptsize±2.37}}$ & $50.47\textcolor{gray}{\text{\scriptsize±1.04}}$ &  
& $46.85\textcolor{gray}{\text{\scriptsize±0.61}}$ & $50.08\textcolor{gray}{\text{\scriptsize±1.00}}$
& $75.22\textcolor{gray}{\text{\scriptsize±0.31}}$ & $58.13\textcolor{gray}{\text{\scriptsize±0.74}}$ &  
& $63.92\textcolor{gray}{\text{\scriptsize±0.42}}$ & $71.63\textcolor{gray}{\text{\scriptsize±0.48}}$
& $48.91\textcolor{gray}{\text{\scriptsize±0.13}}$ & $55.03\textcolor{gray}{\text{\scriptsize±0.00}}$ \\

\midrule

\multirow{6}{*}{Taxi-NYC}

& \multirow{3}{*}{Short}
& MAE & \cellcolor{cyan!8}\textbf{2.19} & 6.40 & 8.07 &   & 9.32 & 6.25  &  6.98 & 7.01 &
& \cellcolor{green!10}$2.50\textcolor{gray}{\text{\scriptsize±0.07}}$ & $2.63\textcolor{gray}{\text{\scriptsize±0.06}}$
& $2.66\textcolor{gray}{\text{\scriptsize±0.17}}$ & $2.64\textcolor{gray}{\text{\scriptsize±0.07}}$ &  
& $3.34\textcolor{gray}{\text{\scriptsize±0.02}}$ & $3.96\textcolor{gray}{\text{\scriptsize±0.04}}$
& $20.06\textcolor{gray}{\text{\scriptsize±0.00}}$ & $19.39\textcolor{gray}{\text{\scriptsize±0.01}}$ &  
& $3.99\textcolor{gray}{\text{\scriptsize±0.00}}$ & $4.23\textcolor{gray}{\text{\scriptsize±0.01}}$
& $3.57\textcolor{gray}{\text{\scriptsize±0.11}}$ & $9.55\textcolor{gray}{\text{\scriptsize±0.00}}$ \\

&
& RMSE & \cellcolor{cyan!8}\textbf{4.98} & 15.47 & 19.28 &   & 20.90 & 15.09 &  19.73 & 17.46 &
& \cellcolor{green!10}$6.73\textcolor{gray}{\text{\scriptsize±0.20}}$ & $6.98\textcolor{gray}{\text{\scriptsize±0.22}}$
& $7.20\textcolor{gray}{\text{\scriptsize±0.47}}$ & $7.10\textcolor{gray}{\text{\scriptsize±0.21}}$ &  
& $8.32\textcolor{gray}{\text{\scriptsize±0.08}}$ & $12.08\textcolor{gray}{\text{\scriptsize±0.28}}$
& $73.55\textcolor{gray}{\text{\scriptsize±0.00}}$ & $73.17\textcolor{gray}{\text{\scriptsize±0.00}}$ &  
& $11.99\textcolor{gray}{\text{\scriptsize±0.03}}$ & $12.17\textcolor{gray}{\text{\scriptsize±0.01}}$
& $10.14\textcolor{gray}{\text{\scriptsize±0.41}}$ & $22.68\textcolor{gray}{\text{\scriptsize±0.00}}$ \\

&
& MAPE(\%) & \cellcolor{cyan!8}\textbf{13.06} &  36.42 & 48.12 &  & 37.24 & 32.93 & 35.15 & 32.71 &
& \cellcolor{green!10}$27.09\textcolor{gray}{\text{\scriptsize±0.34}}$ & $28.96\textcolor{gray}{\text{\scriptsize±0.25}}$
& $28.34\textcolor{gray}{\text{\scriptsize±1.65}}$ & $28.78\textcolor{gray}{\text{\scriptsize±0.43}}$ &  
& $39.61\textcolor{gray}{\text{\scriptsize±0.38}}$ & $37.84\textcolor{gray}{\text{\scriptsize±1.19}}$
& $50.00\textcolor{gray}{\text{\scriptsize±0.11}}$ & $50.37\textcolor{gray}{\text{\scriptsize±0.85}}$ &  
& $35.92\textcolor{gray}{\text{\scriptsize±0.17}}$ & $37.53\textcolor{gray}{\text{\scriptsize±0.19}}$
& $32.60\textcolor{gray}{\text{\scriptsize±0.58}}$ & $42.33\textcolor{gray}{\text{\scriptsize±0.00}}$ \\

\cmidrule{3-27}

& \multirow{3}{*}{Long}
& MAE & \cellcolor{cyan!8}\textbf{3.37} & 12.26 & 11.22 &  & 12.70 & 12.24 &  OOM & 12.85 &
& \cellcolor{green!10}$3.51\textcolor{gray}{\text{\scriptsize±0.02}}$ & $3.58\textcolor{gray}{\text{\scriptsize±0.06}}$
& $3.58\textcolor{gray}{\text{\scriptsize±0.06}}$ & $4.91\textcolor{gray}{\text{\scriptsize±0.04}}$ &  
& $4.11\textcolor{gray}{\text{\scriptsize±0.03}}$ & $4.61\textcolor{gray}{\text{\scriptsize±0.02}}$
& $21.46\textcolor{gray}{\text{\scriptsize±0.02}}$ & $19.63\textcolor{gray}{\text{\scriptsize±0.02}}$ &  
& $7.35\textcolor{gray}{\text{\scriptsize±0.01}}$ & $7.20\textcolor{gray}{\text{\scriptsize±0.01}}$
& $5.14\textcolor{gray}{\text{\scriptsize±0.03}}$ & $9.55\textcolor{gray}{\text{\scriptsize±0.00}}$ \\

&
& RMSE & \cellcolor{cyan!8}\textbf{7.78} & 30.06 & 27.07 &  & 30.48 & 29.75 &  OOM & 32.11 &
& \cellcolor{green!10}$9.58\textcolor{gray}{\text{\scriptsize±0.15}}$ & $9.77\textcolor{gray}{\text{\scriptsize±0.15}}$
& $9.79\textcolor{gray}{\text{\scriptsize±0.26}}$ & $14.10\textcolor{gray}{\text{\scriptsize±0.17}}$ &  
& $10.89\textcolor{gray}{\text{\scriptsize±0.12}}$ & $14.40\textcolor{gray}{\text{\scriptsize±0.48}}$
& $74.52\textcolor{gray}{\text{\scriptsize±0.02}}$ & $73.50\textcolor{gray}{\text{\scriptsize±0.01}}$ &  
& $21.88\textcolor{gray}{\text{\scriptsize±0.04}}$ & $21.77\textcolor{gray}{\text{\scriptsize±0.03}}$
& $14.39\textcolor{gray}{\text{\scriptsize±0.08}}$ & $22.69\textcolor{gray}{\text{\scriptsize±0.00}}$ \\

&
& MAPE(\%) & \cellcolor{cyan!8}\textbf{16.19} & 52.10 & 57.97 &  & 58.22 & 53.01 & OOM & 50.76 &
& \cellcolor{green!10}$33.07\textcolor{gray}{\text{\scriptsize±0.46}}$ & $33.89\textcolor{gray}{\text{\scriptsize±0.33}}$
& $33.89\textcolor{gray}{\text{\scriptsize±1.15}}$ & $38.46\textcolor{gray}{\text{\scriptsize±0.55}}$ &  
& $43.75\textcolor{gray}{\text{\scriptsize±1.54}}$ & $38.90\textcolor{gray}{\text{\scriptsize±0.77}}$
& $64.65\textcolor{gray}{\text{\scriptsize±0.08}}$ & $50.35\textcolor{gray}{\text{\scriptsize±0.60}}$ &  
& $51.52\textcolor{gray}{\text{\scriptsize±0.07}}$ & $50.81\textcolor{gray}{\text{\scriptsize±0.21}}$
& $45.00\textcolor{gray}{\text{\scriptsize±1.23}}$ & $42.32\textcolor{gray}{\text{\scriptsize±0.00}}$ \\

\midrule

\multirow{6}{*}{Tdrive-BJ}

& \multirow{3}{*}{Short}
& MAE & \cellcolor{cyan!8}\textbf{2.38} &  8.05 &  10.87 &  & 10.94 & 7.96 & OOM & 8.56  &
& $5.17\textcolor{gray}{\text{\scriptsize±0.19}}$ & $4.98\textcolor{gray}{\text{\scriptsize±0.16}}$
& $7.31\textcolor{gray}{\text{\scriptsize±0.41}}$ & \cellcolor{green!10}$4.19\textcolor{gray}{\text{\scriptsize±0.04}}$ &  
& $9.03\textcolor{gray}{\text{\scriptsize±0.25}}$ & $4.83\textcolor{gray}{\text{\scriptsize±0.27}}$
& $43.76\textcolor{gray}{\text{\scriptsize±0.00}}$ & $43.30\textcolor{gray}{\text{\scriptsize±0.20}}$ &  
& $6.87\textcolor{gray}{\text{\scriptsize±0.03}}$ & $9.66\textcolor{gray}{\text{\scriptsize±0.01}}$
& $6.64\textcolor{gray}{\text{\scriptsize±0.12}}$ & $21.91\textcolor{gray}{\text{\scriptsize±0.00}}$ \\

&
& RMSE & \cellcolor{cyan!8}\textbf{5.29} &  21.24 & 24.34 &  & 25.32 & 20.18 & OOM & 20.82 &
& $11.82\textcolor{gray}{\text{\scriptsize±0.36}}$ & $11.24\textcolor{gray}{\text{\scriptsize±0.29}}$
& $15.85\textcolor{gray}{\text{\scriptsize±1.04}}$ & \cellcolor{green!10}$10.80\textcolor{gray}{\text{\scriptsize±0.09}}$ &  
& $17.00\textcolor{gray}{\text{\scriptsize±0.36}}$ & $12.29\textcolor{gray}{\text{\scriptsize±0.30}}$
& $134.60\textcolor{gray}{\text{\scriptsize±0.00}}$ & $133.84\textcolor{gray}{\text{\scriptsize±0.10}}$ &  
& $19.34\textcolor{gray}{\text{\scriptsize±0.06}}$ & $22.36\textcolor{gray}{\text{\scriptsize±0.03}}$
& $16.70\textcolor{gray}{\text{\scriptsize±0.19}}$ & $44.59\textcolor{gray}{\text{\scriptsize±0.00}}$ \\

&
& MAPE(\%) & \cellcolor{cyan!8}\textbf{4.49} & 10.95 & 19.56 &  & 15.79 & 11.49 & OOM & 10.94 &
& $21.38\textcolor{gray}{\text{\scriptsize±1.40}}$ & $14.64\textcolor{gray}{\text{\scriptsize±1.27}}$
& $14.06\textcolor{gray}{\text{\scriptsize±1.65}}$ & \cellcolor{green!10}$10.61\textcolor{gray}{\text{\scriptsize±0.41}}$ &  
& $34.87\textcolor{gray}{\text{\scriptsize±0.45}}$ & $14.58\textcolor{gray}{\text{\scriptsize±1.35}}$
& $61.58\textcolor{gray}{\text{\scriptsize±2.26}}$ & $49.66\textcolor{gray}{\text{\scriptsize±0.97}}$ &  
& $10.86\textcolor{gray}{\text{\scriptsize±0.17}}$ & $17.55\textcolor{gray}{\text{\scriptsize±0.09}}$
& $12.95\textcolor{gray}{\text{\scriptsize±0.71}}$ & $26.65\textcolor{gray}{\text{\scriptsize±0.00}}$ \\

\cmidrule{3-27}

& \multirow{3}{*}{Long}
& MAE & \textbf{12.92} &  19.84 &  17.21 &  & 23.68 & OOM &  OOM & 22.64 &
& \cellcolor{cyan!8}$8.37\textcolor{gray}{\text{\scriptsize±0.42}}$ & \cellcolor{green!10}$9.30\textcolor{gray}{\text{\scriptsize±0.09}}$
& $10.94\textcolor{gray}{\text{\scriptsize±0.61}}$ & $14.45\textcolor{gray}{\text{\scriptsize±0.37}}$ &  
& $12.54\textcolor{gray}{\text{\scriptsize±0.05}}$ & $12.25\textcolor{gray}{\text{\scriptsize±0.04}}$
& $49.78\textcolor{gray}{\text{\scriptsize±0.02}}$ & $44.99\textcolor{gray}{\text{\scriptsize±0.29}}$ &  
& $18.10\textcolor{gray}{\text{\scriptsize±0.05}}$ & $20.65\textcolor{gray}{\text{\scriptsize±0.04}}$
& $12.72\textcolor{gray}{\text{\scriptsize±0.00}}$ & $21.93\textcolor{gray}{\text{\scriptsize±0.00}}$ \\

&
& RMSE & \textbf{23.54} & 43.94 & 38.44 &  & 49.65 & OOM &  OOM & 46.92 &
& \cellcolor{cyan!8}$19.44\textcolor{gray}{\text{\scriptsize±0.71}}$ & \cellcolor{green!10}$20.45\textcolor{gray}{\text{\scriptsize±0.14}}$
& $23.49\textcolor{gray}{\text{\scriptsize±0.79}}$ & $32.30\textcolor{gray}{\text{\scriptsize±0.65}}$ &  
& $25.54\textcolor{gray}{\text{\scriptsize±0.21}}$ & $26.85\textcolor{gray}{\text{\scriptsize±0.45}}$
& $137.82\textcolor{gray}{\text{\scriptsize±0.01}}$ & $134.66\textcolor{gray}{\text{\scriptsize±0.15}}$ &  
& $42.55\textcolor{gray}{\text{\scriptsize±0.13}}$ & $43.88\textcolor{gray}{\text{\scriptsize±0.08}}$
& $27.26\textcolor{gray}{\text{\scriptsize±0.00}}$ & $44.62\textcolor{gray}{\text{\scriptsize±0.00}}$ \\

&
& MAPE(\%) & \cellcolor{green!10}\textbf{20.62} & 24.49 &  22.70 &  & 86.22 & OOM & OOM & 25.93 &
& \cellcolor{green!10}$15.59\textcolor{gray}{\text{\scriptsize±0.91}}$ & $21.00\textcolor{gray}{\text{\scriptsize±1.68}}$
& $23.63\textcolor{gray}{\text{\scriptsize±3.61}}$ & $23.65\textcolor{gray}{\text{\scriptsize±0.67}}$ &  
& $41.76\textcolor{gray}{\text{\scriptsize±1.28}}$ & $24.25\textcolor{gray}{\text{\scriptsize±2.43}}$
& $73.46\textcolor{gray}{\text{\scriptsize±2.81}}$ & $55.91\textcolor{gray}{\text{\scriptsize±1.51}}$ &  
& $24.89\textcolor{gray}{\text{\scriptsize±0.17}}$ & $31.87\textcolor{gray}{\text{\scriptsize±0.32}}$
& $24.41\textcolor{gray}{\text{\scriptsize±0.00}}$ & $26.67\textcolor{gray}{\text{\scriptsize±0.00}}$ \\

\midrule

\rowc

\multicolumn{3}{c}{\textbf{$\bf 1^{st}$ count}} 
& \textbf{17} & 0 & 0 & & 0 & 0 & 0 & 0 &  & 4 & 0 & 2 & 1 &  & 0 & 0 & 0 & 0 &  & 0 & 0 & 0 & 0 \\

\bottomrule

\end{tabular}
}
\label{tab:few_full_grid}
\end{table*}

\subsection{More Result on Scaling Evaluation}
\label{appendix_more_rq2}

We provide complete experimental results on the PEMS dataset benchmark, including the pre-trained data capacity of WorldST, the parameter capacity used by the \model~model, and the partitioning parameters of \format. These results are presented in Tables ~\ref{tab:ratio}, ~\ref{tab:layer}, ~\ref{tab:temporal} and ~\ref{tab:spatial}.

\begin{table}[htbp!] %
    \centering
    \begin{minipage}{0.48\linewidth}
        \centering
\centering
\caption{Impact of the scale of pre-training datasets on downstream forecasting performance.}
\label{tab:ratio}
\renewcommand{\arraystretch}{1.2}
\resizebox{\linewidth}{!}{

\begin{tabular}{lllccccc} 
\toprule 
\multicolumn{2}{c}{\multirow{2}{*}{\textbf{Dataset}}} &
\multirow{2}{*}{\textbf{Metric}} & \multicolumn{5}{c}{\textbf{Pre-training Data Ratio}} \\
\cmidrule(lr){4-8} 
 &  &  & 1 & 2 & 4 & 5 & 10 \\ 
\midrule

\multirow{6}{*}{{PEMS03}} 
 & \multirow{3}{*}{Short} & MAE      & 21.87 & 19.73 & 18.43 & 16.78 & {15.62} \\ 
 &                        & RMSE     & 33.47 & 30.60 & 28.28 & 26.29 & {23.24} \\ 
 &                        & MAPE(\%) & 27.67 & 18.96 & 19.70 & 16.35 & {15.24} \\ 
\cmidrule(lr){2-8} 
 & \multirow{3}{*}{Long}  & MAE      & 28.52 & 27.00 & 24.21 & 20.55 & {17.03} \\ 
 &                        & RMSE     & 42.81 & 40.68 & 37.07 & 32.14 & {26.30} \\ 
 &                        & MAPE(\%) & 43.53 & 29.16 & 26.60 & 20.29 & {15.96} \\ 
\midrule

\multirow{6}{*}{{PEMS04}} 
 & \multirow{3}{*}{Short} & MAE      & 28.37 & 25.90 & 24.43 & 22.85 & {20.95} \\ 
 &                        & RMSE     & 43.31 & 40.00 & 37.95 & 36.14 & {32.21} \\ 
 &                        & MAPE(\%) & 23.11 & 18.23 & 17.89 & 15.64 & {13.42} \\ 
\cmidrule(lr){2-8}
 & \multirow{3}{*}{Long}  & MAE      & 36.78 & 35.82 & 32.80 & 29.15 & {23.00} \\ 
 &                        & RMSE     & 54.16 & 53.72 & 49.82 & 45.26 & {35.26} \\ 
 &                        & MAPE(\%) & 21.75 & 28.24 & 25.84 & 20.63 & {15.03} \\ 
\midrule

\multirow{6}{*}{{PEMS07}} 
 & \multirow{3}{*}{Short} & MAE      & 35.19 & 31.67 & 29.30 & 27.13 & {23.71} \\ 
 &                        & RMSE     & 51.07 & 46.90 & 43.33 & 40.61 & {34.99} \\ 
 &                        & MAPE(\%) & 21.75 & 14.80 & 14.38 & 12.97 & {11.26} \\ 
\cmidrule(lr){2-8}
 & \multirow{3}{*}{Long}  & MAE      & 46.10 & 44.72 & 40.38 & 35.36 & {26.39} \\ 
 &                        & RMSE     & 65.03 & 64.69 & 58.91 & 52.35 & {38.93} \\ 
 &                        & MAPE(\%) & 33.35 & 22.84 & 20.88 & 16.90 & {12.09} \\ 
\midrule

\multirow{6}{*}{{PEMS08}} 
 & \multirow{3}{*}{Short} & MAE      & 24.40 & 21.38 & 19.90 & 18.60 & {17.70} \\ 
 &                        & RMSE     & 35.32 & 31.72 & 29.86 & 28.60 & {25.80} \\ 
 &                        & MAPE(\%) & 24.29 & 15.40 & 15.64 & 14.61 & {13.18} \\ 
\cmidrule(lr){2-8}
 & \multirow{3}{*}{Long}  & MAE      & 31.02 & 29.57 & 26.60 & 23.27 & {19.19} \\ 
 &                        & RMSE     & 44.05 & 43.00 & 39.46 & 35.40 & {28.57} \\ 
 &                        & MAPE(\%) & 34.02 & 21.92 & 20.85 & 17.54 & {13.77} \\ 
\bottomrule

\end{tabular}
}
    \end{minipage}
    \hfill 
    \begin{minipage}{0.48
    \linewidth}
        \centering
\centering
\caption{Performance comparison with different numbers of spatial and temporal layers in \model.}
\label{tab:layer}
\renewcommand{\arraystretch}{1.2}
\resizebox{\linewidth}{!}{

\begin{tabular}{lllccccc} 
\toprule 
\multicolumn{2}{c}{\multirow{2}{*}{{\textbf{Dataset}}}} &
\multirow{2}{*}{{\textbf{Metric}}} & \multicolumn{5}{c}{{\textbf{Layer Number}}} \\
\cmidrule(lr){4-8} 
 &  &  & 2 & 4 & 6 & 8 & 10 \\ 
\midrule

\multirow{6}{*}{{PEMS03}} 
 & \multirow{3}{*}{Short} & MAE      & 19.73 & 16.89 & 15.28 & {15.62} & 16.06  \\ 
 &                        & RMSE     & 28.92 & 26.22 & 23.79 & {23.24} & 25.27  \\ 
 &                        & MAPE(\%) & 23.01 & 16.29 & 15.20 & {15.24} & 15.01  \\ 
\cmidrule(lr){2-8} 
 & \multirow{3}{*}{Long}  & MAE      & 17.87 & 17.31 & 17.04 & {17.03} & 17.26  \\ 
 &                        & RMSE     & 27.84 & 26.87 & 26.51 & {26.30} & 26.70  \\ 
 &                        & MAPE(\%) & 18.61 & 17.37 & 17.01 & {15.96} & 16.98  \\ 
\midrule

\multirow{6}{*}{{PEMS04}} 
 & \multirow{3}{*}{Short} & MAE      & 26.14 & 22.83 & 20.89 & {20.95} & 21.83  \\ 
 &                        & RMSE     & 38.89 & 36.14 & 33.11 & {32.21} & 34.68  \\ 
 &                        & MAPE(\%) & 19.87 & 15.76 & 14.64 & {13.42} & 14.74  \\ 
\cmidrule(lr){2-8}
 & \multirow{3}{*}{Long}  & MAE      & 25.45 & 24.56 & 24.24 & {23.00} & 24.48  \\ 
 &                        & RMSE     & 39.91 & 38.49 & 38.08 & {35.26} & 38.27  \\ 
 &                        & MAPE(\%) & 18.48 & 17.74 & 17.37 & {15.03} & 17.49  \\ 
\midrule

\multirow{6}{*}{{PEMS07}} 
 & \multirow{3}{*}{Short} & MAE      & 31.31 & 27.19 & 24.55 & {23.71} & 25.59  \\ 
 &                        & RMSE     & 44.15 & 40.92 & 36.86 & {34.99} & 38.51  \\ 
 &                        & MAPE(\%) & 17.03 & 12.33 & 11.92 & {11.26} & 11.60  \\ 
\cmidrule(lr){2-8}
 & \multirow{3}{*}{Long}  & MAE      & 30.36 & 29.05 & 28.68 & {26.39} & 29.25  \\ 
 &                        & RMSE     & 45.18 & 43.49 & 42.94 & {38.93} & 43.57  \\ 
 &                        & MAPE(\%) & 15.26 & 13.76 & 13.57 & {12.09} & 13.93  \\ 
\midrule

\multirow{6}{*}{{PEMS08}} 
 & \multirow{3}{*}{Short} & MAE      & 22.12 & 19.29 & 17.14 & {17.70} & 17.80  \\ 
 &                        & RMSE     & 31.64 & 29.48 & 26.38 & {25.80} & 27.56  \\ 
 &                        & MAPE(\%) & 18.60 & 13.53 & 13.68 & {13.18} & 12.88  \\ 
\cmidrule(lr){2-8}
 & \multirow{3}{*}{Long}  & MAE      & 20.38 & 19.66 & 19.46 & {19.19} & 19.94  \\ 
 &                        & RMSE     & 31.21 & 30.34 & 30.07 & {28.57} & 30.45  \\ 
 &                        & MAPE(\%) & 16.33 & 14.47 & 14.36 & {13.77} & 15.11  \\ 
\bottomrule

\end{tabular}
}
    \end{minipage}
\end{table}

\begin{table}[htbp!] %
    \centering
    \begin{minipage}{0.48\linewidth}
        \centering
\centering
\caption{Performance comparison on temporal sizes.}
\label{tab:temporal}
\footnotesize
\setlength{\tabcolsep}{6pt}       
\renewcommand{\arraystretch}{1.0}
\newcolumntype{Y}{>{\centering\arraybackslash}X}

\begin{tabularx}{\textwidth}{lllYYYY} 
\toprule

\multicolumn{2}{l}{\multirow{2}{*}{\textbf{Dataset}}} & 
\multirow{2}{*}{\textbf{Metric}} & 
\multicolumn{4}{c}{\textbf{Temporal Window Size}} \\ 
\cmidrule(lr){4-7} 
 & & & 24 & 48 & 64 & 128 \\ 
\midrule

\multirow{6}{*}{{PEMS03}} 
 & \multirow{3}{*}{Short} & MAE      & 14.84 & {15.62} & 18.77 & 19.79 \\ 
 &                        & RMSE     & 23.19 & {23.24} & 28.93 & 30.45 \\ 
 &                        & MAPE(\%) & 14.10 & {15.24} & 17.93 & 20.38 \\ 
\cmidrule(lr){2-7} 
 & \multirow{3}{*}{Long}  & MAE      & 23.15 & {17.03} & 17.69 & 25.82 \\ 
 &                        & RMSE     & 35.75 & {26.30} & 27.48 & 40.28 \\ 
 &                        & MAPE(\%) & 22.50 & {15.96} & 17.79 & 26.10 \\ 
\midrule

\multirow{6}{*}{{PEMS04}} 
 & \multirow{3}{*}{Short} & MAE      & 20.80 & {20.95} & 24.74 & 25.97 \\ 
 &                        & RMSE     & 33.18 & {32.21} & 38.99 & 40.14 \\ 
 &                        & MAPE(\%) & 14.20 & {13.42} & 16.99 & 18.17 \\ 
\cmidrule(lr){2-7}
 & \multirow{3}{*}{Long}  & MAE      & 31.48 & {23.00} & 24.69 & 33.00 \\ 
 &                        & RMSE     & 47.25 & {35.26} & 38.87 & 50.39 \\ 
 &                        & MAPE(\%) & 23.04 & {15.03} & 17.57 & 22.72 \\ 
\midrule

\multirow{6}{*}{{PEMS07}} 
 & \multirow{3}{*}{Short} & MAE      & 24.05 & {23.71} & 30.33 & 32.25 \\ 
 &                        & RMSE     & 36.75 & {34.99} & 45.11 & 47.41 \\ 
 &                        & MAPE(\%) & 10.84 & {11.26} & 14.49 & 16.26 \\ 
\cmidrule(lr){2-7}
 & \multirow{3}{*}{Long}  & MAE      & 38.01 & {26.39} & 29.35 & 41.12 \\ 
 &                        & RMSE     & 55.61 & {38.93} & 43.76 & 59.82 \\ 
 &                        & MAPE(\%) & 19.08 & {12.09} & 14.26 & 20.64 \\ 
\midrule

\multirow{6}{*}{{PEMS08}} 
 & \multirow{3}{*}{Short} & MAE      & 16.53 & {17.70} & 22.27 & 22.63 \\ 
 &                        & RMSE     & 25.88 & {25.80} & 33.26 & 33.72 \\ 
 &                        & MAPE(\%) & 11.37 & {13.18} & 17.46 & 19.05 \\ 
\cmidrule(lr){2-7}
 & \multirow{3}{*}{Long}  & MAE      & 26.00 & {19.19} & 20.05 & 28.62 \\ 
 &                        & RMSE     & 38.50 & {28.57} & 30.81 & 41.91 \\ 
 &                        & MAPE(\%) & 18.42 & {13.77} & 15.50 & 23.06 \\ 
\bottomrule

\end{tabularx}
    \end{minipage}
    \hfill 
    \begin{minipage}{0.48
    \linewidth}
        \centering
\centering
\caption{Performance comparison on spatial sizes.}
\label{tab:spatial}
\footnotesize
\setlength{\tabcolsep}{6pt}       
\renewcommand{\arraystretch}{1.0}
\newcolumntype{Y}{>{\centering\arraybackslash}X} 

\begin{tabularx}{\textwidth}{lllYYYY} 
\toprule

\multicolumn{2}{l}{\multirow{2}{*}{\textbf{Dataset}}} & 
\multirow{2}{*}{\textbf{Metric}} & 
\multicolumn{4}{c}{\textbf{Spatial Window Size}} \\ 
\cmidrule(lr){4-7} 
 & & & 16 & 48 & 64 & 128 \\ 
\midrule

\multirow{6}{*}{{PEMS03}} 
 & \multirow{3}{*}{Short} & MAE      & 15.62 & 15.86 & 16.74 & 17.55 \\ 
 &                        & RMSE     & 23.24 & 25.63 & 27.15 & 29.27 \\ 
 &                        & MAPE(\%) & 15.24 & 15.61 & 16.93 & 17.46 \\ 
\cmidrule(lr){2-7} 
 & \multirow{3}{*}{Long}  & MAE      & 17.03 & 16.71 & 17.44 & 17.53 \\ 
 &                        & RMSE     & 26.30 & 26.68 & 28.11 & 28.77 \\ 
 &                        & MAPE(\%) & 15.96 & 17.27 & 18.11 & 18.04 \\ 
\midrule

\multirow{6}{*}{{PEMS04}} 
 & \multirow{3}{*}{Short} & MAE      & 20.95 & 22.65 & 23.38 & 20.35 \\ 
 &                        & RMSE     & 32.21 & 36.00 & 37.28 & 35.24 \\ 
 &                        & MAPE(\%) & 13.42 & 15.42 & 15.88 & 16.19 \\ 
\cmidrule(lr){2-7}
 & \multirow{3}{*}{Long}  & MAE      & 23.00 & 24.73 & 25.37 & 21.27 \\ 
 &                        & RMSE     & 35.26 & 38.70 & 39.94 & 36.80 \\ 
 &                        & MAPE(\%) & 15.03 & 17.61 & 18.15 & 17.73 \\ 
\midrule

\multirow{6}{*}{{PEMS07}} 
 & \multirow{3}{*}{Short} & MAE      & 23.71 & 27.31 & 28.02 & 29.09 \\ 
 &                        & RMSE     & 34.99 & 41.29 & 42.39 & 43.97 \\ 
 &                        & MAPE(\%) & 11.26 & 13.53 & 15.03 & 15.00 \\ 
\cmidrule(lr){2-7}
 & \multirow{3}{*}{Long}  & MAE      & 26.39 & 29.87 & 30.52 & 30.62 \\ 
 &                        & RMSE     & 38.93 & 44.53 & 45.54 & 45.66 \\ 
 &                        & MAPE(\%) & 12.09 & 15.24 & 15.50 & 15.91 \\ 
\midrule

\multirow{6}{*}{{PEMS08}} 
 & \multirow{3}{*}{Short} & MAE      & 17.70 & 17.44 & 17.87 & 13.83 \\ 
 &                        & RMSE     & 25.80 & 28.24 & 28.74 & 25.53 \\ 
 &                        & MAPE(\%) & 13.18 & 16.58 & 18.45 & 16.77 \\ 
\cmidrule(lr){2-7}
 & \multirow{3}{*}{Long}  & MAE      & 19.19 & 18.66 & 18.72 & 14.01 \\ 
 &                        & RMSE     & 28.57 & 29.98 & 30.21 & 26.13 \\ 
 &                        & MAPE(\%) & 13.77 & 18.00 & 17.50 & 17.14 \\ 
\bottomrule

\end{tabularx}
    \end{minipage}
\end{table}

\section{More Disscuison}\label{disscuison}

\subsection{Ethics, Fairness, and Limitations}

Despite the significant efforts we have made in this work, we still emphasize the following potential risks:

\subsubsection{Privacy and Anonymity.}
Although \dataset~aggregates data at the flow level to protect anonymity, theoretical re-identification risks persist. We strictly adhere to differential privacy principles during pre-processing to mitigate these risks, ensuring that no single user's data significantly influences macroscopic patterns.

\subsubsection{Fairness in Modeling.}
Data coverage is inherently biased towards developed regions with better sensor infrastructure, potentially leading to a ``rich-get-richer'' performance disparity. We acknowledge this inevitable reality and highlight that addressing such regional fairness remains a critical direction for future research.

\subsubsection{Limitations.}
Currently, \model~focuses on structured Euclidean Sensor-based and Grid-based data. It does not yet fully integrate unstructured modalities, such as text (\eg, traffic incident reports) or visual feeds (\eg, CCTV footage), which represents a key future avenue for developing truly multi-modal urban foundation models.

\subsection{Future Direction}

Based on these analyses, we identify three critical avenues for future work to advance urban spatio-temporal foundation models:


\subsubsection{Democratizing Urban AI via Few-Shot Transfer.} 
To mitigate the ``rich-get-richer'' disparity~\citep{lin2024fairstg} in model performance, future research must focus on data-efficient adaptation techniques. We aim to develop robust few-shot learning or parameter-efficient fine-tuning (PEFT) strategies that allow the foundation model—trained on data-rich metropolises—to be effectively adapted to underdeveloped regions with sparse sensor infrastructure, thereby promoting global equity in urban intelligence.

\subsubsection{Multi-Modal Alignment with LLMs.} 
Current models are confined to numerical sensor readings. A key frontier is aligning these quantitative spatio-temporal representations with qualitative semantic knowledge from Large Language Models~\citep{liu2025st,liu2025urbanmind}. Integrating unstructured text (e.g., event descriptions, traffic reports) will enhance the model's interpretability and reasoning capabilities, enabling it to explain \textit{why} a congestion occurs, not just predict \textit{when}.

\subsubsection{From Forecasting to Decision Making.} 
Forecasting is merely the first step. We envision evolving \model~from a passive predictor into an active \textit{World Model} for urban system. By accurately simulating complex urban dynamics, the foundation model can serve as a simulator for Reinforcement Learning (RL) agents~\citep{wu2025spatiotemporal}, empowering downstream tasks such as traffic signal control and emergency dispatch optimization.

\end{document}